%% file: approx_ml.tex
\documentclass[sigconf,10pt]{acmart}

\usepackage{booktabs} % For formal tables

\pdfoutput=1

\input{vldb_preamble}

\newif\ifsubmit
\submitfalse

\newif\iftech
\techtrue  %or
\ifsubmit
\newcommand{\barzan}[1]{}
\newcommand{\yongjoo}[1]{}
\newcommand{\tofix}[1]{\xspace{\color{red} #1}\xspace}
\else
\newcommand{\barzan}[1]{}
\newcommand{\yongjoo}[1]{}
\newcommand{\tofix}[1]{\xspace{\color{red} #1}\xspace}
\fi
\newcommand{\cancut}[1]{}

\newcommand{\ph}[1]{\vspace{2mm} \noindent \textbf{#1}\;}

\newcommand{\ignore}[1]{}

\newcommand{\system}{\textsc{BlinkML}\xspace}
\newcommand{\systemtitle}{BlinkML\xspace}

\newcommand{\gradual}{IncEstimator\xspace}

\newcommand{\mcs}{MCS\xspace}

\newcommand{\xx}{\bm{x}}

\newcommand{\ee}{\bm{\varepsilon}}

\DeclareMathOperator{\argmin}{arg\,min}
\DeclareMathOperator{\argmax}{arg\,max}
\DeclareMathOperator{\erf}{erf}

\DeclareMathOperator{\pr}{Pr}
\DeclareMathOperator{\cov}{Cov}
\DeclareMathOperator{\var}{Var}
\DeclareMathOperator{\tr}{tr}
\DeclareMathOperator{\E}{E}
\newcommand{\1}[1]{\mathds{1}\left[#1\right]}

\newcommand{\power}{\texttt{Power}\xspace}
\newcommand{\gas}{\texttt{Gas}\xspace}
\newcommand{\criteo}{\texttt{Criteo}\xspace}
\newcommand{\higgs}{\texttt{HIGGS}\xspace}
\newcommand{\yelp}{\texttt{Yelp}\xspace}
\newcommand{\mnist}{\texttt{MNIST}\xspace}

\acmArticle{4}
\acmPrice{15.00}

\copyrightyear{2019} 
\acmYear{2019} 
\setcopyright{acmlicensed}
\acmConference[SIGMOD '19]{2019 International Conference on Management of Data}{June 30-July 5, 2019}{Amsterdam, Netherlands}
\acmBooktitle{2019 International Conference on Management of Data (SIGMOD '19), June 30-July 5, 2019, Amsterdam, Netherlands}
\acmPrice{15.00}
\acmDOI{10.1145/3299869.3300077}
\acmISBN{978-1-4503-5643-5/19/06}

\begin{document}
\title[\systemtitle: Efficient Maximum Likelihood Estimation]{\systemtitle: Efficient Maximum Likelihood Estimation with Probabilistic Guarantees}

% Updat the paper id
\author{Yongjoo Park, Jingyi Qing, Xiaoyang Shen, Barzan Mozafari}
\affiliation{%
 \institution{University of Michigan, Ann Arbor}
}
\email{{pyongjoo,jyqing,xyshen,mozafari}@umich.edu}

%
%\author{Yongjoo Park}
%\affiliation{%
%  \institution{University of Michigan}
%  \city{Ann Arbor}
%  \state{Michigan}
%}
%\email{pyongjoo@umich.edu}
%
%\author{Jingyi Qing}
%\affiliation{%
%  \institution{University of Michigan}
%  \city{Ann Arbor}
%  \state{Michigan}
%}
%\email{jyqing@umich.edu}
%
%\author{Xiaoyang Shen}
%\affiliation{%
%  \institution{University of Michigan}
%  \city{Ann Arbor}
%  \state{Michigan}
%}
%\email{xyshen@umich.edu}
%
%\author{Barzan Mozafari}
%\affiliation{%
%  \institution{University of Michigan}
%  \city{Ann Arbor}
%  \state{Michigan}
%}
%\email{mozafari@umich.edu}

% \renewcommand{\shortauthors}{B. Trovato et al.}

\begin{abstract}
The rising volume of datasets has made training machine learning (ML) models
 a major computational cost in the enterprise.
 %ta inja
 Given the \emph{iterative} nature of model and parameter tuning, many analysts use a small sample of their entire data during their \emph{initial} stage of analysis to make quick decisions (e.g., what features or hyperparameters to use)
%  what type of model to pursue), 
 and use the entire dataset only in later stages (i.e., when they have converged to a specific model). This sampling, however, is performed in an ad-hoc fashion. Most practitioners
% are not able to
cannot
precisely capture the effect of sampling on the quality of their model, and eventually on their decision-making process during the tuning phase. Moreover, without systematic support for sampling operators, many optimizations and reuse opportunities are lost.

In this paper, we introduce \system, a system for \emph{fast, quality-guaranteed ML training.} 
% for \tofix{the models based on maximum likelihood estimation (MLE).} 
\system allows users to make error-computation tradeoffs: instead of training a model on their full data (i.e., \emph{full model}), \system can quickly train an \emph{approximate model} with quality guarantees using a sample. 
The quality guarantees ensure that, with high probability, the approximate model makes the same predictions as the full model.
\system currently supports any ML model that relies on \emph{maximum likelihood estimation} (MLE), which 
% \tofix{The MLE-based models}
includes Generalized Linear Models (e.g., linear regression, logistic regression, max entropy classifier, Poisson regression) as well as PPCA (Probabilistic Principal Component Analysis). Our experiments show that \system can speed up the training of large-scale ML tasks by 6.26$\times$--629$\times$
while guaranteeing the same predictions, with 95\% probability, as the full model.
\end{abstract}

\maketitle

\input{introduction}

\input{overview}

% \input{archi}

% methods

\input{accuracy}

\input{sample}

% Experiment

\input{exp}

% Others

\input{related}

\input{conclusion}

% \vspace{2mm}

% \bibliographystyle{ACM-Reference-Format}
\bibliographystyle{abbrv}
\bibliography{biblio/approxml,biblio/related,biblio/approximate,biblio/mozafari}

\appendix
% % \balance

\input{abstraction_example}

\input{proofs}

% \input{additional_exp}

\input{model_sim_def}

\iftech
\input{raw_data}
\else
\fi

\end{document}

%% file: vldb_preamble.tex
\usepackage[T1]{fontenc}

\usepackage[utf8]{inputenc}
\usepackage{graphicx}
\usepackage{xcolor}
\usepackage{balance}
\usepackage{booktabs,multirow} % For formal tables
\usepackage[shortlabels]{enumitem}
\setenumerate{1.,nolistsep,noitemsep,leftmargin=10pt}
\usepackage[font={small}]{caption}
\usepackage[font={small}]{subcaption}
\usepackage[noend]{algpseudocode}
\usepackage{xspace}

\usepackage{amsmath}
\usepackage{amssymb}
\usepackage{mathtools}
\usepackage{bm}
\usepackage{dsfont}
\usepackage[normalem]{ulem}

\usepackage{amsthm}
\newtheorem{theorem}{Theorem}
\newtheorem{lemma}{Lemma}
\theoremstyle{definition}

\usepackage{tikz}
\usetikzlibrary{calc,pgfplots.groupplots,arrows.meta,shapes}
\usetikzlibrary{decorations.pathreplacing}
\usepackage{pgfplots}
\usepackage{pgfplots}
\usepgfplotslibrary{fillbetween}

\pgfplotsset{every axis/.append style={
                    label style={font=\scriptsize},
                    tick label style={font=\scriptsize}
}}

\input{pgfplot_styles}

\usepackage{pifont}% http://ctan.org/pkg/pifont
\newcommand{\xmark}{\ding{55}}%
\newcommand{\minus}{\scalebox{0.75}[1.0]{$-$}}

\usepackage{listings}

\usepackage{hyperref}
\hypersetup{
    colorlinks=true,
    allcolors=blue,
}
\usepackage[noabbrev,capitalize]{cleveref}

\definecolor{vintageblack}{HTML}{484043}
\definecolor{vintageyellow}{HTML}{FFC805}
\definecolor{vintagegreen}{HTML}{3C948B}
\definecolor{vintageorange}{HTML}{FF4D25}
\definecolor{vintagepurple}{HTML}{913A5D}

\renewcommand{\footnotesize}{\scriptsize}

%% file: pgfplot_styles.tex
\definecolor{vintageblack}{HTML}{484043}
\definecolor{vintageyellow}{HTML}{FFC805}
\definecolor{vintagegreen}{HTML}{3C948B}
\definecolor{vintageorange}{HTML}{FF4D25}

\pgfplotsset{
    compat=newest,
    every axis legend/.append style={font=\scriptsize, column sep=5pt},
    every axis/.append style={font=\scriptsize},
    /pgfplots/ybar legend/.style={
        /pgfplots/legend image code/.code={
        \draw[##1,/tikz/.cd, bar width=6pt, yshift=-0.25em, bar shift=0pt, xshift=0.8em]
        plot coordinates {(0cm,0.8em)};
        }
    }
}

%% file: introduction.tex
%!TEX root = approx_ml.tex

\section{Introduction}
\label{sec:intro}

% \barzan{note for later: 1) inject somewhere that we extend a popular ML library that's supported by most dbms
% 2) see if we can claim that our implementation can sometimes write and send proper SQL queries
% for faster sampling}

% \ph{Motivation}
% While data management systems have been quite successful in supporting
% 	traditional OLAP-style (i.e., SQL-based) analytics,
% 		they have not been as successful in
% 			attracting modern analytics, which are increasingly statistical (i.e., machine learning workloads).
% Consequently, nearly every analytical database vendors
% 	provides integration interfaces for R~\cite{citeAsManyAsYouCan}, Python libraries~\cite{CiteAsManyAsPossible}, and even TensorFlow~\cite{CiteDBMSLinksThatSupportTensorFlow}
% 		and \barzan{see what else they support}~\cite{blah}.

While data management systems have been widely successful in supporting traditional OLAP-style
 analytics,
they have \emph{not} been equally successful in attracting modern machine learning  (ML) workloads.
To circumvent this,
most analytical database vendors have added integration layers for popular ML libraries in Python
(e.g., Oracle's cx\textunderscore Oracle \cite{cxoracle},  SQL Server's pymssql \cite{pymssql},
and DB2's ibm\textunderscore db  \cite{ibmdb})
or   R
(e.g., Oracle's RODM \cite{rodm},
 SQL Server's  RevoScaleR  \cite{revoscaler},
 and
 DB2's  ibmdbR \cite{ibmdbR}).
% \tofix{are limited to}
	% has added an interface for integrating
	% popular tools, such as
\ignore{
\tofix{Oracle Machine Learning~\cite{oracleservices},
SQL Server ML Services~\cite{sqlserverml},
Spark driver for IBM DB2~\cite{db2spark}, etc.
Also, open-source communities have developed ML layers on top of existing distributed databases, e.g.,
Spark MLlib~\cite{meng2016mllib},
Apache MADlib on Greenplum~\cite{hellerstein2012madlib},
% Apache Ignite on HDFS~\cite{apacheignite},
Apache Mahout on Spark~\cite{apachemahout}, etc.}}
These interfaces simply allow machine learning algorithms to run on the data \emph{in-situ}.

% 	R~\cite{sparkrstudio,sparkr,implyr,hiver,redshiftr}, Python libraries~\cite{pyspark,impalapy,hivepy,verticapy}, SAS~\cite{sasmysql,sasmssql}, \barzan{what does SAS have to do w/ ML?
% 	i thought u were going to search for more ML-based layers in DBMSs}
% 	 and even TensorFlow~\cite{tensorbigquery}. \barzan{might be better to name actual DBMS products instead of just listing some citations}
% 		% and \barzan{see what else they support}~\cite{blah}.
% \tofix{These interfaces simply allow ML libraries to import the data from relational databases.}

% \barzan{i think u re wrong. look into GreenPlum and many others such as SciDB who claim a perfect integration
% and better performance. 1) u need to make this more comprehensive and 2) enumerate the different ways in which
% DBMSs allow for ML}

However, recent efforts have shown that data management systems have much more to offer.
% \tofix{That is, familiar database functionalities
% 		can deliver significant speedups for various ML workloads.
% Those DBMS-inspired approaches include}
% \barzan{u screwed this up. plz change back to what i had here. plz dont make changes unless i have marked them as red}
% For example,
For example, materialization and reuse opportunities~\cite{zhang2014materialization,anderson2016input,anderson2013brainwash,crankshaw2017clipper,shin2015incremental},
	cost-based optimization of linear algebraic operators~\cite{ghoting2011systemml,boehm2016systemml,cohen2009mad},
	 array-based representations~\cite{stonebraker2013scidb,kersten2011sciql},
	 avoiding denormalization~\cite{kumar2015learning,schleich2016learning,kumar2016join},
 lazy evaluation~\cite{zhang2010efficient},
declarative interfaces~\cite{weimer2011machine,olston2008pig,sparks2017keystoneml},
and query planning~\cite{pan2017hemingway,kumar2016model,sparks2015automating}
 are all readily available (or at least familiar) database functionalities
		that can deliver significant speedups for various ML  workloads.

% \tofix{model management}~\cite{vartak2016m,miao2016modelhub}, and so on.
	 % \barzan{they avoid the join, so they do denormalization, no??}
	 % \barzan{replace this w/ whatever technique arun has for delaying joins}~\cite{AsManyPapersAsYouCan}

% \barzan{this paragraph is not comprehensive both in terms of approaches but more importantly in terms
% of citations. that's why i asked u to look at Arun's related work sections (esp his SIGMOD tutorial).
% u re offending the PC members in this paragraph for no good reason}

One additional but key opportunity that has been largely overlooked is the
\emph{sampling abstraction} offered by nearly every database system.
Sampling operators have been mostly used for approximate query processing
(AQP)~\cite{galakatos2017revisiting,crotty2015vizdom,wander-join,online-agg-mr1,chakrabarti:wavelets,dremel,dbo,mozafari_sigmod2017_dbl}.
However, applying the lessons learned in the data management community regarding AQP,
	we could use a similar sampling abstraction to also speed up an important class of ML workloads.
\ph{Our Goal}
Given that (sub)sampling is already quite common in early stages of
	ML workloads---such as feature selection and hyper-parameter tuning---we propose
	 a high-level system abstraction for training ML models,
		with which analysts can explicitly
		request error-computation trade-offs for several important classes of ML models.
This involves systematic support for
(i)
% predicting and
% guaranteeing the maximum error given a \tofix{sample-based model},
bounding the deviation of the approximate model's predictions
	from those of the full model given a sample size,
% (ii) bounding the deviation of the approximate model's parameters
	% from those of the full model given a sample size,
and (ii) predicting the \emph{minimum sample size} with which the trained model would meet a given prediction
	 error.

\ph{Challenges}
While estimating sampling error is a  well-studied problem
	for SQL queries~\cite{approx_chapter,mozafari_sigmod2014_diagnosis},
	it is more involved for ML models.
There are two types of approaches here:
(i) those that estimate the error before training the model (i.e., \emph{predictive}),
	and (ii) those that estimate the error after a model is trained (i.e., \emph{descriptive}).
A well-known predictive technique  is the so-called
	VC-dimension~\cite{mohri2012foundations}, which upper bounds the generalization error of a model.
However, given that VC-dimension bounds are data-independent,
	they tend to be quite loose in practice \cite{vcdlecture}.
Overestimated error bounds would lead the analysts to use
% means that the analysts still have to use
the entire dataset even if
	similar results could be obtained from a small sample.
\footnote{This is why
	VC-dimensions are sometimes used indirectly, as a comparative measure of quality~\cite{kumar2016join}.}

		% a sample-based model.
	% In many cases, the entire dataset is used anyway.
% The VC dimension can be used for bounding a model's generalization error.
% Since the VC dimension is proportional to a data dimension (in case of linear models),
% the change of the VC dimension (thus, the increment of the generalization error) is used for determining whether to perform a join.
% However, this VC dimension-based approach is too conservative that in many cases,  even when not required.
% Also, this approach cannot estimate the error of a sample-based model.}
%
	% This means in most cases the error bound will be so large that
	% 	analysts still has to use the entire dataset even if it leads to similar results.
		 % as a sample-based model.
% which bounds the generalization error of a model
% 		using
% The most popular tool for the predictive approach is the so-called
% 	VC-dimension, }
		% \barzan{replace with a 4-5 word summary of VC's approach.}
% Another predictive techniques is choosing a representative subset of the training set, then assumes that the model trained on that representative subset well approximates the model

Common techniques for the descriptive approach include
cross-validation~\cite{bishop2006pattern} and Radamacher complexity~\cite{mohri2012foundations}.
Since these techniques are data-dependent, they provide tighter error estimates.
However, they only bound the generalization error.
While useful for evaluating the model's quality on future (i.e., unseen) data,
	the generalization error provides little help in predicting how much
		the model quality would differ
		% be improved
		if the entire dataset were
		% \tofix{a larger dataset is}
		 used instead of the current sample.
Furthermore, when
choosing the minimum sample size needed to achieve a user-specified accuracy,
the descriptive approaches can be quite expensive:
 one would need to train multiple models, each on a different sample size,
	 until a desirable error tolerance is met.
Given that most ML models do not have an incremental training procedure (besides a warm start~\cite{chu2015warm}),
 training multiple models to find an appropriate sample size might  take longer overall than simply training a  model
 on the full dataset (see \cref{sec:exp:sample}).

% \yongjoo{This paragraph is out of context.}
% In fact, a key observation in this paper is the following:
% 	from the analyst's perspective, the relevant question that helps him/her
% 		decide whether to train on a sample rather than the entire data---and if so, on what size of a sample---is
% 			the following:
% 	how much do the parameters (or predictions) of the model trained on a sample of size $s$
% 		  deviate from those of the model trained on the entire dataset (of size $n$, where $s\leq n$)?
% This is the key question that can quantitatively justify the use (or not) of sampling in initial phases of model tuning
% 	or feature engineering/selection.
% \barzan{i noticed i have not said anything about why it's technically challenging. so sprinkle that in between my text or add it here at the end}

\ph{Our Approach}
\system's underlying statistical technique offers tight error bounds for approximate models (\cref{sec:accuracy}), and does so without
 % actually
training multiple approximate models (\cref{sec:sample}).
% \system supports the ML models that relies on maximum likelihood estimation for its training, which
% includes Generalized Linear Models (e.g., linear regression, logistic regression, max entropy classifier, Poisson regression) and Probabilistic Principal Component Analysis.
% Moreover, \system's system abstractions (\cref{sec:archi})
	% significantly reduce the user effort in applying
% these statistical techniques    to new models.

\system's statistical techniques are based on the following observation:
given a test example $\xx$, the model's prediction is simply a function $m(\xx; \theta)$, where $\theta$ is the model
parameter learned during the training phase.
Therefore, if we could understand how $\theta$ would differ when trained on a sample (instead of the entire dataset), we could also infer its impact on the
model's prediction, i.e., $m(\xx; \theta)$.

Specifically, let $\theta_N$ be the model parameter obtained if one trains on the entire dataset (say, of size $N$),
 and $\theta_n$ be the model parameter obtained if one trains on a sample of size $n$.
% \footnote{Note that $\hat{\theta}_n$ is a random variable; a specific model parameter $\theta_n$ trained on a specific sample (of size $n$) is an instance of $\hat{\theta}_n$.}
Obtaining $\theta_n$ is fast when $n \ll N$;
however, $\theta_N$ is unknown unless we use the entire dataset.
% \barzan{u re now talking about a specific value of ur rand var. either use a diff symbol or adjust the origin def of $\theta_n$}
% Unless we use the entire dataset,
Our key idea is to exploit the asymptotic distribution of $\theta_N - \hat{\theta}_n$
to analytically (thus, efficiently) derive the
conditional distribution of $\hat{\theta}_N \mid \theta_n$, where $\hat{\theta}_n$ is the random variable for $\theta_n$, and $\hat{\theta}_N$ represents our (limited) probabilistic knowledge of $\theta_N$ (\cref{thm:param_dist,thm:cond_param_dist}).
A specific model parameter $\theta_n$ trained on a specific sample (of size $n$) is an instance of $\hat{\theta}_n$
The asymptotic distribution of $\theta_N - \hat{\theta}_n$ is available for the ML methods relying on maximum likelihood estimation.

% relying on maximum likelihood estimation for their parameter optimizations, which includes Generalized Linear Models~\cite{bishop2006pattern} and Probabilistic PCA~\cite{tipping1999probabilistic}.
% \tofix{Although this asymptotic distribution may not be available for all ML models, it has been studied for an important class of models that rely on maximum likelihood estimation for their parameter optimizations.}
	% \tofix{the sampling distribution of $\theta_n$},
% \barzan{this is where u were supposed to say something concrete and impressive instead of just saying sampling distribution! use a tech term}
	% to analytically derive the probability distribution of $\theta_N$ ).
% \barzan{1) for what class of models is this possible?
% 2) what do u mean by derive? u explicitly compute it? isn't it expensive?
% do u mean the joint prob dist of thetaN or just its marginals?}
%

This indicates that, while we cannot determine the exact value of $\theta_N$ without training the full model,
	we can  use
  % the conditional distribution of
  $\hat{\theta}_N \mid \theta_n$
	to probabilistically bound the deviation of $\theta_N$ from  $\theta_n$,
	and consequently, the deviation of $m(\xx; \theta_N)$ from $m(\xx; \theta_n)$ (\cref{sec:accuracy:predict}).
% \barzan{you need an assumption here otherwise sounds like u claim for all ML models including NNs!}
% This also means
Moreover,
we can estimate
the deviation of $m(\xx; \theta_N)$ from $m(\xx; \theta_{n})$
for any other sample size, say $n$, using only the model trained on the initial sample of size $n_0$ (\cref{sec:sample}).
In other words, without having to perform additional training, we can efficiently search for the minimum sample size $n$, with which the approximate model, $m(\xx; \theta_{n})$ would be guaranteed, with probability $1-\delta$, not to deviate from  $m(\xx; \theta_{N})$ by more than $\varepsilon$.

% To efficiently perform the above process,
% % Estimating the deviation between an approximate model and the full model 
% \system employs sampling-based approximation steps, which can incur small errors; however, \system also reflects these errors by training a slightly more conservative model (\cref{lemma:err}).
% \barzan{better wording but really OUT of place. turn into  a footnote. also remember to remove this in the camera ready}

\ph{Difference from Previous Work}
% Sampling techniques have been extensively studied in the machine learning literature.
Existing sampling-based techniques are typically designed
for a very specific type of model, such as
non-uniform sampling for linear regression~\cite{derezinski2017unbiased,feldman2011unified,bachem2017practical,ghashami2014relative,drineas2011faster,drineas2012fast,drineas2006sampling,bhojanapalli2015tighter},
logistic regression~\cite{krishnapuram2005sparse,wang2018optimal},
clustering~\cite{feldman2013turning,jaiswal2014simple},
kernel matrices~\cite{gittens2016revisiting,musco2017recursive},
Gaussian mixture models~\cite{lucic2017training}, and
point processes~\cite{li2015efficient}.
In contrast, \system 
exploits uniform random sampling for training a much wider class of models, 
i.e., any MLE-based model; thus, no sampling probabilities need to be determined in advance.
% \barzan{u didn't mention prev models need preprocessing}
\system's contributions also include an efficient accuracy estimation for the approximate model  and 
 an accurate minimum sample size estimation for satisfying a user-requested accuracy.
 (See \cref{sec:related,sec:conclusion} for discussions.)

\ph{Contributions}
We make the following contributions:
\begin{enumerate}
\item We introduce a system (called \system) that offers
error-computation trade-offs for training
any MLE-based ML model,
% ML models.
% \system supports any ML model that relies on maximum likelihood estimation
% for  its training,
% 	which includes Generalized Linear Models (e.g., linear regression, logistic regression, max entropy classifier, Poisson regression) and Probabilistic Principal Component Analysis.
	including Generalized Linear Models 
	(e.g., linear regression, logistic regression, max entropy classifier, Poisson regression)
	and Probabilistic Principal Component Analysis.
% 	\barzan{only two types of models included in MLE? name more}
% 	\yongjoo{added}
(\cref{sec:overview})

\item
We formally study the sampling distribution of an approximate model's parameters,
	which we use to design an efficient algorithm that computes the probabilistic difference between an
 approximate model and a full one, without having to train the latter.
 (\cref{sec:accuracy})
% To achieve this, we formally \barzan{study the quality of an approximate model's parameter (\cref{sec:accuracy:param}), investigate various approaches for efficiently computing the theoretical expression (\cref{sec:hessian}), and develop a fast Monte Carlo method for estimating the accuracy of an approximate model's predictions (\cref{sec:accuracy:predict} and \cref{sec:sample:fast}).}
% Our algorithm supports
% (\cref{sec:accuracy})
% \tofix{We show that, for the ML methods based on maximum likelihood estimation, the probability that a sample-based model deviates no more than a given error tolerance from a full model can be computed.}

%
\item
We develop a technique that can
analytically infer the quality of a new approximate model,
only using a previous model and without having to train the new one.
 This ability enables \system to automatically and efficiently infer the
appropriate sample size
% \system relies on this contribution
% for automatically and efficiently  inferring the minimum sample size
 for satisfying an error tolerance requested by the user.
 (\cref{sec:sample})

\item We empirically validate the statistical correctness and  computational benefits of
	 \system through extensive experiments. (\cref{sec:exp})
\end{enumerate}

%% file: overview.tex
\section{System Overview}
\label{sec:overview}

In this section, we provide an overview of \system.
We describe \system's user interface in \cref{sec:interface}.
In \cref{sec:support}, we formally describe the models supported by \system.
We describe \system's internal workflow in \cref{sec:workflow}.

\input{figures/fig_interface}

\subsection{User Interface}
\label{sec:interface}

In this section, we first describe the interface of a traditional ML library (e.g., scikit-learn~\cite{scikit-learn}, Weka~\cite{hall09:_weka_data_minin_softw}, MLlib~\cite{mlib}), and then present the difference in \system's interface.
To simplify our presentation, here we focus on classification models;
however, our description can be easily generalized to both regression models (e.g., linear regression) and unsupervised learning (e.g., PPCA),
as described in \cref{sec:model:sim}.

\ph{Traditional ML Libraries}
As depicted in \cref{fig:interface} (top),
% For a traditional ML system,
with a typical ML library,
the user provides a \emph{training set} $D \sim \mathcal{D}$ and specifies
a \emph{model class}
(e.g., linear regression, logistic regression, PPCA)
along with model-specific configurations (e.g., regularization coefficients for linear or logistic regression, the number of factors for PPCA).
A training set $D$ is a (multi-)set of $N$ training examples, which we denote as $\{ (\xx_1, y_1), \ldots, (\xx_N, y_N) \}$.
% , sampled from an unknown distribution $\mathcal{D}$.
The $d$-dimensional vector $\xx_i$ is called a \emph{feature vector}, and a real-valued $y_i$ is called a \emph{label}.
 % The meaning of $y_i$ differs depending on the types of ML tasks, which are described in \cref{tab:training_set}.
Then, the traditional ML library outputs a model $m_N$ trained on the given training set.
We call $m_N$ a \emph{full model}.
% \cref{fig:code:sklearn} is an example of a code snippet that trains a logistic regression classifier using scikit-learn (i.e., the \texttt{\textbf{sklearn}} module)~\cite{scikit-learn}.

% The trained model $m_N$ is used differently depending on the type of learning.
% In supervised learning,
In classification tasks,
$m(\xx)$   predicts a class label for an unseen feature vector $\xx$. For example, if $\xx$ encodes a review of a restaurant, a trained logistic regression classifier $m_N(\xx)$
may predict whether the review is positive or negative.

\ignore{
In unsupervised learning, $m_N()$
 returns values
 % parameters
 that
capture certain characteristics of the training set. For example, if $D$ consists of
 bitmap images of handwritten digits,  $m_N()$  might be
 a factor analysis model (e.g., PPCA)
which returns a small number ($q$) of images that most accurately reconstruct the handwritten digit images when linearly combined~\cite{tipping1999probabilistic};
here, the value of $q$ is the model-specific configuration of PPCA.
% $q$ number \barzan{just the number??} of \tofix{underlying factors (or patterns) that best explains hand-written digits;}
When the meaning is clear, we omit the argument
% parameter
(i.e., $\xx$) of $m_N$ and simply use  $m_N(\cdot)$ or $m_N$.
}

\input{figures/tab_notation}

\ph{\system}
 In addition to the inputs required by traditional ML libraries, \system needs one extra input:
an \emph{approximation contract} that consists of
an error bound $\varepsilon$
and a confidence level $\delta$.
% The approximation contract specifies the quality of the approximate model trained by \system,
% using two constants $\varepsilon$ and $\delta$, which indicates the following.
% \cref{fig:interface} depicts this user interaction in comparison to traditional ML systems.
% The approximation contract consists of an error bound $\varepsilon$ and  a confidence level $\delta$,
% and has the following meaning.
% Given an error bound $\varepsilon$ and a confidence level $\delta$,
% Given $\varepsilon$ and $\delta$,
Then,
\system returns an \emph{approximate model} $m_{n}$ such that
the prediction difference between $m_{n}$ and $m_N$ is within $\varepsilon$ with probability at least $1-\delta$.
That is,
\begin{align*}
&\pr[ v(m_n) \le \varepsilon ] \ge 1 - \delta \\
\text{where} \quad &
v(m_n) = E_{\xx \sim \mathcal{D}} (\1{ m_{n}(\xx) \ne m_N(\xx) })
\end{align*}
% where  $v(m_{n},\, m_N)$ is the \emph{model difference} between $m_{n}$ and $m_N$.
% , a function that defines the difference between the two models.
%
where the expectation is over a test set. 
To estimate the above probability, 
\system uses a holdout set that is not used for training the approximate model.
The approximate model $m_{n}$ is trained on a sample of size $n$, where the value of $n$ is automatically inferred by \system.

\vspace{2mm}
\noindent
The following lemma shows that \system's accuracy guarantee 
also 
implies
 a probabilistic  bound on the full model's generalization error.
\begin{lemma}
  \label{lemma:gen}
  Let $\varepsilon_g$ be the generalization error of \system's approximate model;
  that is, $\varepsilon_g = E_{(\xx,y) \sim \mathcal{D}}(\1{ m_n(\xx) \ne y })$.
  Then, the full model's generalization error is bounded as:
  \[
  E_{(\xx,y) \sim \mathcal{D}} (\1{m_N(\xx) \ne y}) \le
    \varepsilon_g + \varepsilon - \varepsilon_g \cdot \varepsilon
  \]
  with probability at least $1 - \delta$.
\end{lemma}

\noindent
We defer the proof to \cref{sec:proofs}.
We also empirically confirm this result 
in \cref{sec:exp:dim}.

\ignore{
\ph{Model Difference}
The model difference depends on the ML task.
For example, in supervised learning (i.e., classification or regression), the model difference captures the expected difference between the predictions of two models. That is,
\begin{align*}
v(m_1, m_2) &= \E[m_1(\xx) \ne m_2(\xx)] \qquad \text{\bf (classification)}\\
v(m_1, m_2) &= \E[(m_1(\xx) - m_2(\xx))^2] \qquad \text{\bf (regression)}
\end{align*}
For unsupervised learning (e.g., PPCA), the model difference captures the difference between the model parameters.
For instance,
\[
v(m_1, m_2) = 1 - \text{cosine}(\theta_1, \theta_2) \qquad \text{\bf (PPCA)}
\]
where $\text{cosine}(\cdot, \cdot)$ indicates the cosine similarity.
}

% set to zero by default, and $\delta$ indicates the probability that the approximate model produces class label predictions \emph{different from} the full models. For regression tasks and unsupervised learning, $\varepsilon$ is set to 1\% relative error (or something comparable depending on tasks) by default, and $\delta$ indicates the probability that the approximate model's output deviates larger than $\varepsilon$ from the full model.

\subsection{Supported Models \& Abstraction}
\label{sec:support}

Here, we formally describe \system's supported models; then, we describe how \system expresses the supported models in an abstract way.

% \yongjoo{Clean up this section.}

\ph{Formal Description of Supported Models}
\system supports any model that can be trained by solving the following convex optimization problem:
\begin{align}
\argmin_\theta \quad & f_n(\theta)
\label{eq:support:obj1} \\
% \end{align}
% where
% \begin{align}
\text{where} \quad
f_n(\theta) &= \frac{1}{n} \sum_{i=1}^{n}\, \minus \log \pr(\xx_i, y_i;\, \theta) + R(\theta)
\label{eq:support:obj2}
\end{align}
Here, $\pr(\xx_i, y_i;\, \theta)$ indicates the likelihood of observing the pair ($\xx_i$, $y_i$) given $\theta$, $R(\theta)$ is the optional regularization term typically used to prevent overfitting, and
$n$ is the number of training examples used. When $n$=$N$, this results in training the full model $m_N$, and otherwise we have an approximate model $m_n$.
Different ML models use different expressions for $\pr(\xx_i, y_i;\, \theta)$.
\cancut{REMOVED FOR SPACE
For example, PPCA uses the Gaussian distribution with covariance matrix $\xx_i \xx_i^\top$  whereas logistic regression uses
the Bernoulli distribution with its mean being the sigmoid function of $\xx_i^\top \theta$.}
We provide specific examples in \cref{sec:model:abstract:example}.

The solution $\theta_n$ to the minimization problem in \cref{eq:support:obj1}   is a value of $\theta$ at which the gradient $g_n(\theta) = \nabla f_n(\theta)$ of the objective function $f_n(\theta)$ becomes zero.\footnote{
 In this work, we assume $\theta_n$ has fully converged to the optimal point, satisfying \cref{eq:abstract:model}. 
 The non-fully converged cases can   be handled by simply 
 adding small error terms to the diagonal elements of $J$ in \cref{thm:param_dist}.}
\, That is,
% The ML models supported by \system are commonly trained by finding model parameters $\theta_n$ at which the expressions $g_n(\theta)$ as follows becomes a zero vector:
\begin{equation}
g_n(\theta_n) = \left[ \frac{1}{n} \sum_{i=1}^n q(\theta_n; \xx_i, y_i) \right] + r(\theta_n) = \bm{0}
\label{eq:abstract:model}
\end{equation}
where
	$q(\theta; \xx_i, y_i)$ denotes $\minus \nabla_\theta \log \pr(\xx_i, y_i; \theta)$ and $r(\theta)$ denotes $\nabla_\theta R(\theta)$.
% We provide concrete examples of obtaining $g_n(\theta)$ in \cref{sec:model:abstract:example}.

% \input{figures/fig_model_specification}

% \input{figures/tab_models}

\ph{Examples of Supported Models}
\system currently supports the following four model classes: linear regression, logistic regression, max entropy classifier, and PPCA.
However,
 \system's core technical contributions (\cref{thm:param_dist,thm:decreasing}  in \cref{sec:accuracy,sec:sample}) can be generalized to any ML algorithms that rely on maximum likelihood estimation.
\cancut{We are currently focused on adding more configuration options for these currently available models, but we plan to extend our supported model classes in the future (e.g., na\"ive Bayes classifier and decision trees).  Although  neural networks are also quite popular, \system's near-term vision is to focus on better understood models that can still be quite costly when faced with large datasets.}

\ignore{
\textbf{Note} that the users of \system are \emph{not} required to understand or derive any mathematical expressions in order to use (and benefit from)
	the approximation contracts offered by \system.
  }
  %for any of these supported models.

\ignore{
To use a specific model class, the user simply invokes the corresponding class name in \system (see \cref{fig:code:ours:a,fig:code:ours:b} for an example of training a \texttt{LogisticRegression}).

\cref{tab:model_classes} summarizes the currently implemented model classes and their corresponding class names.
\textbf{Note} that the users of \system are \emph{not} required to understand or derive any mathematical expressions in order to use (and benefit from)
	the approximation contracts offered by \system for any of these supported models.
}

\input{figures/fig_archi}

\ph{Model Abstraction}
A model class specification  (\mcs) is the abstraction that
allows \system's components to remain generic and not tied to the specific internal logic of the supported ML models.
Each MCS must implement the following two methods:
\begin{enumerate}
\item \texttt{\textbf{diff($m_1$, $m_2$)}}: This function computes the prediction difference between two models $m_1$ and $m_2$, using part of the training set (i.e., holdout set) that was not used during model training.
 % training either of $m_1$ and $m_2$}.
% \tofix{Examples of $v(m_1, m_2)$ can be found in \cref{sec:interface}.}

% For example, for classification (e.g., logistic regression, max entropy classifier), $v(m_1, m_2) = \E[m_1(\xx) \ne m_2(\xx)]$ for $\xx \sim \mathcal{D}$. For regression (e.g., linear regression), $v(m_1, m_2) = \E[(m_1(\xx) - m_2(\xx))^2]$ for $\xx \sim \mathcal{D}$. For PPCA, $v(m_1, m_2)$ computes one minus the average cosine similarity between two models' respective extracted factors.
%
\item \texttt{\textbf{grads}}: This function returns a list of $q(\theta; \xx_i, y_i) + r(\theta)$ for $i = 1, \ldots, n$, as defined in \cref{eq:abstract:model}.
Although iterative optimization algorithms typically rely \emph{only} on the \emph{average} of this list of values (i.e., the gradient $\nabla_\theta\, f_n(\theta)$), the \texttt{grads} function must return individual values (without averaging them), as they are internally used by \system (see \cref{sec:hessian}).
%
% \item \texttt{\textbf{solve}}: This optional function computes the optimal model parameters for  models that have a closed-form expression.
% (Currently, \system implements the optional \texttt{solve} functions for linear regression and PPCA.)
% , as they both have closed-form solutions.)
\end{enumerate}

\cancut{
\cref{code:model:specification} shows an example of a model class specification.}

\noindent 
  \system already includes the necessary \mcs definitions for the currently supported model classes.
%   (see \cref{sec:support}).

\cancut{Regular users therefore do not need to provide their own \mcs.
However, more advanced developers can easily integrate new ML models (as long as they are
  	based on maximum likelihood estimation) by defining new \mcs.}

% To define a new \mcs in \system, the developers must inherit from the main \texttt{\textbf{ModelClass}} class and
%   implement three virtual functions: \texttt{\textbf{diff()}}, \texttt{\textbf{grads()}}, and optionally, \texttt{\textbf{solve()}}.

\input{workflow}

% \input{architecture}

%% file: figures/fig_interface.tex
\begin{figure}[t]
\centering
\begin{tikzpicture}

\node (U1) at (0,0) {\includegraphics[width=12mm]{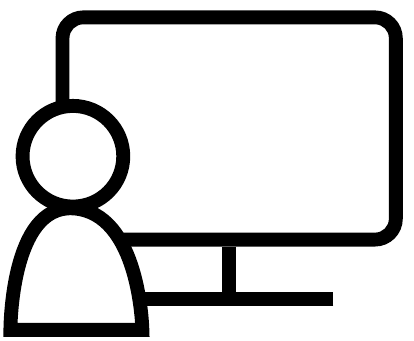}};
\node[draw=vintageblack,ultra thick,fill=none,align=center,rounded corners=1mm,minimum width=16mm,minimum height=13mm,font=\footnotesize] (S1) at ($(U1)+(4.5,0)$) {Traditional\\ML\\Library};
\draw[draw=vintageblack,ultra thick,->] ($(U1.east)+(0,0.2)$) -- ($(S1.west)+(-0.1,0.2)$)
	node[midway,above,font=\it\scriptsize]
	{Training Set};
\draw[draw=vintageblack,ultra thick,<-] ($(U1.east)+(0,-0.2)$) -- ($(S1.west)+(-0.1,-0.2)$)
	node[midway,below,align=center,font=\it\scriptsize]
	{Exact Model\\ (after 2 hours)};
\node[font=\small\bf,align=left,anchor=west] at ($(U1)+(-3,0)$) {Traditional\\ML Library};

\def\y{-2.3}
\node (U2) at (0,\y) {\includegraphics[width=12mm]{figures/app_user.pdf}};
\node[draw=vintageblack,ultra thick,fill=none,align=center,rounded corners=1mm,minimum width=16mm,minimum height=13mm,font=\footnotesize] (S2) at ($(U2)+(4.5,0)$) {\system};
\draw[draw=vintageblack,ultra thick,->] ($(U2.east)+(0,0.2)$) -- ($(S2.west)+(-0.1,0.2)$)
	node[midway,above,align=center,font=\it\scriptsize]
	{Training Set,\\Requested Accuracy (99\%)};
\draw[draw=vintageblack,ultra thick,<-] ($(U2.east)+(0,-0.2)$) -- ($(S2.west)+(-0.1,-0.2)$)
	node[midway,below,align=center,,font=\scriptsize]
	{\textbf{\textit{99\% Accurate Model}}\\ \textbf{\textit{(after 2 mins)}}};
\node[font=\small\bf,align=left,anchor=west] at ($(U2)+(-3,0)$) {\system\\\scriptsize};

% \def\y{-4.6}
% \node (U2) at (0,\y) {\includegraphics[width=12mm]{figures/app_user.pdf}};
% \node[draw=vintageblack,ultra thick,fill=none,align=center,rounded corners=1mm,minimum width=16mm,minimum height=13mm,font=\footnotesize] (S2) at ($(U2)+(4.5,0)$) {\system};
% \draw[draw=vintageblack,ultra thick,->] ($(U2.east)+(0,0.2)$) -- ($(S2.west)+(-0.1,0.2)$)
% 	node[midway,above,align=center,font=\it\scriptsize]
% 	{Training Set,\\Sample Size (10,000)};
% \draw[draw=vintageblack,ultra thick,<-] ($(U2.east)+(0,-0.2)$) -- ($(S2.west)+(-0.1,-0.2)$)
% 	node[midway,below,align=center,,font=\scriptsize]
% 	{\textbf{\textit{90\% Accurate Model}}\\ \textbf{\textit{(after 1 min)}}};
% \node[font=\small\bf,align=left,anchor=west] at ($(U2)+(-3,0)$) {\system\\\scriptsize (\modeb)};

\end{tikzpicture}

\caption{Interaction difference between traditional ML libraries and \system. \system can quickly train an approximate ML model in accordance to a user-specified accuracy request.
% \emph{approximation contract}.
% : \modea or \modeb.
% Latency and accuracy examples are based on our experiment results (\cref{sec:exp}).
}
\label{fig:interface}

\end{figure}

%% file: figures/tab_notation.tex
\begin{table}[t]
  \caption{Notations}
  
  \vspace{-2mm}
  
  \centering
  \small
  \begin{tabular}{l p{70mm}}
    \toprule
    Sym. & Meaning \\
    \midrule
    $N$    & the size of dataset \\
    $n$    & the size of a sample \\
    $D$    & the training set (drawn from a distribution $\mathcal{D}$)\\
    $D_n$  & a size-$n$ random sample of $D$ \\
    % $(\xx, y)$  & a training example \\
    $m_N$  & the full model, which is trained on $D$ \\
    $m_n$  & an approximate model, which is trained on $D_n$ \\
    $\theta_N$ & the parameter of the full model \\
    $\theta_n$ & the parameter of an approximate model \\
    % $v(m_n, m_N)$    & the \emph{difference} between the two models \\
    $v(m_n)$ & the probability that $m_n$ makes a different prediction than $m_N$
    (for the test set) \\
    $\varepsilon$    & the error bound on $v(m_n)$ \\
    $\delta$         & the probability of error bound violation \\
    % $f_n(\theta)$ & an optimization object for training with $D_n$ \\
    % $g_n(\theta)$ & the gradient of $f_n(\theta)$ \\
    $n_0$  & the size of initial training set (10K by default) \\
    $D_0$  & a size-$n_0$ random sample of $D$ \\
    $m_0$  & an initial model trained on $D_0$ \\
    \bottomrule
  \end{tabular}
\end{table}

%% file: figures/fig_archi.tex
\begin{figure}[t]

% \vspace{2mm}

\centering
\begin{tikzpicture}

% COMPONENT
\node[draw=vintageblack,thick,minimum height=10mm,minimum width=15mm,font=\bf\footnotesize,align=center] (MO) at (0,0) {Model\\Trainer};
% \node[draw=none,font=\bf\footnotesize,align=center,minimum height=0mm,minimum width=0mm,anchor=north]
%     at ($(MO.north)+(0,-0.1)$) {Model\\ Trainer};
% \node[draw=vintageblack,minimum height=0mm,minimum width=0mm,anchor=south west,font=\scriptsize] (S)
%     at ($(MO.south west)+(0.12,0.1)$) {\texttt{scipy.optimize}};

\node[draw=vintageblack,thick,minimum height=10mm,minimum width=15mm,align=center,font=\bf\footnotesize,anchor=north west] (MAA)
    at ($(MO.north east)+(0.1,0)$) {Model\\ Accuracy\\ Estimator};

\node[draw=vintageblack,thick,minimum height=10mm,minimum width=15mm,align=center,font=\bf\footnotesize,anchor=north west] (SSE)
    at ($(MAA.north east)+(0.1,0)$) {Sample\\Size\\Estimator};

\node[draw=vintageblack,thick,minimum height=6mm,minimum width=48mm,align=center,font=\bf\footnotesize,anchor=south west] (C)
    at ($(MO.north west)+(0,0.4)$) {\system Coordinator};

% OUTLINE
\node[draw=vintageblack,minimum height=23mm,minimum width=50mm,anchor=north west,anchor=north west] (B)
  at ($(C.north west)+(-0.1,0.1)$) {};

% USER
\node (U2) at ($(B.west)+(-2.2,0.7)$) {\includegraphics[width=12mm]{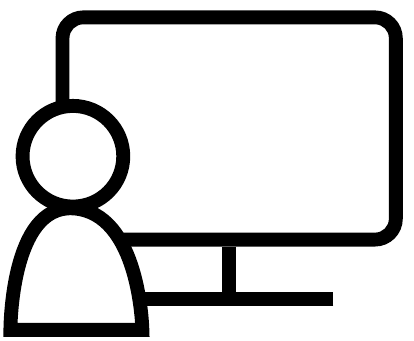}};

\node[anchor=west,draw=vintageblack,rounded corners=1mm,thick,font=\bf\small,minimum height=8mm,minimum width=15mm,align=center] (D)
  at ($(U2.west)+(0.1,-1.2)$) {Training\\ Set};

% FLOW
\draw[{Latex[length=2mm,width=6mm]}-{Latex[length=2mm,width=6mm]},
  line width=1.5mm,vintageblack] ($(C.west)+(-1.6,0)$) -- ($(C.west)+(-0.2,0)$);

\draw[{Latex[length=2mm,width=6mm]}-{Latex[length=2mm,width=6mm]},
  line width=1.5mm,vintageblack] ($(D.east)+(0.1,0.1)$) -- ($(C.west)+(-0.2,-0.3)$);

\draw[{Latex[length=1.5mm,width=5mm]}-{Latex[length=1.5mm,width=5mm]},
  line width=1.5mm,vintageorange] ($(MO.north)+(-0.3,-0.1)$) -- ($(MO.north)+(-0.3,0.5)$);
\draw[{Latex[length=1.5mm,width=5mm]}-{Latex[length=1.5mm,width=5mm]},
  line width=1.5mm,vintageorange] ($(MAA.north)+(-0.3,-0.1)$) -- ($(MAA.north)+(-0.3,0.5)$);
\draw[{Latex[length=1.5mm,width=5mm]}-{Latex[length=1.5mm,width=5mm]},
  line width=1.5mm,vintageorange] ($(SSE.north)+(-0.3,-0.1)$) -- ($(SSE.north)+(-0.3,0.5)$);

\end{tikzpicture}

\vspace{-2mm}
\caption{Architecture of \system.
% All communications are through Coordinator.
% For \modea, orange flows are used.
% For \modeb, green flows are used.
% Computation-intensive operations (in orange) are either processed locally or distributed to multiple workers using Spark, depending on the sizes of workloads.
}
\label{fig:archi}
\end{figure}

%% file: workflow.tex
% \input{figures/fig_archi}

\subsection{System Workflow}
\label{sec:workflow}

We describe the workflow between \system's components depicted in \cref{fig:archi}.
First, Coordinator obtains a size-$n_0$ sample $D_0$ of the training set $D$.
We call $D_0$ the \emph{initial training set} (10K by default).
% \tofix{By default, $n_0$ is set to 10K; this choice kept the initial training fast for all cases we tested while enabling accurate characterization of model-specific properties (which is presented in \cref{sec:accuracy:param}).}
% \tofix{Jingyi: Is it possible to mention why we chose this value as default?}
 Coordinator then invokes Model Trainer to train an \emph{initial model} $m_0$ on $D_0$, and subsequently invokes Model Accuracy Estimator to
 	 estimate the accuracy $\varepsilon_0$  of $m_0$ (with confidence $1-\delta$).
	If $\varepsilon_0$ is smaller than or equal to the user-requested error bound $\varepsilon$,
		Coordinator simply returns the initial model to the user.
Otherwise, Coordinator prepares to train a second model, called the \emph{final model} $m_{n}$. To determine the sample size $n$ required for the final model to satisfy the error
bound, Coordinator consults Sample Size Estimator to estimate the smallest $n$ with which the model difference between $m_{n}$ and $m_N$ (i.e., the \emph{unknown} full model)
	would not exceed $\varepsilon$ with probability at least $1-\delta$.
Note that this operation of Sample Size Estimator does not rely on Model Trainer; that is, no additional (approximate) models are trained for estimating $n$.
Finally, Coordinator invokes Model Trainer (for a second time) to train on a sample of size $n$ and return $m_{n}$ to the user.
Therefore, in the worst case, at most two approximate models are trained.

%% file: accuracy.tex
%!TEX root = approx_ml.tex

\section{Model Accuracy Estimator}
\label{sec:accuracy}

% Model Accuracy Estimator is the component that estimates the accuracy of an approximate model, i.e., the probability of the approximate model making different predictions than the full model.
% This section describes \tofix{this} estimation process.
% To understand Model Accuracy Estimator, it is crucial to understand the statistical properties of \cref{eq:abstract:model} from \cref{sec:model:abstract}.
% In \cref{sec:accuracy:model}, we provide an abstract mathematical form that encapsulates those supported ML models.

Model Accuracy Estimator estimates the accuracy of an approximate model. 
That is,
given an approximate model $m_n$ and a confidence level $\delta$,
Model Accuracy Estimator computes $\varepsilon$, such that $v(m_n) \le \varepsilon$ with probability at least $1 - \delta$.
% The second operations is that given a trained model $m_n$ and an approximation contract function $v(m_1, m_2; \varepsilon)$, it estimates the minimum sample size $n'$ such that $\pr(v(m_{n'}, m_N; \varepsilon) = 1) \ge 1 - \delta$, where $m_{n'}$ is an approximate model trained on a simple random sample of size $n'$.
In \cref{sec:acc:overview}, we first overview the process of Model Accuracy Estimator.
In \cref{sec:accuracy:param}, we establish the statistical properties of
the models supported by \system. Then, in \cref{sec:accuracy:predict}, we explain how \system exploits these statistical properties to estimate the accuracy of an approximate model.
Finally, \cref{sec:hessian} describes how to efficiently compute those statistics.

% In \cref{sec:hessian}, we show how to efficiently compute certain statistics required for expressing these   properties.

\subsection{Accuracy Estimation Overview}
\label{sec:acc:overview}

To compute the probabilistic upper bound $\varepsilon$, Model Accuracy Estimator exploits the fact that both $m_n$ and $m_N$ are essentially the same function (i.e., $m(\xx)$) but with different model parameters
(i.e., $\theta_n$ and $\theta_N$).
Although we cannot compute the exact value of $\theta_N$ without training the full model $m_N$,
	we can still estimate its probability distribution (we explain this in \cref{sec:accuracy:param}).
% Model Accuracy Estimator uses this probability distribution to
% \tofix{obtain}
% % estimate $m_N$'s probability distribution (or more accurately, 
% the probability distribution of the output of $m_N(\xx)$.
The probability distribution of $\theta_N$ is then used to estimate the distribution of $v(m_n)$, which is the quantity we aim to upper bound.
% \barzan{find out what's correct when u use this as a verb, upperbound or upper bound or upper-bound. go through the paper and be consistent} \yongjoo{updated to ``upper bound''}
The upper bound $\varepsilon$ is determined by
	simply finding the value that is larger than $v(m_n)$ 
	for $100 \cdot (1-\delta)$\% of the holdout examples.
% 	with probability at least $1-\delta$ 
	(\cref{sec:accuracy:predict}).

\input{figures/fig_acc_intuition}

\subsection{Model Parameter Distribution}
\label{sec:accuracy:param}

In this section, we present how to probabilistically express the parameter $\theta_N$ of the (unknown) full model $m_N$ given only the parameter $\theta_n$ of an approximate model $m_n$.
% This probabilistic expression is used in \cref{sec:accuracy:predict} to estimate the accuracy of the approximate model itself.
Let $\hat{\theta}_n$ be a random variable  representing the distribution of the approximate model's parameters; $\theta_n$ is simply one instance of $\hat{\theta}_n$. We also use $\hat{\theta}_N$ to represent our (limited) knowledge of $\theta_N$.
Then, our goal is to obtain the distribution of $\theta_n - \hat{\theta}_N$, and then use this distribution to estimate the prediction difference between $m(\theta_n)$ and $m(\hat{\theta}_N)$.

\ph{Intuition}
% Let $\theta_n$ and $\theta_N$  be the parameters of $m_n$ and $m_N$, respectively.
% The parameters  of an approximate model $m_n$ is the optimal value at which \cref{eq:abstract:model} becomes a zero vector. The parameter $\theta_N$ of a full model $m_N$ is the optimal value when $N$ number of training examples are used (instead of $n$).
% Also, 
% derive the conditional distribution of $\hat{\theta}_N \mid \theta_n$.
Since $\theta_n$ is the value
% \barzan{point?} \yongjoo{updated to value} 
that satisfies \cref{eq:abstract:model} for $n$ training examples (instead of $N$), we can obtain the difference $\alpha J$ between $g_n(\theta_n)$ and $g_N(\hat{\theta}_N)$. 
In addition, we can  obtain the relationship $H$ between $g_n(\theta_n) - g_N(\hat{\theta}_N)$ and $\theta_n - \hat{\theta}_N$ using the Taylor expansion of $g_n(\theta)$.
Then, we can finally derive the difference between $\theta_n$ and $\hat{\theta}_N$.

\cref{fig:acc:intuition} depicts this idea intuitively.
In the figure, the slope $H$ 
% \barzan{above u used H for relationship. r u explaining
% fig 3? if so, mention fig 3 again somewhere in this sentence} 
% \yongjoo{added}
captures the surface of the gradient,
and $\alpha J$ captures the variance of the gradient;
% \barzan{not clear what `which' is referring to} \yongjoo{updated}
$\alpha J$ decreases as $n$ increases.
Thus,
if a model is more flexible (e.g., smaller regularization coefficients),
% \barzan{really dangerous: u re saying small reg coef means less complex. 
% thisi s not the real meaning of regularization. reword}
% \yongjoo{updated}
the slope becomes more moderate (i.e., smaller elements in $H$), which
% it\tofix{it??}
% leads to a gentle 
% \barzan{gentle is not an adj for slope! reword} \yongjoo{updated to moderate, but gentle was actually \#1 suggestion}
% slope (i.e., smaller elements in $H$), 
leads to 
a larger distance between $\hat{\theta}_N$ and $\theta_n$  given the same $\alpha J$.
In other words, the approximate model will be less accurate given a fixed sample size.
% In this case, to ensure high accuracy of an approximate model, \system automatically uses a larger sample to reduce $\alpha J$.
Below, we formally present this idea. 
% In
% \cref{sec:exp:model}, 
% \tofix{we also empirically study the tradeoff between the model complexity (i.e., regularization and the number of parameters) and the sample size.}
\yongjoo{removed above because out of place.}
% model complexity
% \barzan{complexity is about num of params! is this waht u meant?} 
% on \system. 
% \barzan{instead of this say we study the complexity-quality tradeoff for a fixed sample size}
To account for these differences between models, \system automatically adjusts its sample size $n$ when it trains an approximate model to satisfy the requested error bound (\cref{sec:sample}).
% \barzan{blah}, \yongjoo{updated} 
% as will be described in \cref{sec:sample}.

% In the following section, this conditional distribution is used to estimate the accuracy of $m_n$.

% In the remainder of this section, we first derive the distribution of $\hat{\theta}_n - \hat{\theta}_N$, and then use this distribution to derive the conditional distribution of $\hat{\theta}_N \mid \theta_n$.

% The parameter $\theta_N$ of the exact model $m_N$ satisfies \cref{} when the entire training set is used.

\ph{Parameter Distribution} 
% \barzan{terrible header name} \yongjoo{updated}
The following theorem provides the distribution of $\hat{\theta}_n - \hat{\theta}_N$
(its proof is in \cref{sec:proofs}).

\begin{theorem}
Let $J$ be the Jacobian of $g_n(\theta) - r(\theta)$ evaluated at $\theta_n$, and let $H$ be the Jacobian of $g_n(\theta)$ evaluated at $\theta_n$. Then,
\[
\hat{\theta}_n \, - \, \hat{\theta}_N \rightarrow
\mathcal{N}( \bm{0},\, \alpha\, H^{\minus 1} J H^{\minus 1} ), \qquad
\alpha = \frac{1}{n} - \frac{1}{N}
% \Sigma_n = \left( \frac{1}{n} \left( 1 - \frac{n}{N} \right) + \frac{1}{N} \right)\,
% \left(\nabla_{\theta\theta}\,f_n(\hat{\theta}_n) \right)^{\minus 1}
\]
as $n \rightarrow \infty$ and $N \rightarrow \infty$. $\mathcal{N}$ denotes a normal distribution,
which means that $\hat{\theta}_n \, - \, \hat{\theta}_N$ asymptotically follows a multivariate normal distribution with covariance matrix $\alpha\, H^{\minus 1} J H^{\minus 1}$.
% \yongjoo{This theorem must say that $N$ also approaches infinity.}
\label{thm:param_dist}
\end{theorem}

Directly computing $H$ and $J$ requires $\Omega(d^2)$ space
% \barzan{BLAH}, \yongjoo{added ``space''}
    where $d$ is the number of features.
    This
    can be prohibitively expensive when $d$ is large. To address this, 
        these quantities
        % \barzan{`these quantities'?} \yongjoo{updated}
        are
        indirectly  computed (as described in \cref{sec:hessian} and \cref{sec:sample:fast}), reducing the computational cost to only $O(d)$.
    Our empirical study shows that \system can scale up to datasets with a million features (\cref{sec:exp}).

In \cref{thm:param_dist}, $J$ is essentially the covariance matrix of gradients (computed on individual examples). Since \system uses uniform random sampling, estimating $J$ is simpler; however, even when non-uniform random sampling is used, $J$ can still be estimated if we know the sampling probabilities. By assigning those sampling probabilities in a task-specific way, one could obtain higher accuracy, which we leave as future work.

% The above theorem is a generalization of the sampling distribution of the maximum likelihood estimator \cite{efron1978assessing,newey1994large}.
% We defer the proof to \cref{sec:proofs}.
%
The following corollary provides   the conditional distribution of $\hat{\theta}_N \mid \theta_n$
(proof in \cref{sec:proofs}).

\begin{mycorollary}
Without any \emph{a priori} knowledge of $\theta_N$,
\[
\hat{\theta}_N \mid \theta_n \rightarrow
\mathcal{N}( \theta_n,\, \alpha\, H^{\minus 1} J H^{\minus 1} ), \qquad
\alpha = \frac{1}{n} - \frac{1}{N}
\]
as $n \rightarrow \infty$ and $N \rightarrow \infty$.
\label{thm:cond_param_dist}
\end{mycorollary}

% In \cref{sec:hessian}, we describe how to efficiently compute $H$ and $J$.
% \yongjoo{Describe how to scale to high-dimensional data.}
% \tofix{\cref{sec:appendix:discussion} discusses the   limitations of extremely high-dimensional data.}
The following section uses the conditional distribution $\hat{\theta}_N \mid \theta_n$ (in \cref{thm:cond_param_dist}) to obtain an error bound on the approximate model.

\subsection{Error Bound on Approximate Model}
\label{sec:accuracy:predict}

In this section, we describe how Model Accuracy Estimator estimates the accuracy of an approximate model $m_n$.
Specifically, we show how to estimate $v(m_n)$ without training $m_N$.
% Using the notations defined in \cref{sec:interface},
% this is equivalent
% to computing $\varepsilon$ such that $E_{\xx \sim \mathcal{D}} ( m_{n}(\xx) \ne m_N(\xx) ) \le \varepsilon$ with probability at least $1-\delta$. For this, we use the conditional distribution $\hat{\theta}_N \mid \theta_n$ derived in the previous section.

Let $h(\theta_N)$ denote the probability density function of the normal distribution with mean $\theta_n$ and covariance matrix $\alpha H^{\minus1} J H^{\minus1}$ (obtained in \cref{thm:cond_param_dist}).
Then, we aim to find the error bound $\varepsilon$ of an approximate model that holds with probability at least $1 - \delta$. That is,
% that satisfies:
\begin{align}
  % p(\varepsilon)
  &\pr_{\theta_N} [ v(m_n) \le \varepsilon ] \ge 1 - \delta \qquad \text{where} \nonumber \\
  % p(\varepsilon)
  % &=
  &\pr_{\theta_N} [ v(m_n) \le \varepsilon ]
  =
  \int \1{ v(m_n; \theta_N) \le \varepsilon}
  \, h(\theta_N)\; d \theta_N, \label{eq:acc:objective} \\
  &m_n(\xx) = m(\xx; \; \theta_n), \quad
  m_N(\xx) = m(\xx; \; \theta_N \mid \theta_n) \nonumber
\end{align}
where
the integration is over the domain of $\theta_N$ ($\in \mathcal{R}^d$);
$v(m_n; \theta_N)$ is the error of $m_n$ when the full model's parameter is
$\theta_N$;
and $\1{\cdot}$ is the indicator function that returns 1 if its argument is true and returns 0 otherwise.
Since the above expression involves the model's (blackbox) prediction function $m(\xx)$,
it cannot be analytically computed in general.

To compute \cref{eq:acc:objective},
 % (which we denote by $p(\varepsilon)$),
\system's Model Accuracy Estimator uses
% approximately computes the integration $p_v$ in \cref{eq:acc:objective} using
the empirical distribution of $h(\theta_N)$ as follows. Let $\theta_{N,1}, \ldots, \theta_{N,k}$ be i.i.d.~samples drawn from $h(\theta_N)$. Then,
\begin{align}
% p(\varepsilon) =
&\int \1{ v(m_n; \theta_N) \le \varepsilon }
\, h(\theta_N)\; d \theta_N
% \label{eq:int}
% \nonumber \\
\approx \frac{1}{k} \sum_{i=1}^k
\1{ v(m_n; \theta_{N,i}) \le \varepsilon }
% = \tilde{p}(\varepsilon)
\label{eq:monte}
\end{align}
% The value of $\tilde{p}(\varepsilon)$ converges
% The quantity in \cref{eq:monte} converges to the quantity in \cref{eq:acc:objective} as $k \rightarrow \infty$; thus, the estimate is \emph{consistent}.
To take into account the approximation error in \cref{eq:monte},
% Note that using this empirical distribution can incur
\yongjoo{updated}
% \barzan{`an additional'?} \yongjoo{this is the only error in this context} error. To account for this potential error,
\system 
uses conservative estimates on $\varepsilon$ as formally stated in the following lemma (see \cref{sec:proofs} for proof).
\begin{lemma}
\label{lemma:err}
If $\varepsilon$ satisfies
  \begin{align*}
  &\frac{1}{k} \sum_{i=1}^k
  \1{ v(m_n; \theta_{N,i}) \le \varepsilon }
    = \frac{1 - \delta}{0.95} + \sqrt{\frac{\log 0.95 }{\minus 2 k}} \\
  \text{then} \quad & \quad
  \pr[v(m_n) \le \varepsilon] \ge 1 - \delta.
  \end{align*}
\end{lemma}

\noindent
The above lemma implies that by using a larger $k$ (i.e., number of sampled values), we can obtain a tighter $\varepsilon$.
To obtain a large   $k$, an efficient sampling algorithm is necessary.
Since $\hat{\theta}_N$ follows a normal distribution,
one can simply use an existing library, such as \texttt{\textbf{numpy.random}}.
% there are already standard libraries, such as \texttt{\textbf{numpy.random}}.
However, \system uses its own fast, custom sampler to avoid directly computing the covariance matrix $H^{\minus1} J H^{\minus1}$ (see \cref{sec:sample:fast}).

\subsection{Computing Necessary Statistics}
\label{sec:hessian}

We present three methods---(1) ClosedForm, (2) InverseGradients, and (3) ObservedFisher---for computing $H$.
Given $H$,
computing $J$ is straightforward since $J = H - J_r$, where $J_r$ is the Jacobian of $r(\theta)$. 
\system uses ObservedFisher by default since it achieves high memory-efficiency by avoiding the direct computations of $H$.

\ph{Method 1: ClosedForm}
ClosedForm uses the analytic form of the Jacobian $H(\theta)$ of $g_n(\theta)$, and sets $\theta = \theta_n$ by the definition of $H$. For instance, $H(\theta)$ of L2-regularized logistic regression is expressed as follows:
\[
H(\theta) = \frac{1}{n} X^\top Q X + \beta I
\]
where $X$ is an $n$-by-$d$ matrix whose $i$-th row is $\xx_i$, and $Q$ is a $d$-by-$d$ diagonal matrix whose $i$-th diagonal entry is $\sigma(\theta^\top \xx_i) (1 - \sigma(\theta^\top \xx_i))$, and $\beta$ is the coefficient of L2 regularization. When $H(\theta)$ is available, as in the case of logistic regression, ClosedForm is fast and exact.

However, inverting $H$ is computationally expensive when $d$ is large. Also, using ClosedForm is less straightforward when obtaining analytic expression of $H(\theta)$ is non-trivial.

% Thus, \system supports ClosedForm only for linear regression and logistic regression. For other methods, such as max entropy classifier and PPCA, either of the following two methods is used.

\ph{Method 2: InverseGradients}
InverseGradients numerically computes $H$ by relying on the Taylor expansion of $g_n(\theta)$: $g_n(\theta_n + d\theta) \approx g_n(\theta_n) + H d\theta$. Since $g_n(\theta_n) = \bm{0}$, the Taylor expansion simplifies to:
\[
g_n(\theta_n + d\theta) \approx H d\theta
\]
The values of $g_n(\theta_n + d\theta)$ and $g_n(\theta_n)$ are computed using the \texttt{grads} function provided by the \mcs. The remaining question is what values of $d\theta$ to use for computing $H$. Since $H$ is a $d$-by-$d$ matrix, \system uses $d$ number of linearly independent $d\theta$ to fully construct $H$. That is, let $P$ be $\epsilon I$, where $\epsilon$ is a small real number ($10^{\minus6}$ by default). Also, let $R$ be the $d$-by-$d$ matrix whose $i$-th column is $g_n(\theta_n + P_{\cdot,i})$ where $P_{\cdot,i}$ is the $i$-th column of $P$. Then,
$H \approx R P^{\minus1}$.
% \[
% R \approx H P
% \quad \Rightarrow \quad
% H \approx R P^{\minus1}
% \]

Since InverseGradients only relies on the \texttt{\textbf{grads}} function, it is applicable to all supported models. Although InverseGradients is accurate, it is still computationally inefficient for high-dimensional data, since the \texttt{\textbf{grads}} function must be called $d$ times. We study its runtime overhead in \cref{sec:exp:hessian}.

\ph{Method 3: ObservedFisher}
ObservedFisher numerically computes $H$ by relying on the information matrix equality~\cite{newey1994large}.\footnote{ObservedFisher is also inspired by Hessian-free optimizations~\cite{lin2008trust,martens2010deep}.}
According to the information matrix equality, the covariance matrix $C$ of $q(\theta_n; \xx_i, y_i)$, for $i = 1, \ldots, n$, is asymptotically identical to $J$ (i.e., as $n \rightarrow \infty$). In addition, given $J$, we can simply obtain $H$
as $H = C + J_r$. Our empirical study in \cref{sec:exp:hessian} shows that ObservedFisher is highly accurate for $n \ge 5K$.

Instead of computing the $d$-by-$d$ matrix $C$ directly, ObservedFisher takes a slightly different approach. That is, ObservedFisher computes \emph{factors} $U$ and $\Sigma$ such that $C = U \, \Sigma^2 \, U^\top$, where $U$ is a $d$-by-$n$ matrix, and $\Sigma$ is an $n$-by-$n$ diagonal matrix. As described below,  this factor-based approach is significantly more efficient when $d$ is large,
hence allowing 
\system to scale up to high-dimensional data.

Specifically,
let $Q$ be the $n$-by-$d$ matrix whose $i$-th row is $q(\theta_n; \xx_i, y_i)$. Then, ObservedFisher performs the singular value decomposition of $Q^\top$ to obtain $U$, $\Sigma$, and $V$ such that $Q^\top = U\, \Sigma \, V^\top$. Then,
the following relationship holds:
% ObservedFisher relies on the following relationship for expressing $J$:
\begin{equation}
C = Q^\top Q = U\, \Sigma^2\, U^\top
% \approx U\, \Sigma_+^2\, U^\top
\label{eq:jacobian}
\end{equation}
As stated above, ObservedFisher never computes $C$;
it only stores $U$ and $\Sigma$, which are used directly for obtaining samples from $\mathcal{N}(\bm{0}, H^{\minus 1} J H^{\minus 1})$ (see \cref{sec:sample:fast}).
% where $\Sigma_+$ is a diagonal matrix with only positive singular values. Note that the diagonal entries of $\Sigma_+^2$ are the eigenvalues of $J$; because they are also all positive, $J$ is positive definite.
The cost of singular value decomposition is $O(\min(n^2 d, n d^2))$. When $d \ge n$, this time complexity becomes $O(n^2 d) = O(d)$ for a fixed sample size ($n_0$ = 10K by default).
Moreover, ObservedFisher requires only a single call of the \texttt{\textbf{grads}} function.
\cref{sec:exp:hessian} empirically studies the relationship between $n$ and ObservedFisher's runtime.

%% file: figures/fig_acc_intuition.tex
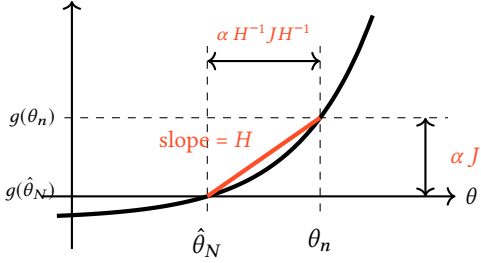
\begin{figure}[t]
\centering

\begin{tikzpicture}
  \draw[->,black,thick] (-3.4,0.2) -- (2.1,0.2)
    node[right,font=\small] {$\theta$};

  \draw[->,black,thick] (-3,-0.5) -- (-3,2.8);
    % node[midway,left,font=\small,xshift=0mm,yshift=11mm] {$f_n(\theta)$};

  \draw[domain=-3.2:1.0,smooth,variable=\x,black,ultra thick]
    plot ({\x}, {0.3 * (exp(\x+1.2) - 1.0) + 0.2});

  % x-axis ticks
  \draw[dashed] (-1.2,-0) -- (-1.2,2.2);
  \node[anchor=north] at (-1.2,-0.1) {$\hat{\theta}_N$};
  \draw[dashed] (0.3,-0) -- (0.3,2.2);
  \node[anchor=north] at (0.3,-0.1) {$\theta_n$};

  % Y-axis ticks
  % \draw[dashed] (-3.1,0.3) -- (2,0.3);
  \node[anchor=east,font=\footnotesize] at (-3.1,0.3) {$g(\hat{\theta}_N)$};

  \draw[dashed] (-3.1,1.246) -- (2,1.246);
  \node[anchor=east,font=\footnotesize] at (-3.1,1.246) {$g(\theta_n)$};

  % diff
  \draw[<->,thick] (1.7,0.2) -- (1.7,1.246)
    node[midway,right,font=\small,xshift=2mm,text=vintageorange]
      {$\alpha\, J$};

  \draw[<->,thick] (-1.2,2.0) -- (0.3,2.0)
    node[midway,above,font=\footnotesize,yshift=1mm,text=vintageorange]
      {$\alpha\, H^{\minus 1} J H^{\minus 1}$};

  \draw[ultra thick,vintageorange] (-1.2,0.2) -- (0.3,1.246)
    node[midway,xshift=0mm,yshift=2mm,anchor=east,font=\small] {slope = $H$};

\end{tikzpicture}

\vspace{-2mm}

\caption{
The variances $\alpha\, H^{-1} J H^{-1}$ of sampling-based model parameters are obtained using two model/data-aware quantities (i.e., $\alpha J$ and $H$).}
\label{fig:acc:intuition}
\end{figure}

%% file: sample.tex
%!TEX root = approx_ml.tex
\section{Sample Size Estimator}
\label{sec:sample}

% Given an error bound $\varepsilon$, a confidence $\delta$, and the initial model $m_0$,

Sample Size Estimator estimates the minimum sample size $n$ such that $E(m_n(\xx) \ne m_N(\xx))$ is not larger than the requested error bound $\varepsilon$ with probability at least $1 - \delta$.
In this process, Sample Size Estimator does not train any additional approximate models; it only relies on the initial model $m_0$ given to this component.

\input{figures/fig_search_intuition}

\subsection{Quality Estimation sans Training}
\label{sec:sample:joint_dist}

This section explains how Sample Size Estimator computes the probability of $E_{\xx} (m_n(\xx) \ne m_N(\xx)) \le \varepsilon$ given the initial model $m_0$.
% This section presents that given the initial model $m_0$, how Sample Size Estimator computes the probability of .
Since both models---$m_{n}(\xx) = m(\xx; \hat{\theta}_{n})$ and $m_N(\xx) = m(\xx; \hat{\theta}_N)$---are uncertain, Sample Size Estimator uses the joint probability distributions of $\hat{\theta}_{n}$ and $\hat{\theta}_N$ to compute the probability.
To make it clear that both models involve uncertain   parameters, we use the following notation:
\[
v(m_n, m_N; \theta_n, \theta_N) =
E_{\xx \sim \mathcal{D}} (\1{m(\xx; \theta_n) \ne m(\xx; \theta_N)})
\]
The computed probability, i.e., $\pr(v(m_n, m_N; \theta_n, \theta_N) \le \varepsilon)$,
is then used in the following section for finding $n$ that makes the probability at least $1-\delta$.

Like Model Accuracy Estimator, Sample Size Estimator computes the probability using the i.i.d.~samples from the joint distribution $h(\theta_{n}, \theta_N)$ of $(\theta_{n}, \theta_N)$ as follows:
\begin{align}
&\pr(v(m_n, m_N; \theta_n, \theta_N) \le \varepsilon) \\
&= \iint \1{v(m_n, m_N; \theta_n, \theta_N) \le \varepsilon} \,
h(\theta_{n}, \theta_N)\, d\theta_{n}\, d\theta_N \nonumber \\
&\approx \frac{1}{k} \sum_{i=1}^k
\1{v(m_n, m_N; \theta_{n,i}, \theta_{N,i}) \le \varepsilon}
% = \tilde{p}_v(n'0)
\label{eq:approx_pv}
\end{align}
where the integration is over the domain of $\theta_n$ and the domain of $\theta_N$, both of which are $\mathcal{R}^d$.
To offset the approximation error in \cref{eq:approx_pv}, \system makes conservative estimates using \cref{lemma:err}.
% \tofix{properly taken into account by \cref{lemma:err}.} 
% \barzan{reword.} \yongjoo{updated}

To obtain i.i.d.~samples, $(\theta_{n',i}, \theta_{N,i})$ for $i = 1, \ldots, k$, from $h(\theta_{n'}, \theta_N)$, Sample Size Estimator uses the following:
\[
\pr(\theta_{n}, \theta_N \mid \theta_0) = \pr(\theta_N \mid \theta_{n})\, \pr(\theta_{n} \mid \theta_0)
\]
where the conditional distributions, $\theta_N \mid \theta_{n}$ and $\theta_{n} \mid \theta_0$, are obtained using \cref{thm:cond_param_dist}. That is, Model Accuracy Estimator uses the following two-stage sampling procedure. It first samples $\theta_{n,i}$ from $\mathcal{N}(\theta_0,\, \alpha_1 H^{\minus1} J H^{\minus1})$ where $\alpha_1 = (1/n_0 - 1/n)$; then, samples $\theta_{N,i}$ from $\mathcal{N}(\theta_{n,i},\, \alpha_2 H^{\minus1} J H^{\minus1})$ where $\alpha_2 = (1/n - 1/N)$. This process is repeated for every $i = 1, \ldots, k$ to obtain $k$ pairs of $(\theta_{n,i}, \theta_{N,i})$.
\cref{fig:search:intuition} depicts this process.
% Finally, $\tilde{p}_v(n)$ is obtained by counting the fraction of the cases in which $\1{v(m({}\cdot{}; \theta_{n',i}), m({}\cdot{}; \theta_{N,i})) \le \varepsilon}$ returns 1.

\subsection{Sample Size Searching}
\label{sec:sample:search}

To find the minimum $n$ such that $\pr(v(m_n, m_N; \theta_n, \theta_N) \le \varepsilon) \ge 1 - \delta$, Sample Size Estimator uses binary search, exploiting that the probability tends to be an increasing function of $n$. We first provide an intuitive explanation; then, we present a formal argument.

Observe that $\pr(v(m_n, m_N; \theta_n, \theta_N) \le \varepsilon)$ relies on the two model parameters $\theta_{n}$ and $\theta_N$. If $\theta_{n} = \theta_N$, the probability is trivially equal to 1. According to \cref{thm:param_dist}, the difference between those two parameters, i.e., $\hat{\theta}_{n} - \hat{\theta}_N$, follows a normal distribution whose covariance matrix shrinks by a factor of $1/n - 1/N$. Therefore, those parameter values become closer as $n \rightarrow N$, which implies that the probability must increase toward 1 as $n \rightarrow N$. The following theorem formally shows that $\pr(v(m_n, m_N; \theta_n, \theta_N) \le \varepsilon)$ is guaranteed to be an increasing function for a large class of cases (its proof is in \cref{sec:proofs}).

\begin{theorem}
Let $h(\theta; \gamma C)$ be the probability density function of a normal distribution with mean $\theta_N$ and covariance matrix $\gamma\, C$, where $\gamma$ is a real number, and $C$ is an arbitrary positive semidefinite matrix. Also, let $B$ be the box area of $\theta$ such that $v(m_n, m_N; \theta_n, \theta_N) \le \varepsilon$.
Then, the following function $p_v(\gamma)$
\[
p_v(\gamma) = \int_B \, h(\theta; \gamma C)\, d\theta
\]
is a decreasing function of $\gamma$.
\label{thm:decreasing}
\end{theorem}

Since binary search is used, Sample Size Estimator needs to compute 
$\frac{1}{k} \sum_{i=1}^k
\1{v(m_n, m_N; \theta_{n,i}, \theta_{N,i}) \le \varepsilon}$
(in \cref{eq:approx_pv})
% $\tilde{p}_v(n)$ 
for different values of $n$; in total, $O(\log_2(N-n_0))$ times. Thus, a fast mechanism for producing i.i.d.~samples is desirable. The following section describes \system's optimizations.

\subsection{Optimizations for Fast Sampling}
\label{sec:sample:fast}

This section describes how to quickly generate i.i.d.~samples from the normal distribution with covariance matrix $(1/n - 1/N)\, H^{\minus1} J H^{\minus1}$.
% Sample Size Estimator's sampling mechanism.
% For computing $\pr(v(m_n, m_N; \theta_n, \theta_N) \le \varepsilon)$, Sample Size Estimator relies on the i.i.d.~samples from the normal distribution with covariance matrix $(1/n' - 1/N)\, H^{\minus1} J H^{\minus1}$.
A basic approach would be to use off-the-shelf functions, such as the one shipped in the \texttt{\textbf{numpy.random}} module, for every different $n$. Albeit simple, this basic approach involves many redundant operations that could be avoided. We describe two orthogonal approaches to reduce the redundancy.
% which enables much faster sampling operations.

\ph{Sampling by Scaling}
We can avoid invoking a sampling function multiple times for different $n$ by exploiting the structural similarity of the covariance matrices associated with different $n$.
Let $\hat{\theta}_{n} \sim \mathcal{N}(\bm{0}, \, (1/n - 1/N)\, H^{\minus1} J H^{\minus1})$, and let $\hat{\theta}_0 \sim \mathcal{N}(\bm{0}, \, H^{\minus1} J H^{\minus1})$. Then, there exists the following relationship:
\[
\hat{\theta}_n = \sqrt{1/n - 1/N} \; \hat{\theta}_0.
\]
This indicates that we can first draw i.i.d.~samples from the unscaled distribution $\mathcal{N}(\bm{0}, H^{\minus1} J H^{\minus1})$; then, we can scale those sampled values by $\sqrt{1/n - 1/N}$ whenever the i.i.d.~samples from $\mathcal{N}(\bm{0}, (1/n - 1/N)\, H^{\minus1} J H^{\minus1})$ are needed.

\ph{Avoiding Direct Covariance Computation}
When $r(\theta) = \beta \theta$ in \cref{eq:abstract:model} (i.e., no regularization or L2 regularization),
Sample Size Estimator avoids the direct computations of $H^{\minus1} J H^{\minus1}$.
% \tofix{which is the key to \system's scalability to high-dimensional training data.}
Instead, it simply draws samples from the standard normal distribution and applies an appropriate linear transformation $L$ to the sampled values ($L$ is obtained shortly).
This approach is used in conjunction with ObservedFisher, 
which is \system's default strategy for computing its necessary statistics (\cref{sec:hessian}).

Avoiding the direct  computation of
the covariances  has two benefits. First, we can completely avoid the $\Omega(d^2)$ cost of computing/storing $H^{\minus1} J H^{\minus1}$.
Second, sampling from the standard normal distribution is much faster
because no dependencies need to be enforced among sampled values.

We use the following relationship:
\[
z \sim \mathcal{N}(\bm{0},\, I)
\quad \Rightarrow \quad
L\, z \sim \mathcal{N}(\bm{0},\, L\, L^\top).
\]
That is, if there exists $L$ such that $L\, L^\top = H^{\minus1} J H^{\minus1}$,  we can obtain the samples of $\hat{\theta}_0$ by multiplying $L$ to the samples drawn from the standard normal distribution.

Specifically, Sample Size Estimator performs the following for obtaining $L$.
Observe from \cref{eq:jacobian} that $J = U\, \Sigma^2\, U^\top$. Since $H = J + \beta$, $H = U\, (\Sigma^2 + \beta I)\, U^\top$. Thus,
\begin{align*}
&H^{\minus1} J H^{\minus1} =
U\, (\Sigma^2 + \beta I)^{\minus1}\, U^\top \;
U\, \Sigma^2\, U^\top \;
U\, (\Sigma^2 + \beta I)^{\minus1}\, U^\top \\
&\Rightarrow
H^{\minus1} J H^{\minus1} = (U \Lambda)\, (U \Lambda)^\top = L \, L^\top
\end{align*}
where $\Lambda$ is a diagonal matrix whose $i$-th diagonal entry is $s_i / (s_i^2 + \beta)$, where $s_i$ is the $i$-th singular value of $J$ contained in $\Sigma$.
Note that both $U$ and $\Sigma$ are already available
as part of computing the necessary statistics in \cref{sec:hessian}.
% \footnote{The concrete description on how to compute the necessary statistics is described in \cref{sec:hessian}.)}
% ObservedFisher is used for computing necessary statistics in \cref{sec:hessian}.
Thus, computing $L$ only involves a simple matrix multiplication.

%% file: figures/fig_search_intuition.tex
\begin{figure}[t]
\centering

\begin{tikzpicture}
  \begin{axis}[
      width=60mm,
      height=45mm,
      view={0}{90},
      xtick=\empty,
      ytick=\empty,
      axis lines=none,
      clip=false,
    ]
    \addplot3[
      samples=50,
      contour gnuplot = {
        labels=false,
        number=5,
        contour label style={
          nodes={text=black},
          /pgf/number format/fixed,
          /pgf/number format/fixed zerofill=true,
          /pgf/number format/precision=1,}},
       contour/draw color={black},
       contour/label distance=5000pt,
       vintageblack,
       opacity=0.6,
       % dashed,
    ]
    {exp(- x^2 + x*y - y^2};

    \node[circle,fill=black,scale=0.5] (A) at (0,0) {};
    \node[anchor=south east,font=\bf] at (A) {$\theta_0$};

    \node[circle,fill=black,scale=0.5] (B) at (-0.8,-0.6) {};
    \node[anchor=north east,font=\bf] at (B) {$\theta_n$};

    \node[circle,fill=black,scale=0.5] (C) at (-0.3,-1.0) {};
    \node[anchor=north west,font=\bf] at (C) {$\theta_N$};

    % lines
    \draw[->,vintageorange,ultra thick] (A) -- (B);
    \draw[->,vintageorange,ultra thick] (B) -- (C);
    
    % second sample
    \node[circle,fill=black,scale=0.5] (E) at (0.6,-0.4) {};
    \node[anchor=north west,font=\bf] at (E) {$\theta_n$};

    \node[circle,fill=black,scale=0.5] (F) at (0.9,0.4) {};
    \node[anchor=south west,font=\bf] at (F) {$\theta_N$};

    % second lines
    \draw[->,vintagegreen,ultra thick,dashed] (A) -- (E);
    \draw[->,vintagegreen,ultra thick,dashed] (E) -- (F);

    % legend
    \draw[->,vintageorange,ultra thick] (axis cs:-2.5,1) -- (axis cs:-2.2,1);
    \node[text=black,anchor=west,font=\small] at (axis cs:-2.16,1) {sample \#1};
    
    \draw[->,vintagegreen,ultra thick,dashed] (axis cs:-2.5,0.5) -- (axis cs:-2.2,0.5);
    \node[text=black,anchor=west,font=\small] at (axis cs:-2.16,0.5) {sample \#2};

  \end{axis}
\end{tikzpicture}

\vspace{-2mm}
\caption{\system repeats this parameter sampling process multiple times to estimate the accuracy of an approximate model $m_n$ (with param $\theta_n$) without having to train it.}
\label{fig:search:intuition}
\end{figure}
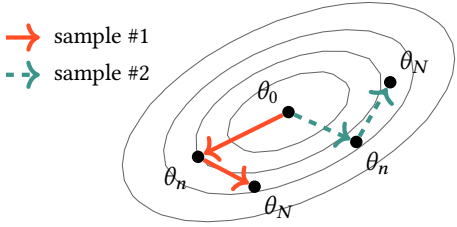

%% file: exp.tex
%!TEX root = approx_ml.tex

\section{Experiments}
\label{sec:exp}

% In addition to our theoretical guarantees in \cref{sec:accuracy,sec:sample},
% 	we also evaluate \system empirically to answer several questions:
% \begin{enumerate}
% \item How faster is approximate training in practice?
% \item Does \system meet the user-requested accuracies?
% \item Is \system's estimated minimum sample size optimal?
% % \item Is \system's estimated model accuracy reliable?
% \item  How large is the runtime overhead of \system?
% \end{enumerate}

%ta inja
% \vspace{1mm}
% \noindent
% Overall,
Our experimental results show the following:
% \yongjoo{Update these numbers.}
\begin{enumerate}

\item \system reduces training time
by 84.04\%--99.84\% (i.e., 6.26$\times$--629$\times$) when training 95\% accurate models.
and by 7.20\%--96.47\% (i.e., 1.07$\times$--28.31$\times$ faster) when training 99\% accurate models.
(\cref{sec:exp:training})
%, we study training time savings for different ML models, datasets, and requested accuracies.

\item The actual accuracy of \system's approximate models is, in most cases, even higher than the requested accuracy. (\cref{sec:exp:guarantee})
%, we compare requested and actual accuracies.

\item \system's estimated minimum sample sizes are close to optimal.
%, we compare \system's estimated minimum sample size to the ones obtained by a (slow) na\"ive approach. We also show that \system's training time is almost optimal;
 % In fact, \system's training time was  only slightly higher than an oracle that had the \emph{perfect knowledge} of the optimal sample size \emph{a priori}.
(\cref{sec:exp:sample})

\item
% \tofix{ dont write like thatu need to keep the diff items parallel}
% \system was scalable to 
\system is  highly effective and accurate 
even for high-dimensional data, and its runtime overhead is much smaller than the time needed for training a full model.
(\cref{sec:exp:dim})

\item
\system's default statistics computation method, ObservedFisher,
is both accurate and efficient.
% the accuracy and efficiency of ObservedFisher 
% compared to other methods (presented in \cref{sec:hessian}).
% \barzan{rephrase to sound like a fact to be parallel to other bullets,
% e,.g. OF is significantly more accurate and mnore eff blah}
% \yongjoo{updated}
(\cref{sec:exp:hessian})

\item 
% \barzan{rephrase to sound like a statement not a report like prev items}
% \yongjoo{updated}
\system offers significant benefits in hyperparameter optimization compared to full model training. (\cref{sec:exp:hyper})

\item
\system's sample size estimation is adaptive to the properties of models. (\cref{sec:exp:model})

% \item The \tofix{estimated} accuracies of \system's approximate models (in \modeb) \tofix{were higher than the actual accuracies of the models.} \barzan{what does `correct' mean??
% say how close they were in percentages or something like that}
% (\cref{sec:exp:assessment})

% \item The runtime overheads of \system was only
% 0.39\%--0.44\%
% % \tofix{1.57 sec--19.0 sec}
% for \modea and
% 0.03\%--0.12\% for \modeb
% compared to the time needed for training a full model.
% % \tofix{0.50 sec--1.64 sec for \modeb}.
%  % for \modeb \tofix{of (already fast) approximate model training time.
%  % For instance, the 22\% overhead was only 0.50 secs.}
%  (\cref{sec:exp:runtime})

% \item
% For interested readers, we  also describe how to compute the statistics used in \cref{sec:accuracy} and empirically study the pros and cons of different approaches
% in \cref{sec:hessian}.

% We also  studied the pros and cons of the two statistics computation methods: InverseGradients and ObservedFisher.
% ()
 \end{enumerate}

% We also  study the pros and cons of InverseGradients and ObservedFisher, the
% \tofix{general statistics computation methods}
% \barzan{awkward phrase. reword plz}
 % described in \cref{sec:exp:hessian}.

\input{figures/tab_dataset}

\input{figures/fig_exp_time_saving}

\subsection{Experiment Setup}
\label{sec:exp:setup}

Here, we present our computational
environment as well as the
 different models and datasets used in our experiments.

\ph{Models}
We tested \system with four different ML models:
% We also present what \texttt{diff} functions were used for them.
\begin{enumerate}
\item \textbf{Linear Regression (Lin).}
Lin is the standard linear regression model with L2 regularization coefficient $\beta$ set as 0.001.
Different values of $\beta$ are tested in \cref{sec:exp:model}.

\item \textbf{Logistic Regression (LR).}
LR is the standard logistic regression (binary) classifier with L2-regularization coefficient $\beta$ set as 0.001.
% LR is a binary classifier. LR is trained by finding parameter values of a sigmoid prediction function such that the misclassification rate is minimized on the training set.
% To avoid overfitting, LR is often trained with regularization.
% We used L2-regularized LR with  coefficient $\beta = 0.001$.
%  Given that LR has no closed-form solution, it is typically trained using an iterative optimization method (e.g., the optimization libraries in \texttt{\textbf{scipy.optimize}}).

% For LR, \system's \texttt{diff} function computes the expected misclassification, i.e., $\varepsilon = \E[m_N(\xx) \ne m_n(\xx)]$, using a holdout set.\footnote{A holdout set is a part of the training set not used for actual model training.}
%
\item \textbf{Max Entropy Classifier (ME).}
ME is the standard max entropy (multiclass) classifier with L2-regularization coefficient $\beta$ set as 0.001.
% ME is a multi-class classifier. ME is trained by finding parameter values of a softmax prediction function such that the misclassification rate is minimized on the training set.
% Similar to LR,
% we used a L2-regularized ME with regularization coefficient $\beta = 0.001$, and since there is no closed-form solution for ME,
% we used the optimization libraries in \texttt{scipy.optimize}.
% an iterative optimization method.
 % For ME, \system's \texttt{diff} function computes the expected misclassification, i.e., $\varepsilon = \E[m_N(\xx) \ne m_n(\xx)]$, using a holdout set.

\item \textbf{PPCA.}
PPCA is the standard probabilistic principal component analysis model~\cite{tipping1999probabilistic}, with the number of factors $q$ set as 10.
% PPCA is a factor-analysis method. PPCA is trained by finding a small number ($q$) of vectors that can most accurately reconstruct the training set  using linear transformations~\cite{tipping1999probabilistic}.
% <<<<<<< HEAD
%   The optimal parameters (i.e., factors) can be found by performing eigendecomposition of the sample covariance matrix \tofix{of} a training set.
%  \barzan{can u come up with a better rewording?}
%   \tofix{For PPCA, we disabled the use of Spark since computing the covariance locally was
%   significantly faster.}
%   \barzan{i realized there was no platform diagram ever in the paper. there was no diagram that shows
%   ok this works on these platforms and may spawn something on Spark and send it back to blah.
%   if u disagree with this, let's talk in person}
% =======
  % The optimal parameters (i.e., factors) can be found by performing eigendecomposition of a sample covariance matrix constructed from the training set.
  % For PPCA, we disabled the use of Apache Spark since computing the covariance locally was
  % significantly faster.
% >>>>>>> c1727491e7e99ecdd7d4ed59b34cf105ec2167d0
% For PPCA, \system's \texttt{diff} function computes one minus the average cosine similarity between extracted factors, i.e., $\varepsilon = 1\, -\, (1/j)\allowbreak \sum_{i=1}^j \text{cosine}(m_N(j), m_n(j))$, where $m_N(j)$ and $m_n(j)$ are the $j$-th factors of respective models.
\end{enumerate}

\vspace{1mm}
\noindent
% \tofix{
% As stated above, \system's automatic adjustments to different regularization parameters is empirically studied in \cref{sec:exp:model}.
\system is configured to use the BFGS optimization algorithm for low-dimensional datasets ($d < 100$) and to use a memory-efficient alternative, called L-BFGS, for high-dimensional datasets ($d \ge 100$).

% As mentioned earlier,  an optimization method must be chosen for  LR and ME.
% Based on our empirical studies, \system
% uses the BFGS optimization when the dataset is low-dimensional ($d < 100$).
% For high-dimensional datasets ($d \ge 100$), \system uses a memory-efficient alternative called L-BFGS.
%Note that \system's other components (i.e., Model Accuracy Assessor and Sample Size Estimator) are agnostic to the choice of an optimization algorithm.

% \barzan{plz go through ur paper and make sure u always have tilde right before \cite{??} to make sure they don't separate between two lines}

\ph{Datasets}
We used six real-world datasets.
The key characteristics of these datasets are summarized in \cref{tab:datasets}.

\vspace{1mm}
\noindent \emph{Linear Regression:}
\begin{enumerate}
  \item \textbf{\gas}:
  This dataset contains chemical sensor readings exposed to gas mixtures at varying concentration levels~\cite{fonollosa2015reservoir}.
  We use the sensor reading as a target variable and gas concentration levels as independent variables.

  \item \textbf{\power}:
  This dataset contains power consumption measurements~\cite{uci_power}. We use the household power consumption as a target variable and global power consumptions as independent variables.
\end{enumerate}

\vspace{1mm}
\noindent \emph{Logistic Regression:}
\begin{enumerate}
  \setcounter{enumi}{2}
\item \textbf{\criteo}:
  This is a click-through rates dataset made publicly available by Criteo Labs~\cite{criteo}.
  The label of each example indicates if an ad was clicked or not.

\item \textbf{\higgs}:
  This is a Higgs bosons simulation dataset~\cite{baldi2014searching}.
  Each example is a pair of physical properties of an environment and a binary indicator of Higgs bosons production.

% \item \textbf{\yelp}:
%   This is a \tofix{collection of} publicly available Yelp reviews~\cite{yelpdata}. Each example is a pair of an English review and a rating (between 0 and 5). For LR,
%   we converted ratings of 0--2 to 0 (negative) and ratings of 3--5 to 1 (positive).
\end{enumerate}

\vspace{1mm}
\noindent \emph{Max Entropy Classifier:}
\begin{enumerate}
  \setcounter{enumi}{4}
\item \textbf{\mnist}:
  This is a hand-written digits image dataset (a.k.a. infinite MNIST~\cite{infmnist}).
  Each example is a pair of an image (intensity value per pixel) and the actual digit in the image.
 % When applying PPCA on \mnist, we used $q$=10.
  % \rev{C12}{the intensity values in $[0,255]$ per pixel of}
  % a larger version  of a hand-written digit dataset called MNIST.

\item \textbf{\yelp}:
%   See above for description.
%   For ME, the original categories (between 0 and 5) were used.
   This is a collection of publicly available Yelp reviews~\cite{yelpdata}. Each example is a pair of an English review and a rating (between 0 and 5).
\end{enumerate}

\vspace{1mm}
\noindent \emph{Probabilistic Principal Component Analysis:}
\begin{enumerate}
  \setcounter{enumi}{6}
  \item \textbf{\mnist}: We use the features of \mnist for PPCA.

  \item \textbf{\higgs}: We use the features of \higgs for PPCA.
\end{enumerate}

\noindent
For each dataset, we used 80\% for training and 20\% for testing.

\ignore{
\begin{enumerate}
\item \textbf{\higgs.}
% We used \higgs for LR.
 % and PPCA ($q=2$).
\item \textbf{\yelp.}
% The original dataset contains 737,530 distinct words; we performed feature selection and used the 1,000 most frequently used words.
% Among 737,530 distinct words, we used the 1,000--100 thousands most frequently-used words to remove mostly-zero columns
%  When applying  LR to \yelp, we formulated a binary classification task by converting  ratings of 0--2 to 0 (negative) and ratings of 3--5 to 1 (positive).

\item \textbf{\mnist.} This is an image dataset (a.k.a. infinite MNIST~\cite{infmnist}), which
contains the bitmap images of hand-written digits.
% a larger version  of a hand-written digit dataset called MNIST.
 Each training example is a pair of a bitmap image and the actual digit in the image.
When applying PPCA on \mnist, we used $q$=10.
\end{enumerate}
}

\ph{Environment}
All of our experiments were conducted on an EC2 cluster with one \texttt{m5.4xlarge} node as a master and five \texttt{m5.2xlarge} nodes as workers.\footnote{\texttt{m5.4xlarge} instances had 16 CPU cores and 64 GB memory. \texttt{m5.2xlarge} instances had 8 CPU cores and 32 GB memory.}
  % EC2 instance (16 CPU cores, 64 GB memory, Ubuntu 16.04). For Spark, we used five additional instances of the \texttt{m5.2xlarge} type (8 CPU cores, 32 GB memory).}
 We used Python 3.6 shipped with Conda \cite{conda} and Apache Spark 2.2 shipped with Cloudera Manager 5.11.

% \cancut{Our latency measurements do not include the time for data (un)loading and the time for distributing data to multiple Spark workers.} \barzan{why? does this make our numbers look better or worse?
% might want to consider removing this sentence to avoid unnecessary questions}

\subsection{Training Time Savings}
\label{sec:exp:training}

This section measures the time savings by \system, compared to training the full model. We used eight model and dataset combinations: (Lin, \gas), (Lin, \power), (LR, \criteo), (LR, \higgs), (ME, \mnist), (ME, \yelp), (PPCA, \mnist), and (PPCA, \higgs).
For Lin, LR and ME, we varied the requested accuracy  $(1-\varepsilon) \times 100\%$  from 80\% to 99\%. For PPCA, we varied the requested accuracy ($(1-\varepsilon) \times 100\%$) from 90\% to 99.99\%. We fixed $\delta$ at 0.05. For each case, we repeated \system's training 20 times.

\input{figures/fig_exp_acc_guarantee}

\cref{fig:exp:time_saving} shows \system's speedups and training time savings in comparison to full model training.
The full model training times were
  345 seconds for (Lin, \gas),
  876 seconds for (Lin, \power),
  5,727 seconds for (LR, \criteo),
  530 seconds for (LR, \higgs),
  35,361 seconds for (ME, \mnist),
  3,048 seconds for (ME, \yelp),
  35 seconds for (PPCA, \mnist),
  and 2 seconds for (PPCA, \higgs).

Two patterns were observed.
First, as expected, \system took relatively longer for training a more accurate approximate model. This is because \system's Sample Size Estimator correctly estimated that a larger sample was needed to satisfy the requested accuracy.
Second, the relative training times were longer for complex models (ME took longer than LR). This is because multi-class classification (by ME) involves more possible class labels; thus, even a small error in parameter values could lead to misclassification, so a larger sample was needed to sufficiently upper bound the chances of a misclassification.
Nevertheless, training 95\% accurate ME models on \mnist still took only 1.53\% of the time needed for training a full model. In other words, even in this case, we observed a 65.27$\times$ speedup, or equivalently, 98.47\% savings in training time.
In all cases, \system's ability to automatically infer appropriate sample sizes and train approximate models on those samples led to significant training time savings. In the subsequent section, we analyze the actual accuracies of those approximate models.

% \input{figures/fig_exp_sss}

% \input{figures/fig_exp_sss_time}

% exp for sample size estimation
\input{figures/fig_exp_baseline}

\subsection{Accuracy Guarantees}
\label{sec:exp:guarantee}

As stated in \cref{eq:approx_pv} in \cref{sec:sample},
the actual accuracy of \system's approximate model is   guaranteed to never be less than the requested accuracy, with probability at least $1-\delta$. In this section, we also empirically validate \system's accuracy guarantees.
Specifically,
	we ran an experiment to compare the requested accuracies and the actual accuracies of the approximate models
		returned by \system.
% \tofix{The actual accuracies were computed using the actual full models}
% \barzan{what does
% this even mean?? ur contradicting ur previous sentence here. makes no sense}
% \tofix{(which were, of course, neither trained nor used for training approximate models).}
% \barzan{again, makes no sense}
 % According to Theorem~\ref{??},

We varied the requested accuracy from $80\%$ to $99\%$ and
	requested a confidence level of 95\%, i.e.,  $\delta=0.05$.
The results are shown in \cref{fig:exp:acc}  for
the same combinations of models and datasets used in the previous section.

% different combinations of
% 	 models and datasets.

In each case, 5th percentile of the actual accuracies was higher than the requested accuracy.
In other words, in 95\% of the cases, the delivered accuracy was higher than the requested one, confirming that \system's probabilistic accuracy guarantees were satisfied.

 Notice that,
 in some cases, e.g.,  (LR, \criteo), (ME, \mnist),
the actual accuracies remained identical even though the requested accuracies were different.
This was due to  \system's design. Recall that \system first trains an initial model $m_0$ and then
	 trains a subsequent model only when the estimated model difference $\varepsilon_0$ of $m_0$ is higher than the requested error $\varepsilon$.
	 In the aforementioned cases, the initial models were already accurate enough; therefore, no additional  models needed to be trained.
	 Consequently, the actual accuracies did not vary in those cases. In other words,
	 the actual accuracies of those initial models were higher than the requested accuracies.

\input{figures/fig_overhead_hyperparam}

\subsection{Sample Size Estimation}
\label{sec:exp:sample}

Sample Size Estimator (SSE) is responsible for
	estimating the minimum sample size, which is a crucial operation in \system.
Too  large a sample eliminates the training time savings; likewise, too small a sample can violate the accuracy guarantees.
In this section, we examine SSE's operations in \system.
We analyze both the accuracy and the efficiency of SSS.

\ph{Accuracy}
To analyze the accuracy of SSS, we   implemented three other
baselines: 
% \barzan{u CONSTANTLY go back and forth between present and past 
% tenses in your expr section. plz make consistent. i wont be able to fix all those
% instances}
FixedRatio, RelativeRatio, and IncEstimator. FixedRatio always used 1\% samples for training approximate models. RelativeRatio used $(1-\varepsilon)*10$\% samples for training approximate models (e.g., 9.5\% sample for 95\% requested accuracy).
IncEstimator gradually increased the sample size until the approximate model trained on that  sample satisfied the requested accuracy; the sample size at $k$-th iteration was $1000 \cdot k^2$. We tested these three baselines and \system on both (Lin, \power) and (LR, \criteo).

\cref{fig:exp:sss:a} shows the results. Since FixedRatio and RelativeRatio set the sample sizes regardless of the model, they either failed to satisfy
the requested accuracies or were overly costly. In contrast, IncEstimator and \system adjusted their sample sizes according to the models and the requested accuracies; hence, they were able to satisfy the requested accuracies. However,   IncEstimator was much more expensive than \system, as described next.

% \barzan{in many places u say both in addition and also in the same sentence.
% search the paper and remove one}
\ph{Efficiency}
To measure the efficiency, we measured the training times of \system and 
all three baselines. To show the overhead of SSS, we also measured \system's training time \emph{excluding} the time spent by SSS (referred to as ``\system's pure training time'').
\cref{fig:exp:sss:b} shows the results. In this figure, the runtimes of IncEstimator were significantly larger than those of \system. For instance, IncEstimator took 5,704 seconds for a 99\% accurate model of (LR, \criteo), while \system took
only 228 seconds (i.e., 25$\times$ faster than
 IncEstimator). Moreover, the runtime overhead of SSS was small enough to keep \system's entire approximate training fast enough.
We also study this overhead more systematically, as reported in the following section.

\input{figures/tab_exp_hessian}

\subsection{Impact of Data Dimension}
\label{sec:exp:dim}

In this section, we analyze the impact of data dimension (i.e., the number of features) on \system's  runtime overhead, generalization error, and  number of optimization iterations.
% In this study, we used (LR, \criteo) and varied the number of features from 1,000 to 100 thousands.

\ph{Runtime Overhead}
To analyze \system's runtime overhead, we separately measured (1) the time taken 
for training an initial model (using a sample of size $n_0$), (2) the time
taken by ObservedFisher for computing statistics,
% \barzan{reword} \yongjoo{updated}
(3) the time taken for sample size estimation, and (4) the time taken for training a final model.
\cref{fig:exp:overhead} shows the results for  (LR, \criteo).
As expected, the times for statistics computations and sample size estimation increased as the number of features increased. However, the overall training time was still significantly smaller than training the full model (e.g., 0.8\% for 100K features).
% In the following section, we empirically study statistics computations in more detail.

\ph{Generalization Error}
\cref{fig:exp:generror} compares the generalization error (or equivalently, test error) of full models and \system's approximate models.
Since \system probabilistically guarantees that its approximate models produce the same predictions as the full models (for the same number of features), the generalization errors of \system's approximate models were highly similar to 
those of the full models.
Moreover, the full model's generalization errors were within the predicted bounds  for more than 95\% of the test cases (as stated in \cref{lemma:gen}).
% As stated in \cref{lemma:gen}, the approximate model's generalization error
% This result confirms \cref{lemma:gen} that we can estimate the full model's generalization error using the generalization error of \system's model.

\ph{Number of Iterations}
Lastly, to better understand the reason behind \system's training time savings, we measured the number of iterations taken by an optimization algorithm (i.e., L-BFGS). 
\cref{fig:exp:itrcount} shows the results. In general, the number of iterations for \system were comparable to those for full model training.
The number  of iterations were similar between the full  training and \system's training, indicating  that  \system's time savings are indeed due to faster gradient computations (because of sampling).

\subsection{Statistics Computation}
\label{sec:exp:hessian}

% In \cref{sec:hessian}, we presented two general
% methods---InverseGradients and ObservedFisher---for computing important statistics, i.e., $H$ and $J$.

In this section, we compare ClosedForm, InverseGradients, and ObservedFisher in terms of both accuracy and efficiency.
 This analysis serves as an empirical justification for why ObservedFisher is \system's default choice.

\ph{Accuracy}
% To compare the accuracy of the three methods, we using (Lin, \power) for different sample sizes. 
To compare the accuracy, 
we first estimated the variances of the parameters using each of ClosedForm, InverseGradients, and ObservedFisher.
% \barzan{reword} \yongjoo{updated}
Then, we computed the ratio between those estimated variances and the actual variances.
% \barzan{incomprehensible} \yongjoo{updated}
Thus, a ratio close to 1 would indicate high accuracy.
% It is desirable that the estimates are conservative (ratio being not smaller than 1.0) while being close to the true values (ratio being close to 1.0).
\cref{sec:exp:statistics} shows the results
for (Lin, \power). 
In general, 
for small samples ($n \le $ 1000),
ObservedFisher was relatively inaccurate in comparison to other methods.
% \barzan{what??} \yongjoo{updated}
However, as the sample size increased, the accuracy of all methods 
% increased 
improved (i.e., the ratio approached 1.0); further, 
the accuracy of ObservedFisher was comparable to other methods.
% \barzan{u just said the opposite in prev 2 sentences}
% \yongjoo{updated the earlier sentences}
Recall that ClosedForm is applicable to 
% \barzan{applicable to?} \yongjoo{updated} 
certain types of models (e.g., Lin and LR) while InverseGradients and ObservedFisher are applicable to
% \barzan{applicable to?} 
all MLE-based models.
% \barzan{all MLE-based models?} \yongjoo{updated}
Overall, the estimated variances of the parameters were larger than the actual variances, showing that \system's probabilistic guarantees were met.
Below, we compare the runtime overhead of InverseGradients and ObservedFisher.
% \barzan{not english} \yongjoo{changed}

\ph{Efficiency}
For an in-depth comparison of InverseGradients and ObservedFisher,
we used two more combinations (LR, \higgs) and (ME, \mnist). Recall from \cref{tab:datasets} that \higgs is a low-dimensional dataset ($d = 28$) while \mnist is high-dimensional ($d = 784$).
We used each method to compute the covariance matrix $H^{\minus 1} J H^{\minus 1}$, which determines the parameter distribution in \cref{thm:param_dist}  (\cref{sec:accuracy}).
We measured the runtime of each method and calculated the accuracy of its estimated covariance matrix.
To measure the accuracy, we calculated the average Frobenius norm, i.e., $(1/d^2)\, \|C_t - C_e\|_F$,
 where $C_t$ was the true covariance matrix and $C_e$ was the estimated covariance matrix.

\cref{tab:exp:hessian} summarizes the results of this experiment. For the low-dimensional data (\higgs), the runtimes and accuracies of the two methods were comparable. 
While their accuracies remained comparable  for the high-dimensional data (\mnist),
 their performance differed. Since InverseGradients had to invoke the \texttt{grads} function repeatedly (i.e., $d$ times), its runtime  increased drastically.
In contrast, ObservedFisher had to call the \texttt{grads} function only once.

\subsection{Hyperparameter Optimization}
\label{sec:exp:hyper}

This section studies \system's overall benefit for performing  hyperparameter optimization.
Specifically, we compared \system to a traditional approach (i.e., full model training)
% \barzan{what is it called? citation?}
in searching for an optimal combination of a feature set and a hyperparameter.
% Specifically, we compared the accuracies of the models (with different feature sets and hyperparameters) trained by \system and a traditional ML system.
As performed by the Random Search hyperparameter optimization method~\cite{bergstra2011algorithms},
we first generated a sequence of (pairs of) a randomly chosen feature set and a regularization coefficient.
Then, we let \system train a series of 95\% accurate models using the feature set and the regularization coefficient in the sequence.
Similarly, we let the traditional approach train a series of exact models using the same combination of the feature set and the regularization coefficient.

\input{figures/fig_hyperparam}

\cref{fig:exp:hyperparam} shows the result, where each dot  represents a model. 
Within half an hour, \system trained 961 models while the traditional approach was able to train only 3.
Both \system and the traditional approach found the second-best model (with test accuracy 74\%) at the second iteration (since they used the same sequence); however, \system took only 1.03 seconds while the traditional approach took 817 seconds.
After 387 seconds (about 6 mins), \system found the best model (with test accuracy 75\%; found at iteration \#91), while the traditional approach could not find the model in an hour.
The sizes of the samples used by \system varied between 10,000 and 9,211,426, depending on the feature set and the regularization coefficient. Note that this is  expected, since \system automatically chooses the smallest sample size that is still large enough to train an accurate model. 
We study \system's sample size estimation in more detail
in the following section.

\input{figures/fig_exp_model_complexity}

\subsection{Impact of Model Complexity on Estimated Sample Sizes}
\label{sec:exp:model}

We also study the impact of model complexity (i.e., regularization coefficients and the number of parameters) on \system's estimated sample size.
Intuitively, if a model is more complex (i.e., smaller regularization coefficients or more parameters),
a larger sample would be needed to satisfy a user-requested error bound.

\cref{fig:exp:model} shows the results.
The left subfigure (\cref{fig:exp:model:a}) shows that as the regularization coefficient increased, the estimated sample size decreased.
The right subfigure (\cref{fig:exp:model:b}) shows that as the model had a larger number of parameters, the estimated sample size increased.
Both observations are consistent with our intuition  (\cref{sec:accuracy}).

%% file: figures/tab_dataset.tex
\begin{table}[t]
  \caption{Datasets used in our experiments}
  \label{tab:datasets}

\vspace{-2mm}

\centering
\small
\begin{tabular}{l r r r}
\toprule
\textbf{Dataset} & \textbf{\# of Rows} ($N$) & \textbf{Dimension} ($d$) & \textbf{Size} \\
\midrule
\gas    & 4,178,504   & 57        & 1.9 GB \\
\power  & 2,075,259   & 114       & 1.8 GB \\
\criteo & 45,840,616  & 998,922   & 2.86 GB \\
\higgs  & 11,000,000  & 28        & 7.5 GB \\
\mnist  & 8,000,000   & 784       & 47.5 GB \\
\yelp   & 5,261,667   & 100,000   & 487 MB \\
\bottomrule
\end{tabular}
\end{table}

%% file: figures/fig_exp_time_saving.tex
\begin{figure*}[t]

\pgfplotsset{timesaving/.style={
        width=45mm,
        height=30mm,
        xmin=0.5,
        xmax=9.5,
        ymin=1,
        ymax=100000,
        xlabel=Requested Accuracy ($(1-\varepsilon) \times 100$\%),
        xlabel near ticks,
        ylabel near ticks,
        ylabel style={align=center},
        xtick={1, 2, 3, 4, 5, 6, 7, 8, 9},
        xticklabels={80\%, 85\%, 90\%, 95\%, 96\%, 97\%, 98\%, 99\%, 100\%},
        xticklabel style={rotate=60,yshift=-0.5mm,anchor=east,xshift=0mm},
        ytick={1, 10, 100, 1000, 10000, 100000},
        yticklabels={1$\times$, 10$\times$, 100$\times$, $10^3\times$, $10^4\times$, $10^5\times$},
        every y tick label/.append style={color=vintageorange},
        % yticklabels={0\%, 20\%, 40\%, 60\%, 80\%, 100\%},
        % yticklabels={0 hr, 0.5 hr, 1 hr, 1.5 hr, 2.0 hr},
        % extra y ticks={10, 20, 30, 40},
        % extra y tick style={yticklabels={}},
        ylabel shift=-7pt,
        xlabel shift=-2pt,
        legend style={
            at={(0,1.2)},anchor=south west,column sep=2pt,
            draw=black,fill=white,line width=.5pt,
            /tikz/every even column/.append style={column sep=20pt},
            font=\small,
        },
        legend columns=3,
        legend cell align={left},
        every axis/.append style={font=\footnotesize},
        ymajorgrids=false,
        % yminorgrids,
        minor grid style=lightgray,
        % nodes near coords,
        % every node near coord/.append style={font=\scriptsize,anchor=west,rotate=20}
    }}

% \centering
\begin{subfigure}[b]{0.24\textwidth}
\centering
\begin{tikzpicture}

\begin{axis}[timesaving,
% ybar,bar width=1.5mm,
axis y line*=right,
ylabel shift=-7pt,
axis x line=none,
ymin=0,
ymax=100,
ytick={0, 20, 40, 60, 80, 100},
yticklabels={0\%, 20\%, 40\%, 60\%, 80\%, 100\%},
every y tick label/.append style={color=vintagegreen},
]

\pgfplotsset{
  /pgfplots/mylegend/.style={
    legend image code/.code={
      \draw [vintageorange,ultra thick] (0cm,0cm) -- (0.6cm,0cm);
      \node [circle,minimum size=0.2pt,scale=0.8,fill=vintageorange] at (0.17cm,0) {};
    },
  },
}

% \addplot[sharp plot,mark=*,mark size=1.5,mark options={fill=vintageorange},
% ultra thick,draw=vintageorange]
% table[x=x,y=y] {
% x y
% % 1  167.8  167.8$\times$
% -1  0
% -2  0
% };
\addlegendimage{mylegend}
\addlegendentry{Speedup (left Y-axis)}

\addplot[
mark=triangle*,mark size=2.5,mark options={fill=vintagegreen,solid},draw=vintagegreen,opacity=1,
ultra thick,dashed]
table[x=x,y=y] {
x y
1  99.92
2  99.93
3  99.90
4  99.84
5  99.66
6  99.42
7  99.04
8  96.47
9  1
};

\addlegendentry{Time Saving (right Y-axis)}
\end{axis}

\begin{axis}[timesaving,
axis y line*=left,
ymode=log,
ymajorgrids=false,
ymax=10000,
]
\addplot[mark=*,mark size=1.5,mark options={fill=vintageorange},
ultra thick,draw=vintageorange,
]
table[x=x,y=y] {
x y
1  1326.9231
2  1363.6364
3  1023.7389
4  629.3324
5  291.1638
6  171.0885
7  104.2324
8  28.3132
9  1
};
\end{axis}
\end{tikzpicture}

\vspace{-2mm}
\caption{Lin, \gas}
\end{subfigure}
~
\begin{subfigure}[b]{0.24\textwidth}
\centering
\begin{tikzpicture}

\begin{axis}[timesaving,
% ybar,bar width=1.5mm,
axis y line*=right,
axis x line=none,
ymin=0,
ymax=100,
ytick={0, 20, 40, 60, 80, 100},
yticklabels={0\%, 20\%, 40\%, 60\%, 80\%, 100\%},
every y tick label/.append style={color=vintagegreen},
]

\pgfplotsset{
  /pgfplots/mylegend/.style={
    legend image code/.code={
      \draw [vintageorange,ultra thick] (0cm,0cm) -- (0.6cm,0cm);
      \node [circle,minimum size=0.2pt,scale=0.8,fill=vintageorange] at (0.17cm,0) {};
    },
  },
}

% \addplot[sharp plot,mark=*,mark size=1.5,mark options={fill=vintageorange},
% ultra thick,draw=vintageorange]
% table[x=x,y=y] {
% x y
% % 1  167.8  167.8$\times$
% -1  0
% -2  0
% };

\addplot[
mark=triangle*,mark size=2.5,mark options={fill=vintagegreen,solid},draw=vintagegreen,opacity=1,
ultra thick,dashed]
table[x=x,y=y] {
x y
1  99.84
2  99.84
3  99.84
4  98.62
5  98.49
6  98.33
7  97.98
8  96.03
9  0
};

% \addplot[mark=*,mark size=1.5,mark options={fill=vintagegreen},
% ultra thick,draw=vintagegreen]
% table[x=x,y=y] {
% x y
% % 1  367.3  367.3$\times$
% 2  99.73
% 3  96.68
% 4  92.51
% 5  65.84
% 6  33.42
% };

% \addlegendentry{\higgs}
% \addlegendentry{\yelp (binarized)}

% \draw[vintageorange,thick] (axis cs:0.5,6060) -- (axis cs:5.5,6060);
\end{axis}

\begin{axis}[timesaving,
axis y line*=left,
ymode=log,
ymajorgrids=false,
ymax=1000,
]
\addplot[mark=*,mark size=1.5,mark options={fill=vintageorange},
ultra thick,draw=vintageorange]
table[x=x,y=y] {
x y
1  626.5864
2  637.0412
3  633.5177
4  72.5580
5  66.3077
6  60.0000
7  49.5973
8  25.1670
9  1
};
\end{axis}
\end{tikzpicture}

\vspace{-2mm}
\caption{LR, \criteo}
\end{subfigure}
~
\begin{subfigure}[b]{0.24\textwidth}
\centering
\begin{tikzpicture}
\begin{axis}[timesaving,
% ybar,bar width=1.5mm,
axis y line*=right,
axis x line=none,
ymin=0,
ymax=100,
ytick={0, 20, 40, 60, 80, 100},
yticklabels={0\%, 20\%, 40\%, 60\%, 80\%, 100\%},
ymajorgrids=false,
every y tick label/.append style={color=vintagegreen},
]

% \addplot[mark=*,mark size=1.5,mark options={fill=vintageorange},
% ultra thick,draw=vintageorange]
% table[x=x,y=y] {
% x y
% % 1  16809.4  16809.4$\times$99.99\%
% 2  99.99
% 3  99.99
% 4  99.95
% 5  97.57
% 6  91.98
% };

\addplot[
mark=triangle*,mark size=2.5,mark options={fill=vintagegreen,solid},draw=vintagegreen,opacity=1,
ultra thick,dashed]
table[x=x,y=y] {
x y
1  99.84
2  99.84
3  99.49
4  98.47
5  97.69
6  96.15
7  91.70
8  75.03
9  0
};
\end{axis}

\begin{axis}[timesaving,
axis y line*=left,
ylabel shift=-8pt,
ymode=log,
ymax=1000,
xlabel=Requested Accuracy ($(1-\varepsilon) \times 100\%$),
ymajorgrids=false,
]
\addplot[mark=*,mark size=1.5,mark options={fill=vintageorange},
ultra thick,draw=vintageorange]
table[x=x,y=y] {
x y
1  634.733441
2  639.5550733
3  194.2912088
4  65.26698536
5  43.31810219
6  25.97170809
7  12.04709682
8  4.005477925
9  1
};
\end{axis}
\end{tikzpicture}

\vspace{-2mm}
\caption{ME, \mnist}
\end{subfigure}
~
\begin{subfigure}[b]{0.24\textwidth}
\centering
\begin{tikzpicture}

\begin{axis}[timesaving,
axis y line*=right,
axis x line=none,
% ybar,bar width=1.5mm,
xmin=0.5,
xmax=8.5,
xtick={1, 2, 3, 4, 5, 6, 7, 8},
xticklabels={90\%, 95\%, 99\%, 99.5\%, 99.9\%, 99.95\%, 99.99\%, 100\%},
% xlabel shift=0mm,
xlabel=Requested Accuracy ($1-\varepsilon$),
ymin=0,
ymax=100,
ytick={0, 20, 40, 60, 80, 100},
yticklabels={0\%, 20\%, 40\%, 60\%, 80\%, 100\%},
every y tick label/.append style={color=vintagegreen},
]

\addplot[
mark=triangle*,mark size=2.5,mark options={fill=vintagegreen,solid},draw=vintagegreen,opacity=1,
ultra thick,dashed]
table[x=x,y=y] {
x y
1  93.71
2  93.60
3  93.17
4  92.43
5  87.46
6  82.00
7  58.46
8  0
};
% \addlegendentry{\higgs}
% \addlegendentry{\mnist}
\end{axis}

\begin{axis}[timesaving,
axis y line*=left,
ylabel shift=-8pt,
ymode=log,
xmin=0.5,
xmax=8.5,
ymax=100,
xtick={1, 2, 3, 4, 5, 6, 7, 8},
xticklabels={90\%, 95\%, 99\%, 99.5\%, 99.9\%, 99.95\%, 99.99\%, 100\%},
xlabel shift=-2mm,
ymajorgrids=false,
]
\addplot[mark=*,mark size=1.5,mark options={fill=vintageorange},
ultra thick,draw=vintageorange]
table[x=x,y=y] {
x y
1  15.9091
2  15.6250
3  14.6444
4  13.2075
5  7.9727
6  5.5556
7  2.4072
8  1
};
\end{axis}
\end{tikzpicture}

\vspace{-2mm}
\caption{PPCA, \mnist}
\end{subfigure}

%%%%%%%%%%%%%%
% SECOND ROW
%%%%%%%%%%%%%%

\vspace{2mm}

\begin{subfigure}[b]{0.24\textwidth}
\centering
\begin{tikzpicture}

\begin{axis}[timesaving,
% ybar,bar width=1.5mm,
axis y line*=right,
ylabel shift=-7pt,
axis x line=none,
ymin=0,
ymax=100,
ytick={0, 20, 40, 60, 80, 100},
yticklabels={0\%, 20\%, 40\%, 60\%, 80\%, 100\%},
every y tick label/.append style={color=vintagegreen},
]

\pgfplotsset{
  /pgfplots/mylegend/.style={
    legend image code/.code={
      \draw [vintageorange,ultra thick] (0cm,0cm) -- (0.6cm,0cm);
      \node [circle,minimum size=0.2pt,scale=0.8,fill=vintageorange] at (0.17cm,0) {};
    },
  },
}

\addplot[
mark=triangle*,mark size=2.5,mark options={fill=vintagegreen,solid},draw=vintagegreen,opacity=1,
ultra thick,dashed]
table[x=x,y=y] {
x y z
1  99.85
2  99.84
3  99.84
4  99.24
5  98.88
6  97.39
7  94.56
8  81.55
9  1
};

\end{axis}

\begin{axis}[timesaving,
axis y line*=left,
ymode=log,
ymajorgrids=false,
ymax=1000,
]
\addplot[mark=*,mark size=1.5,mark options={fill=vintageorange},
ultra thick,draw=vintageorange]
table[x=x,y=y] {
x y
1  683.8461
2  630.1252
3  628.1371
4  131.3639
5  89.0878
6  38.3336
7  18.3902
8  5.4188
9  1
};
\end{axis}
\end{tikzpicture}

\vspace{-2mm}
\caption{Lin, \power}
\end{subfigure}
~
\begin{subfigure}[b]{0.24\textwidth}
\centering
\begin{tikzpicture}

\begin{axis}[timesaving,
% ybar,bar width=1.5mm,
axis y line*=right,
axis x line=none,
ymin=0,
ymax=100,
ytick={0, 20, 40, 60, 80, 100},
yticklabels={0\%, 20\%, 40\%, 60\%, 80\%, 100\%},
every y tick label/.append style={color=vintagegreen},
]

\pgfplotsset{
  /pgfplots/mylegend/.style={
    legend image code/.code={
      \draw [vintageorange,ultra thick] (0cm,0cm) -- (0.6cm,0cm);
      \node [circle,minimum size=0.2pt,scale=0.8,fill=vintageorange] at (0.17cm,0) {};
    },
  },
}

% \addplot[sharp plot,mark=*,mark size=1.5,mark options={fill=vintageorange},
% ultra thick,draw=vintageorange]
% table[x=x,y=y] {
% x y
% % 1  167.8  167.8$\times$
% -1  0
% -2  0
% };

\addplot[
mark=triangle*,mark size=2.5,mark options={fill=vintagegreen,solid},draw=vintagegreen,opacity=1,
ultra thick,dashed]
table[x=x,y=y] {
x y
1  99.93
2  99.93
3  99.64
4  99.17
5  98.26
6  96.25
7  89.93
8  56.40
9  1
};

% \addplot[mark=*,mark size=1.5,mark options={fill=vintagegreen},
% ultra thick,draw=vintagegreen]
% table[x=x,y=y] {
% x y
% % 1  367.3  367.3$\times$
% 2  99.73
% 3  96.68
% 4  92.51
% 5  65.84
% 6  33.42
% };

% \addlegendentry{\higgs}
% \addlegendentry{\yelp (binarized)}

% \draw[vintageorange,thick] (axis cs:0.5,6060) -- (axis cs:5.5,6060);
\end{axis}

\begin{axis}[timesaving,
axis y line*=left,
ymode=log,
ymajorgrids=false,
ymax=10000,
]
\addplot[mark=*,mark size=1.5,mark options={fill=vintageorange},
ultra thick,draw=vintageorange]
table[x=x,y=y] {
x y
1  1487.2296
2  1490.1873
3  276.7590
4  120.6080
5  57.5334
6  26.6467
7  9.9280
8  2.2936
9  1
};
\end{axis}
\end{tikzpicture}

\vspace{-2mm}
\caption{LR, \higgs}
\end{subfigure}
~
\begin{subfigure}[b]{0.24\textwidth}
\centering
\begin{tikzpicture}
\begin{axis}[timesaving,
% ybar,bar width=1.5mm,
axis y line*=right,
axis x line=none,
ymin=0,
ymax=100,
ytick={0, 20, 40, 60, 80, 100},
yticklabels={0\%, 20\%, 40\%, 60\%, 80\%, 100\%},
ymajorgrids=false,
every y tick label/.append style={color=vintagegreen},
]

\addplot[
mark=triangle*,mark size=2.5,mark options={fill=vintagegreen,solid},draw=vintagegreen,opacity=1,
ultra thick,dashed]
table[x=x,y=y] {
x y
1  94.84
2  94.30
3  92.80
4  84.04
5  76.30
6  63.18
7  36.42
8  7.20
9  0
};

\end{axis}

\begin{axis}[timesaving,
axis y line*=left,
ylabel shift=-8pt,
ymode=log,
ymax=100,
xlabel=Requested Accuracy ($(1-\varepsilon) \times 100\%$),
ymajorgrids=false,
]
\addplot[mark=*,mark size=1.5,mark options={fill=vintageorange},
ultra thick,draw=vintageorange]
table[x=x,y=y] {
x y
1  19.3856
2  17.5394
3  13.8943
4  6.2662
5  4.2189
6  2.7158
7  1.5729
8  1.0776
9  1
};
\end{axis}
\end{tikzpicture}

\vspace{-2mm}
\caption{ME, \yelp}
\end{subfigure}
~
\begin{subfigure}[b]{0.24\textwidth}
\centering
\begin{tikzpicture}

\begin{axis}[timesaving,
axis y line*=right,
axis x line=none,
% ybar,bar width=1.5mm,
xmin=0.5,
xmax=8.5,
xtick={1, 2, 3, 4, 5, 6, 7, 8},
xticklabels={90\%, 95\%, 99\%, 99.5\%, 99.9\%, 99.95\%, 99.99\%, 100\%},
% xlabel shift=0mm,
xlabel=Requested Accuracy ($1-\varepsilon$),
ymin=0,
ymax=100,
ytick={0, 20, 40, 60, 80, 100},
yticklabels={0\%, 20\%, 40\%, 60\%, 80\%, 100\%},
every y tick label/.append style={color=vintagegreen},
]

\addplot[
mark=triangle*,mark size=2.5,mark options={fill=vintagegreen,solid},draw=vintagegreen,opacity=1,
ultra thick,dashed]
table[x=x,y=y] {
x y
1  92.7500
2  92.8000
3  92.7500
4  92.5000
5  91.6500
6  90.5000
7  80.6000
8  0
};
\end{axis}

\begin{axis}[timesaving,
axis y line*=left,
ylabel shift=-8pt,
ymode=log,
xmin=0.5,
xmax=8.5,
ymax=100,
xtick={1, 2, 3, 4, 5, 6, 7, 8},
xticklabels={90\%, 95\%, 99\%, 99.5\%, 99.9\%, 99.95\%, 99.99\%, 100\%},
xlabel shift=-2mm,
ymajorgrids=false,
]
\addplot[mark=*,mark size=1.5,mark options={fill=vintageorange},
ultra thick,draw=vintageorange]
table[x=x,y=y] {
x y
1  13.7931
2  13.8889
3  13.7931
4  13.3333
5  11.9760
6  10.5263
7  5.1546
8  1
};
\end{axis}
\end{tikzpicture}

\vspace{-2mm}
\caption{PPCA, \higgs}
\end{subfigure}

\vspace{-2mm}
\caption{\system's speedups compared to full model training (the accuracies of the approximate models are studied in \cref{fig:exp:acc}).}
\label{fig:exp:time_saving}
% \vspace{2mm}
\end{figure*}
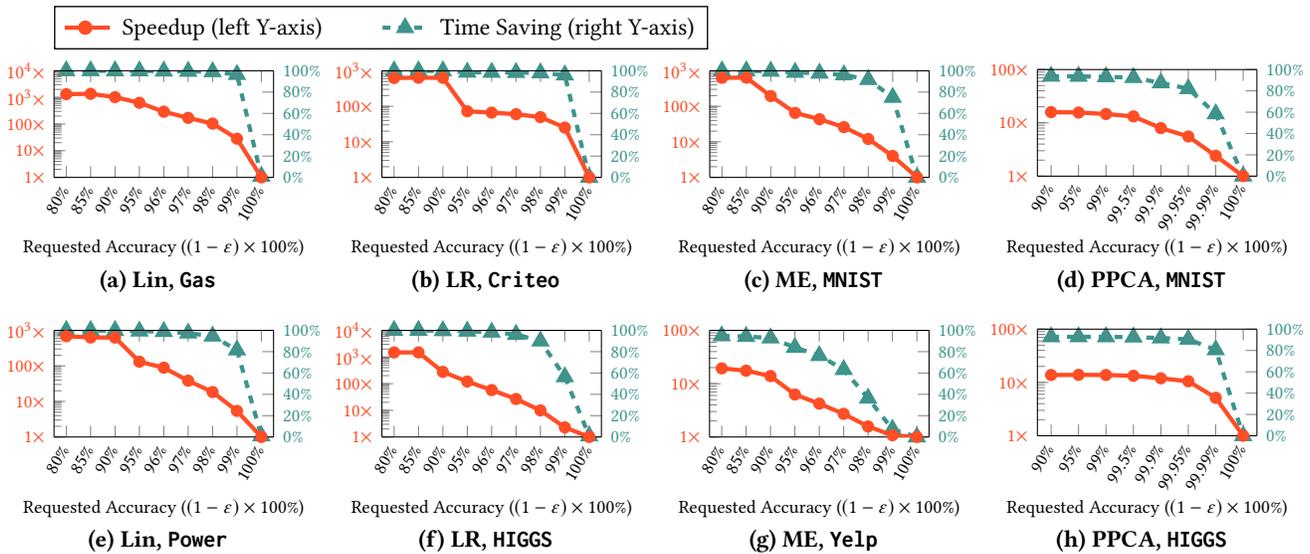

%% file: figures/fig_exp_acc_guarantee.tex
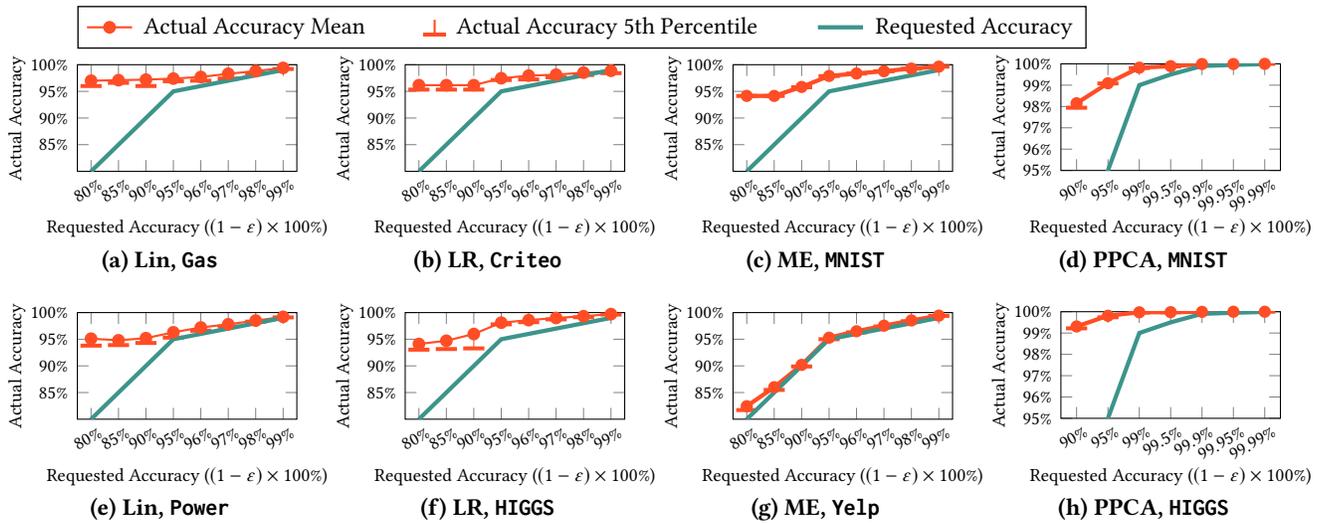
\begin{figure*}[t]

\pgfplotsset{modelagree/.style={
        width=45mm,
        height=30mm,
        xmin=0.5,
        xmax=8.5,
        ymin=80,
        ymax=100,
        xlabel=Requested Accuracy ($(1-\varepsilon) \times 100$\%),
        ylabel=Actual Accuracy,
        xlabel near ticks,
        ylabel near ticks,
        ylabel style={align=center},
        xtick={1, 2, 3, 4, 5, 6, 7, 8},
        xticklabels={80\%, 85\%, 90\%, 95\%, 96\%, 97\%, 98\%, 99\%},
        ytick={85, 90, ..., 100},
        yticklabels={85\%, 90\%, 95\%, 100\%},
        xticklabel style={rotate=30,yshift=-2mm,anchor=east,xshift=1.5mm},
        % yticklabels={0 hr, 0.5 hr, 1 hr, 1.5 hr, 2.0 hr},
        % extra y ticks={10, 20, 30, 40},
        % extra y tick style={yticklabels={}},
        ylabel shift=-5pt,
        xlabel shift=-2pt,
        legend style={
            at={(0,1.15)},anchor=south west,column sep=2pt,
            draw=black,fill=none,line width=.5pt,
            /tikz/every even column/.append style={column sep=20pt},
            font=\small
        },
        legend columns=3,
        every axis/.append style={font=\footnotesize},
        % ymajorgrids,
        % yminorgrids,
        minor grid style=lightgray,
        % nodes near coords,
        % every node near coord/.append style={font=\footnotesize,anchor=west,rotate=60}
    }}

% \centering

\begin{subfigure}[t]{0.24\textwidth}
\centering
\begin{tikzpicture}
\begin{axis}[modelagree,
]

\addplot[mark=*,mark size=2,mark options={fill=vintageorange,draw=vintageorange},
thick,draw=vintageorange,
error bars/.cd, y dir=minus, y explicit,
    error mark options={
      rotate=90,
      vintageorange,
      mark size=4pt,
      line width=1.5pt
    }
]
table[x=x,y=y,
    y error minus expr=\thisrow{y}-\thisrow{neg},
    y error plus expr=\thisrow{pos}-\thisrow{y}] {
x y neg pos
1  97.0  96.0  97.8
2  97.1  96.6  97.7
3  97.2  96.0  98.0
4  97.4  96.9  98.0
5  97.7  97.0  98.3
6  98.3  97.4  98.9
7  98.8  98.4  99.2
8  99.4  99.2  99.5
};

\addlegendentry{Actual Accuracy Mean}

\pgfplotsset{
  /pgfplots/mylegendimage/.style={
    legend image code/.code={
      \draw [vintageorange,thick] (0cm,0.1cm) -- (0cm,-0.1cm);
      \node [inner sep=0,minimum height=0.5mm,minimum width=3mm,fill=vintageorange]
        at (0cm,-0.1cm) {};
    },
  },
}
\addlegendimage{mylegendimage}
\addlegendentry{Actual Accuracy 5th Percentile}

\addplot[mark=none,
ultra thick,draw=vintagegreen,
]
table[x=x,y=y] {
x y
1  80
2  85
3  90
4  95
5  96
6  97
7  98
8  99
};

\addlegendentry{Requested Accuracy}

\end{axis}
\end{tikzpicture}

\vspace{-2mm}
\caption{Lin, \gas}
\end{subfigure}
~
\begin{subfigure}[t]{0.24\textwidth}
\centering
\begin{tikzpicture}
\begin{axis}[modelagree,
]

\addplot[mark=*,mark size=2,mark options={fill=vintageorange,draw=vintageorange},
thick,draw=vintageorange,
error bars/.cd, y dir=minus, y explicit,
    error mark options={
      rotate=90,
      vintageorange,
      mark size=4pt,
      line width=1.5pt
    }
]
table[x=x,y=y,
    y error minus expr=\thisrow{y}-\thisrow{neg},
    y error plus expr=\thisrow{pos}-\thisrow{y}] {
x y neg pos
1  96.16  95.33  96.51
2  96.16  95.33  96.51
3  96.16  95.33  96.51
4  97.46  97.13  98.20
5  97.95  97.24  98.43
6  98.13  97.64  98.49
7  98.51  98.07  98.95
8  98.86  98.42  99.27
};

% \addlegendentry{Actual Accuracy Mean}

\pgfplotsset{
  /pgfplots/mylegendimage/.style={
    legend image code/.code={
      \draw [vintageorange,thick] (0cm,0.1cm) -- (0cm,-0.1cm);
      \node [inner sep=0,minimum height=0.5mm,minimum width=3mm,fill=vintageorange]
        at (0cm,-0.1cm) {};
    },
  },
}
% \addlegendimage{mylegendimage}
% \addlegendentry{Actual Accuracy 5th Percentile}

\addplot[mark=none,
ultra thick,draw=vintagegreen,
]
table[x=x,y=y] {
x y
1  80
2  85
3  90
4  95
5  96
6  97
7  98
8  99
};

\end{axis}
\end{tikzpicture}

\vspace{-2mm}
\caption{LR, \criteo}
\end{subfigure}
~
\begin{subfigure}[t]{0.24\textwidth}
\centering
\begin{tikzpicture}
\begin{axis}[modelagree]
\addplot[mark=*,mark size=1.5,mark options={fill=vintageorange},
ultra thick,draw=vintageorange,
error bars/.cd, y dir=minus, y explicit,
    error mark options={
      rotate=90,
      vintageorange,
      mark size=4pt,
      line width=1.5pt
    }
]
table[x=x,y=y,
    y error minus expr=\thisrow{y}-\thisrow{neg},
    y error plus expr=\thisrow{pos}-\thisrow{y}] {
x y neg pos
1  94.15  94.14  94.15
2  94.15  94.14  94.15
3  95.83  95.77  95.89
4  97.89  97.87  97.91
5  98.35  98.35  98.37
6  98.78  98.77  98.8
7  99.28  99.26  99.29
8  99.64  99.63  99.66
};

\addplot[mark=none,
ultra thick,draw=vintagegreen,
]
table[x=x,y=y] {
x y
1  80
2  85
3  90
4  95
5  96
6  97
7  98
8  99
};

\end{axis}
\end{tikzpicture}

\vspace{-2mm}
\caption{ME, \mnist}
\end{subfigure}
~
\begin{subfigure}[t]{0.24\textwidth}
\centering
\begin{tikzpicture}
\begin{axis}[modelagree,
xmin=0.5,
xmax=7.5,
xtick={1, 2, 3, 4, 5, 6, 7},
xticklabels={90\%, 95\%, 99\%, 99.5\%, 99.9\%, 99.95\%, 99.99\%},
xlabel shift=-2mm,
ymin=0.95,
ymax=1.0,
ytick={0.95, 0.96, 0.97, 0.98, 0.99, 1.0},
yticklabels={95\%, 96\%, 97\%, 98\%, 99\%, 100\%},
xlabel=Requested Accuracy ($(1-\varepsilon) \times 100\%$),
]

\addplot[mark=*,mark size=1.5,mark options={fill=vintageorange},
ultra thick,draw=vintageorange,
error bars/.cd, y dir=minus, y explicit,
    error mark options={
      rotate=90,
      vintageorange,
      mark size=4pt,
      line width=1.5pt
    }
]
table[x=x,y=y,
    y error minus expr=\thisrow{y}-\thisrow{neg},
    y error plus expr=\thisrow{pos}-\thisrow{y}] {
x y neg pos
1  0.9815  0.9794  0.9825
2  0.9909  0.9908  0.9912
3  0.9982  0.998   0.9984
4  0.9989  0.9988  0.9989
5  0.9999  0.9998  0.9999
6  0.9999  0.9999  0.9999
7  1       1       1
};

\addplot[mark=none,
ultra thick,draw=vintagegreen,
]
table[x=x,y=y] {
x y
1  0.90
2  0.95
3  0.99
4  0.995
5  0.999
6  0.9995
7  0.9999
};
\end{axis}
\end{tikzpicture}

\vspace{-2mm}
\caption{PPCA, \mnist}
\end{subfigure}

%%%%%%%%%%%%%%
% SECOND ROW
%%%%%%%%%%%%%%

\vspace{2mm}

\begin{subfigure}[t]{0.24\textwidth}
\centering
\begin{tikzpicture}
\begin{axis}[modelagree,
]

\addplot[mark=*,mark size=2,mark options={fill=vintageorange,draw=vintageorange},
thick,draw=vintageorange,
error bars/.cd, y dir=minus, y explicit,
    error mark options={
      rotate=90,
      vintageorange,
      mark size=4pt,
      line width=1.5pt
    }
]
table[x=x,y=y,
    y error minus expr=\thisrow{y}-\thisrow{neg},
    y error plus expr=\thisrow{pos}-\thisrow{y}] {
x y neg pos
1  95.1  93.8  96.0
2  94.8  93.9  95.9
3  95.2  94.3  96.1
4  96.3  95.3  96.9
5  97.2  96.6  97.6
6  97.8  97.4  98.2
7  98.5  98.3  98.8
8  99.2  99.1  99.4
};

\addplot[mark=none,
ultra thick,draw=vintagegreen,
]
table[x=x,y=y] {
x y
1  80.0
2  85.0
3  90.0
4  95.0
5  96.0
6  97.0
7  98.0
8  99.0
};

\end{axis}
\end{tikzpicture}

\vspace{-2mm}
\caption{Lin, \power}
\end{subfigure}
~
\begin{subfigure}[t]{0.24\textwidth}
\centering
\begin{tikzpicture}
\begin{axis}[modelagree,
]

\addplot[mark=*,mark size=2,mark options={fill=vintageorange,draw=vintageorange},
thick,draw=vintageorange,
error bars/.cd, y dir=minus, y explicit,
    error mark options={
      rotate=90,
      vintageorange,
      mark size=4pt,
      line width=1.5pt
    }
]
table[x=x,y=y,
    y error minus expr=\thisrow{y}-\thisrow{neg},
    y error plus expr=\thisrow{pos}-\thisrow{y}] {
x y neg pos
1  94.11  93.04  95.05
2  94.71  93.16  96.14
3  95.98  93.28  96.91
4  98.11  97.79  98.36
5  98.58  98.20  98.99
6  98.99  98.72  99.21
7  99.32  99.13  99.50
8  99.71  99.63  99.82
};

\pgfplotsset{
  /pgfplots/mylegendimage/.style={
    legend image code/.code={
      \draw [vintageorange,thick] (0cm,0.1cm) -- (0cm,-0.1cm);
      \node [inner sep=0,minimum height=0.5mm,minimum width=3mm,fill=vintageorange]
        at (0cm,-0.1cm) {};
    },
  },
}

\addplot[mark=none,
ultra thick,draw=vintagegreen,
]
table[x=x,y=y] {
x y
1  80.0
2  85.0
3  90.0
4  95.0
5  96.0
6  97.0
7  98.0
8  99.0
};

% \addlegendentry{Actual Accuracy (mean by a circle and 5th percentile by a bar)
% \barzan{add two separate legends: one with a red and circle and one with a red and a bar}}
% \addlegendentry{Requested Accuracy}

% \draw[vintageorange,thick] (axis cs:0.5,6060) -- (axis cs:5.5,6060);

\end{axis}
\end{tikzpicture}

\vspace{-2mm}
\caption{LR, \higgs}
\end{subfigure}
~
\begin{subfigure}[t]{0.24\textwidth}
\centering
\begin{tikzpicture}
\begin{axis}[modelagree]
\addplot[mark=*,mark size=1.5,mark options={fill=vintageorange},
ultra thick,draw=vintageorange,
error bars/.cd, y dir=minus, y explicit,
    error mark options={
      rotate=90,
      vintageorange,
      mark size=4pt,
      line width=1.5pt
    }
]
table[x=x,y=y,
    y error minus expr=\thisrow{y}-\thisrow{neg},
    y error plus expr=\thisrow{pos}-\thisrow{y}] {
x y neg pos
1  82.44  81.71  83.09
2  86.00  85.49  86.49
3  90.19  89.89  90.50
4  95.29  95.00  95.53
5  96.50  96.30  96.69
6  97.54  97.40  97.65
7  98.52  98.38  98.65
8  99.43  99.38  99.51
};

\addplot[mark=none,
ultra thick,draw=vintagegreen,
]
table[x=x,y=y] {
x y
1  80.0
2  85.0
3  90.0
4  95.0
5  96.0
6  97.0
7  98.0
8  99.0
};

\end{axis}
\end{tikzpicture}

\vspace{-2mm}
\caption{ME, \yelp}
\end{subfigure}
~
\begin{subfigure}[t]{0.24\textwidth}
\centering
\begin{tikzpicture}
\begin{axis}[modelagree,
xmin=0.5,
xmax=7.5,
xtick={1, 2, 3, 4, 5, 6, 7},
xticklabels={90\%, 95\%, 99\%, 99.5\%, 99.9\%, 99.95\%, 99.99\%},
xlabel shift=-2mm,
ymin=0.95,
ymax=1.0,
ytick={0.95, 0.96, 0.97, 0.98, 0.99, 1.0},
yticklabels={95\%, 96\%, 97\%, 98\%, 99\%, 100\%},
xlabel=Requested Accuracy ($(1-\varepsilon) \times 100\%$),
]

\addplot[mark=*,mark size=1.5,mark options={fill=vintageorange},
ultra thick,draw=vintageorange,
error bars/.cd, y dir=minus, y explicit,
    error mark options={
      rotate=90,
      vintageorange,
      mark size=4pt,
      line width=1.5pt
    }
]
table[x=x,y=y,
    y error minus expr=\thisrow{y}-\thisrow{neg},
    y error plus expr=\thisrow{pos}-\thisrow{y}] {
x y neg pos
1  0.993073  0.992186  0.994133
2  0.998096  0.997379  0.998922
3  0.999719  0.999613  0.999776
4  0.999781  0.999714  0.999833
5  0.999924  0.999916  0.999936
6  0.999984  0.999979  0.999987
7  0.999988  0.999986  0.999989
};

\addplot[mark=none,
ultra thick,draw=vintagegreen,
]
table[x=x,y=y] {
x y
1  0.90
2  0.95
3  0.99
4  0.995
5  0.999
6  0.9995
7  0.9999
};
\end{axis}
\end{tikzpicture}

\vspace{-2mm}
\caption{PPCA, \higgs}
\end{subfigure}

\vspace{-2mm}
\caption{The correctness of \system. The requested model accuracies were compared to the actual model accuracies. In most cases, 95\% of the actual model accuracies were (or equivalently, the 5th percentile of the actual accuracies was) higher than the requested accuracies.}
\label{fig:exp:acc}
\end{figure*}

%% file: figures/fig_exp_baseline.tex
\begin{figure*}[t]

\pgfplotsset{sss/.style={
        width=48mm,
        height=32mm,
        xmin=0.5,
        xmax=8.5,
        ymin=0,
        ymax=100,
        xlabel=Requested Accuracy,
        ylabel=Sample Size,
        xlabel near ticks,
        ylabel near ticks,
        ylabel style={align=center},
        xtick={1, 2, 3, 4, 5, 6, 7, 8},
        xticklabels={80\%, 85\%, 90\%, 95\%, 96\%, 97\%, 98\%, 99\%},
        ytick={0, 20, ..., 100},
        yticklabels={0\%, 20\%, 40\%, 60\&, 80\%, 100\%},
        % yticklabels={0 hr, 0.5 hr, 1 hr, 1.5 hr, 2.0 hr},
        % extra y ticks={10, 20, 30, 40},
        % extra y tick style={yticklabels={}},
        ylabel shift=-5pt,
        xlabel shift=-5pt,
        legend style={
            at={(0,1.1)},anchor=south west,column sep=2pt,
            draw=black,fill=none,line width=.5pt,
            /tikz/every even column/.append style={column sep=10pt},
            font=\small
        },
        legend cell align={left},
        legend columns=10,
        every axis/.append style={font=\footnotesize},
        % ymajorgrids,
        xticklabel style={rotate=30,yshift=-2mm,anchor=east,xshift=1.5mm},
        % yminorgrids,
        % minor grid style=lightgray,
        % nodes near coords,
        % every node near coord/.append style={font=\footnotesize,rotate=0}
        scaled y ticks=false
    }}

\centering
\begin{subfigure}[t]{0.49\textwidth}
\centering
  \begin{minipage}{0.48\linewidth}
  \begin{tikzpicture}
  \begin{axis}[sss,
    ymin=92,
    ymax=100,
    ytick={90, 92, 94, 96, 98, 100},
    yticklabels={90\%, 92\%, 94\%, 96\%, 98\%, 100\%},
    ylabel=Actual Accuracy,
  ]

  \addplot[mark=x,mark size=2.5,ultra thick,draw=vintageblack,
  mark options={fill=vintageblack}]
  table[x=x,y=y] {
  x y
  1  96.10
  2  96.10
  3  96.10
  4  96.10
  5  96.10
  6  96.10
  7  96.10
  8  96.10
  };

  \addplot[mark=square*,mark size=1.0,ultra thick,draw=vintageblack!50!white,
  mark options={fill=vintageblack!50!white}]
  table[x=x,y=y] {
  x y
  1  98.68
  2  98.74
  3  98.75
  4  98.85
  5  98.78
  6  98.82
  7  98.79
  8  98.84
  };

  \addplot[mark=triangle*,mark size=1.5,ultra thick,draw=vintagegreen,
  mark options={fill=vintagegreen}]
  table[x=x,y=y] {
  x y
  1  94.84
  2  94.84
  3  94.83
  4  97.48
  5  97.48
  6  98.54
  7  98.85
  8  99.29
  };

  \addplot[mark=*,mark size=1.5,ultra thick,draw=vintageorange,
  mark options={fill=vintageorange}]
  table[x=x,y=y] {
  x y
  1  95.06
  2  94.84
  3  95.18
  4  96.28
  5  97.16
  6  97.83
  7  98.54
  8  99.25
  };

  \addlegendentry{FixedRatio}
  \addlegendentry{RelativeRatio}
  \addlegendentry{\gradual}
  \addlegendentry{\system}

  \addlegendimage{mark=*,mark size=1.5,ultra thick,draw=vintagepurple,
    mark options={fill=vintagepurple}}
  \addlegendentry{\system's pure training time}

  \node[anchor=south east,fill=white,draw=black] at (axis cs: 8.5,92) {Lin, \power};

  \end{axis}
  \end{tikzpicture}
  \end{minipage}
  \hfill
  \begin{minipage}{0.48\linewidth}
    \vspace{5mm}

    % LR, Criteo
    \begin{tikzpicture}
    \begin{axis}[sss,
      ymin=92,
      ymax=100,
      ytick={90, 92, 94, 96, 98, 100},
      yticklabels={90\%, 92\%, 94\%, 96\%, 98\%, 100\%},
      ylabel=Actual Accuracy,
    ]

    \addplot[mark=x,mark size=2.5,ultra thick,draw=vintageblack,
    mark options={fill=vintageblack}]
    table[x=x,y=y] {
    x y
    1  98.63
    2  98.63
    3  98.63
    4  98.63
    5  98.63
    6  98.63
    7  98.63
    8  98.63
    };

    \addplot[mark=square*,mark size=1.0,ultra thick,draw=vintageblack!50!white,
    mark options={fill=vintageblack!50!white}]
    table[x=x,y=y] {
    x y
    1  99.01
    2  98.98
    3  99.11
    4  98.98
    5  98.97
    6  98.99
    7  99.05
    8  98.82
    };

    \addplot[mark=triangle*,mark size=1.5,ultra thick,draw=vintagegreen,
    mark options={fill=vintagegreen}]
    table[x=x,y=y] {
    x y
    1  93.95
    2  93.94
    3  96.82
    4  97.00
    5  97.37
    6  98.19
    7  97.99
    8  99.35
    };

    \addplot[mark=*,mark size=1.5,ultra thick,draw=vintageorange,
    mark options={fill=vintageorange}]
    table[x=x,y=y] {
    x y
    1  96.16
    2  96.16
    3  96.16
    4  97.46
    5  97.95
    6  98.13
    7  98.51
    8  98.86
    };

    \node[anchor=south east,fill=white,draw=black] at (axis cs: 8.5,92) {LR, \criteo};

    \end{axis}
    \end{tikzpicture}

  \end{minipage}

\vspace{-2mm}
\caption{Effectiveness of Sample Size Estimator}
\label{fig:exp:sss:a}
\end{subfigure}
\hfill
\begin{subfigure}[t]{0.49\textwidth}
    \begin{minipage}{0.48\linewidth}
    \vspace{5mm}

    % Lin, Power
    \begin{tikzpicture}
    \begin{axis}[sss,
      ymin=1,
      ymax=1000,
      ytick={1, 10, 100, 1000},
      yticklabels={1, 10, 100, 1K},
      ymode=log,
      ylabel=Runtime (sec),
      ylabel shift=-5pt,
    ]

    \addplot[mark=x,mark size=2.5,ultra thick,draw=vintageblack,
    mark options={fill=vintageblack}]
    table[x=x,y=y] {
    x y
    1  2.4
    2  2.4
    3  2.4
    4  2.4
    5  2.4
    6  2.4
    7  2.4
    8  2.4
    };

    \addplot[mark=square*,mark size=1.0,ultra thick,draw=vintageblack!50!white,
    mark options={fill=vintageblack!50!white}]
    table[x=x,y=y] {
    x y
    1  54.4
    2  63.9
    3  60.2
    4  66.1
    5  61.0
    6  60.3
    7  62.0
    8  63.5
    };

    \addplot[mark=triangle*,mark size=1.5,ultra thick,draw=vintagegreen,
    mark options={fill=vintagegreen}]
    table[x=x,y=y] {
    x y
    1  1.8
    2  2.4
    3  1.9
    4  40.1
    5  42.4
    6  115.1
    7  201.8
    8  735.0
    };

    \addplot[mark=*,mark size=1.5,ultra thick,draw=vintageorange,
    mark options={fill=vintageorange}]
    table[x=x,y=y] {
    x y
    1  1.8
    2  2.4
    3  1.9
    4  6.67
    5  9.83
    6  22.85
    7  47.63
    8  161.66
    };

    \addplot[mark=*,mark size=1.5,ultra thick,draw=vintagepurple,
    mark options={fill=vintagepurple}]
    table[x=x,y=y] {
    x y
    1  1.79
    2  2.38
    3  1.86
    4  3.67
    5  6.83
    6  19.85
    7  44.63
    8  158.66
    };

    \node[anchor=north west,fill=white,draw=black] at (axis cs: 0.5,1000) {Lin, \power};

    \end{axis}
    \end{tikzpicture}
    \end{minipage}
    \hfill
    \begin{minipage}{0.48\linewidth}
    \vspace{5mm}

    % LR, Criteo
    \begin{tikzpicture}
    \begin{axis}[sss,
      ymin=1,
      ymax=10000,
      ytick={1, 10, 100, 1000, 10000},
      yticklabels={1, 10, 100, 1K, 10K},
      ymode=log,
      ylabel=Runtime (sec),
      ylabel shift=-5pt,
    ]

    \addplot[mark=x,mark size=2.5,ultra thick,draw=vintageblack,
    mark options={fill=vintageblack}]
    table[x=x,y=y] {
    x y
    1  78.2
    2  78.2
    3  78.2
    4  78.2
    5  78.2
    6  78.2
    7  78.2
    8  78.2
    };

    \addplot[mark=square*,mark size=1.0,ultra thick,draw=vintageblack!50!white,
    mark options={fill=vintageblack!50!white}]
    table[x=x,y=y] {
    x y
    1  700.0
    2  710.0
    3  774.0
    4  721.4
    5  709.3
    6  812.7
    7  786.4
    8  719.8
    };

    \addplot[mark=triangle*,mark size=1.5,ultra thick,draw=vintagegreen,
    mark options={fill=vintagegreen}]
    table[x=x,y=y] {
    x y
    1  9.10
    2  12.39
    3  32.80
    4  115.34
    5  101.21
    6  212.78
    7  822.78
    8  5703.52
    };

    \addplot[mark=*,mark size=1.5,ultra thick,draw=vintageorange,
    mark options={fill=vintageorange}]
    table[x=x,y=y] {
    x y
    1  9.14
    2  8.99
    3  9.04
    4  78.93
    5  86.37
    6  95.45
    7  115.47
    8  227.56
    };

    \addplot[mark=*,mark size=1.5,ultra thick,draw=vintagepurple,
    mark options={fill=vintagepurple}]
    table[x=x,y=y] {
    x y
    1  9.14
    2  8.99
    3  9.04
    4  18.93
    5  26.37
    6  35.45
    7  55.47
    8  167.56
    };

    \node[anchor=south east,fill=white,draw=black] at (axis cs: 8.5,1) {LR, \criteo};

    \end{axis}
    \end{tikzpicture}
    \end{minipage}

  \vspace{-2mm}
  \caption{Efficiency of Sample Size Estimator}
  \label{fig:exp:sss:b}
\end{subfigure}

\vspace{-2mm}
\caption{The effectiveness and efficiency of \system's sample size estimator.
Two baselines (FixedRatio and RelativeRatio) either failed to satisfy
the requested accuracies or were costly. IncEstimator met the requested accuracies, but was often quite slow.}
\label{fig:exp:sss}
\end{figure*}
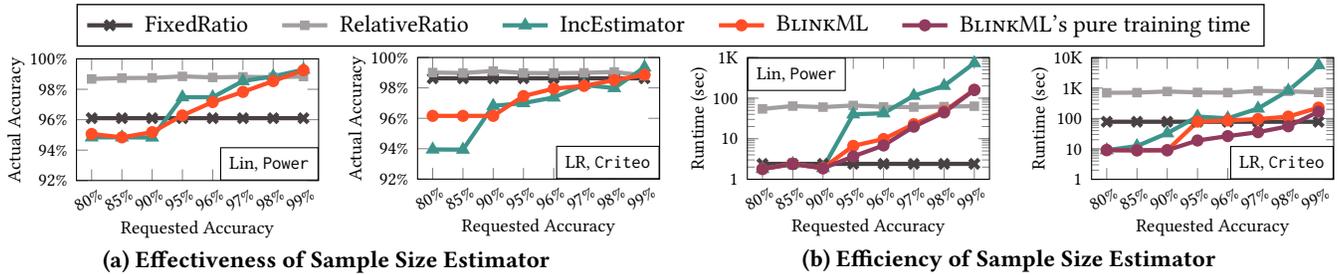

%% file: figures/fig_overhead_hyperparam.tex
\begin{figure*}[t]

\pgfplotsset{hyperparam/.style={
  width=48mm,
  height=29mm,
  xmin=0,
  xmax=1,
  ymin=0,
  ymax=1,
  xlabel=Number of Features,
  xlabel near ticks,
  ylabel near ticks,
  ylabel style={align=center},
  % xtick={1, 2, 3, 4, 5, 6, 7, 8, 9, 10},
  % xticklabels={80\%, 85\%, 90\%, 95\%, 96\%, 97\%, 98\%, 99\%, 99.5\%, 100\%},
  % xticklabel style={rotate=60,yshift=-0.5mm,anchor=east,xshift=0mm},
  % ytick={0, 1800, 3600, 5400, 7200},
  % yticklabels={0 hr, 0.5 hr, 1 hr, 1.5 hr, 2.0 hr},
  % ytick={1, 10, 100, 1000, 10000, 100000},
  % yticklabels={1$\times$, 10$\times$, 100$\times$, $10^3\times$, $10^4\times$, $10^5\times$},
  % every y tick label/.append style={color=vintageorange},
  % yticklabels={0\%, 20\%, 40\%, 60\%, 80\%, 100\%},
  % yticklabels={0 hr, 0.5 hr, 1 hr, 1.5 hr, 2.0 hr},
  % extra y ticks={10, 20, 30, 40},
  % extra y tick style={yticklabels={}},
  ylabel shift=-4pt,
  xlabel shift=-2pt,
  legend style={
      at={(0.0,1.1)},anchor=south west,column sep=2pt,
      draw=black,fill=white,line width=.5pt,
      /tikz/every even column/.append style={column sep=8pt},
  },
  legend columns=3,
  legend cell align={left},
  every axis/.append style={font=\footnotesize},
  ymajorgrids=false,
  % yminorgrids,
  minor grid style=lightgray,
  % nodes near coords,
  % every node near coord/.append style={font=\scriptsize,anchor=west,rotate=20}
}}

\begin{subfigure}[b]{0.50\textwidth}
% \centering
% \hspace*{2mm}
\begin{tikzpicture}

  \begin{axis}[
    hyperparam,
    width=62mm,
    height=32mm,
    xmin=0.5,
    xmax=9.5,
    xtick={1, 2, 3, 4, 5, 6, 7, 8, 9},
    xticklabels={100, 500, 1K, 5K, 10K, 50K, 100K, 500K, 998K},
    ymin=0,
    ymax=300,
    ytick={0, 50, 100, 150, 200, 250, 300},
    ylabel=Time (sec),
    ybar stacked,
    bar width=2mm,
    xtick style={draw=none},
    legend style={
        at={(1.05,1.0)},anchor=north west,column sep=2pt,
        draw=black,fill=white,line width=.5pt,
        /tikz/every even column/.append style={column sep=5pt},
        font=\scriptsize,
    },
    xticklabel style={rotate=30,yshift=-2mm,anchor=east,xshift=1.5mm},
    legend columns=1,
    clip=false,
    xlabel shift=-6pt,
  ]

  \addplot[draw=vintagepurple,fill=vintagepurple]
  table[x=x,y=y] {
    x y
    1 0.34
    2 0.64
    3 0.71
    4 0.59
    5 0.70
    6 0.89
    7 1.36
    8 1.74
    9 2.38
  };

  \addplot[draw=vintagegreen,fill=vintagegreen]
  table[x=x,y=y] {
    x y
    1 0.02
    2 0.08
    3 0.22
    4 1.22
    5 1.80
    6 6.92
    7 11.84
    8 81.74
    9 130.84
  };

  \addplot[draw=vintageorange,fill=vintageorange]
  table[x=x,y=y] {
    x y
    1 0.65
    2 0.70
    3 0.78
    4 1.28
    5 1.67
    6 3.72
    7 7.93
    8 42.77
    9 84.40
  };

  \addplot[draw=vintageblack!50!white,fill=vintageblack!50!white]
  table[x=x,y=y] {
    x y
    1 1.99
    2 6.29
    3 7.20
    4 12.38
    5 8.74
    6 14.46
    7 16.06
    8 16.03
    9 17.42
  };

  \node[anchor=south,font=\footnotesize\it,rotate=20] at (axis cs: 1, 3)  {0.16\%};
  \node[anchor=south,font=\footnotesize\it,rotate=20] at (axis cs: 2, 7.71) {0.19\%};
  \node[anchor=south,font=\footnotesize\it,rotate=20] at (axis cs: 3, 8.91) {0.13\%};
  \node[anchor=south,font=\footnotesize\it,rotate=20] at (axis cs: 4, 15.47) {1.08\%};
  \node[anchor=south,font=\footnotesize\it,rotate=20] at (axis cs: 5, 12.91) {0.11\%};
  \node[anchor=south,font=\footnotesize\it,rotate=20] at (axis cs: 6, 25.99) {0.49\%};
  \node[anchor=south,font=\footnotesize\it,rotate=20] at (axis cs: 7, 37.19) {0.68\%};
  \node[anchor=south,font=\footnotesize\it,rotate=20] at (axis cs: 8, 142.28) {2.16\%};
  \node[anchor=south,font=\footnotesize\it,rotate=20] at (axis cs: 9, 235.04) {3.79\%};

  \addlegendentry{Initial Training}
  \addlegendentry{Statistics Computation}
  \addlegendentry{Sample Size Searching}
  \addlegendentry{Final Training}

  \addlegendimage{empty legend}
  \addlegendentry{\hspace{-4mm}
    \emph{{\normalsize \%} Ratio to Full Training Time}}

  \end{axis}

\end{tikzpicture}

\vspace{-2mm}
\caption{Runtime Overhead}
\label{fig:exp:overhead}
\end{subfigure}
\hfill
\begin{subfigure}[b]{0.23\textwidth}
\vspace{-5mm}

\centering
\begin{tikzpicture}

  \begin{axis}[
    hyperparam,
    ybar,
    bar width=1.3mm,
    xmin=0.5,
    xmax=7.5,
    xtick={1, 2, 3, 4, 5, 6, 7},
    xticklabels={100, 500, 1K, 5K, 10K, 50K, 100K},
    ymin=15,
    ymax=30,
    ytick={0, 5, 10, 15, 20, 25, 30},
    yticklabels={0\%, 5\%, 10\%, 15\%, 20\%, 25\%, 30\%},
    ylabel=Gen.~Error,
    xtick style={draw=none},
    xticklabel style={yshift=1mm},
  ]

  \addplot[draw=vintagegreen,fill=vintagegreen]
  table[x=x,y=y] {
  x y
  1 26.31
  2 21.7
  3 21.01
  4 20.86
  5 20.83
  6 20.84
  7 20.83
  };

  \addplot[draw=vintageorange,fill=vintageorange,
    error bars/.cd, y dir=plus, y explicit,
      error mark options={
        rotate=90,
        black!50!vintageorange,
        mark size=4pt,
        line width=1pt
      }
  ]
  table[x=x,y=y,
    y error minus expr=\thisrow{y}-\thisrow{neg},
    y error plus expr=\thisrow{pos}-\thisrow{y},
  ] {
    x y neg pos
    1 26.4 26.38 27.12
    2 21.8 21.67 22.45
    3 21.04 20.73 21.52
    4 20.77 20.46 21.26
    5 20.68 20.24 21.04
    6 20.73 20.18 20.98
    7 20.71 20.23 21.03
  };
  
    % 1 26.4 26.38 26.45
    % 2 21.8 21.67 21.93
    % 3 21.04 20.73 21.33
    % 4 20.77 20.46 20.94
    % 5 20.68 20.24 20.92
    % 6 20.73 20.18 20.93
    % 7 20.71 20.23 20.92

  \addlegendentry{Full Training}
  \addlegendentry{\system}
  
  \pgfplotsset{
    /pgfplots/mylegendimage/.style={
      legend image code/.code={
        \node [inner sep=0,minimum height=0.5mm,minimum width=3mm,fill=black,anchor=south]
          at (0cm,0.1cm) {};
        \draw [vintageorange,thick] (0cm,0.1cm) -- (0cm,-0.1cm);
      },
    },
  }
  \addlegendimage{mylegendimage}
  \addlegendentry{Predicted Gen.~Error Bound}

  \end{axis}

\end{tikzpicture}

\vspace{-2mm}
\caption{Gen.~Error}
\label{fig:exp:generror}
\end{subfigure}
\hfill
\begin{subfigure}[b]{0.23\textwidth}
\centering
\begin{tikzpicture}

  \begin{axis}[
    hyperparam,
    ybar,
    bar width=1.2mm,
    xmin=0.5,
    xmax=7.5,
    xtick={1, 2, 3, 4, 5, 6, 7},
    xticklabels={100, 500, 1K, 5K, 10K, 50K, 100K},
    ymin=0,
    ymax=40,
    ytick={0, 10, 20, 30, 40},
    yticklabels={0, 10, 20, 30, 40},
    ylabel=\# of Iterations,
    xtick style={draw=none},
    xticklabel style={yshift=1mm},
  ]

  \addplot[draw=vintagegreen,fill=vintagegreen]
  table[x=x,y=y] {
  x y
  1 14
  2 24
  3 28
  4 24
  5 22
  6 27
  7 24
  };

  \addplot[draw=vintageorange,fill=vintageorange]
  table[x=x,y=y] {
    x y
    1 13
    2 27
    3 22
    4 22
    5 21
    6 23
    7 24
  };

  % \addlegendentry{Full Training}
  % \addlegendentry{\system}
  \end{axis}

\end{tikzpicture}

\vspace{-2mm}
\caption{\# of Iterations}
\label{fig:exp:itrcount}
\end{subfigure}

\vspace{-2mm}
\caption{(Left) \system's runtime overhead, (Middle) generalization error (i.e., the errors on test sets), and (Right) the number of iterations taken by an optimization algorithm (L-BFGS) for different numbers of features. In (a), the ratio above each bar is \system's entire training time compared to full training.}
\label{fig:exp:scale}
\end{figure*}
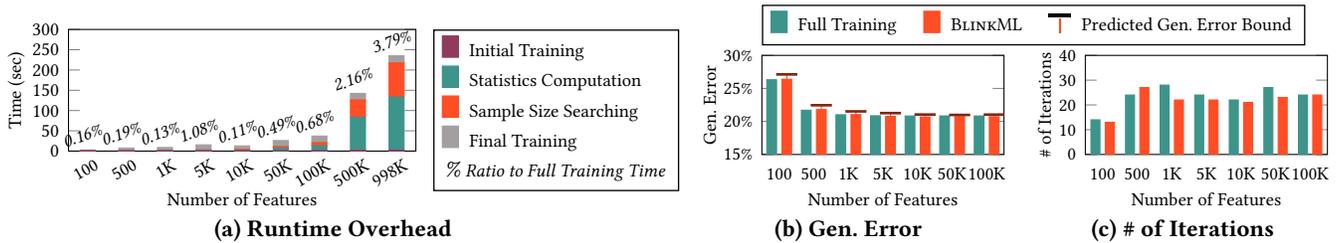

%% file: figures/tab_exp_hessian.tex
\begin{figure*}[t]
  \begin{subfigure}[b]{0.48\linewidth}
    \centering
    \begin{tikzpicture}

      \begin{axis}[
        ybar,
        bar width=1.4mm,
        width=68mm,
        height=32mm,
        % X-axis
        xmin=0.4,
        xmax=7.6,
        xlabel=Sample Size,
        xlabel near ticks,
        ylabel style={align=center},
        xtick={1, 2, 3, 4, 5, 6, 7},
        xticklabels={100, 500, 1K, 5K, 10K, 50K, 100K},
        xtick style={draw=none},
        % Y-axis
        ymin=0.5,
        ymax=2,
        ylabel=Est.~Var / Actual Var,
        ylabel near ticks,
        ytick={0, 0.5, 1.0, 1.5, 2.0},
        ylabel shift=-0pt,
        xlabel shift=-2pt,
        % Legend
        legend style={
            at={(1.05,1)},anchor=north west,column sep=2pt,
            draw=black,fill=white,line width=.5pt,
            /tikz/every even column/.append style={column sep=5pt},
            font=\small,
        },
        legend columns=1,
        legend cell align={left},
        % Others
        every axis/.append style={font=\footnotesize},
        ymajorgrids=false,
        minor grid style=lightgray,
        xticklabel style={yshift=2mm},
        scaled y ticks=false,
      ]

      \addplot[fill=vintageblack,draw=vintageblack]
      table[x=x,y=y] {
      x y
      1  1.4056
      2  1.2636
      3  1.3495
      4  1.2708
      5  1.3143
      6  1.2673
      7  1.2327
      };

      \addplot[fill=vintagegreen,draw=vintagegreen]
      table[x=x,y=y] {
      x y
      1  1.4614
      2  1.2880
      3  1.3719
      4  1.2891
      5  1.3334
      6  1.2856
      7  1.2503
      };

      \addplot[fill=vintageorange,draw=vintageorange]
      table[x=x,y=y] {
      x y
      1  1.7961
      2  1.3903
      3  1.4357
      4  1.2763
      5  1.3202
      6  1.2725
      7  1.2370
      };

      \addlegendentry{ClosedForm}
      \addlegendentry{InverseGradients}
      \addlegendentry{ObservedFisher}

      \draw[opacity=0.8,draw=none,fill=white] (axis cs: 0,0) rectangle (axis cs:7.5,1);
      \draw[red,dashed,ultra thick] (axis cs:0,1) -- (axis cs:8,1);
      \node[anchor=north west,fill=white,draw=red,dashed,thick,font=\footnotesize]
        at (axis cs:0.4,1) {\textsf{\textbf{Optimal Ratio}}};

      \end{axis}
    \end{tikzpicture}

    \vspace{-2mm}
    \caption{Estimation Tightness vs.~Sample Size}
  \end{subfigure}
  \hfill
  \begin{subfigure}[b]{0.48\linewidth}
  \centering
  % \small
  \fontsize{8}{9}\selectfont
    \begin{tabular}{l l r r}
    \toprule
    % \multirow{2}{*}{\textbf{Model, Data}} &
    % \multirow{2}{*}{\textbf{Metric}} &
    % \multicolumn{2}{c}{\textbf{Method}} \\
    % \cmidrule(lr){3-4}
    % \textbf{Dataset} & & IG & OF \\
    \textbf{Model, Data} & \textbf{Metric} & \textbf{IG} & \textbf{OF} \\
    \midrule
    \multirow{2}{*}{LR, \higgs}  & Runtime (sec) & 1.88 & 1.18 \\
                 & Accuracy ($\|\cdot\|_F$) & 0.00332   & 0.0039 \\
    \cmidrule(lr){1-4}
    \multirow{2}{*}{ME, \mnist} & Runtime (sec)  & 357.0 & 3.23 \\
    			 & Accuracy ($\|\cdot\|_F$) & 0.01298   & 0.00847 \\
    \bottomrule
    \end{tabular}

  \vspace{1mm}
  \caption{InverseGradients (IG) vs.~ObservedFisher (OF)}
  \label{tab:exp:hessian}
  \end{subfigure}

\caption{A study of statistics computation methods: (left) a comparison of estimated parameter variances to the actual parameter variances, and (right) a comparison of two statistics computation methods.}
\label{sec:exp:statistics}
\end{figure*}

%% file: figures/fig_hyperparam.tex
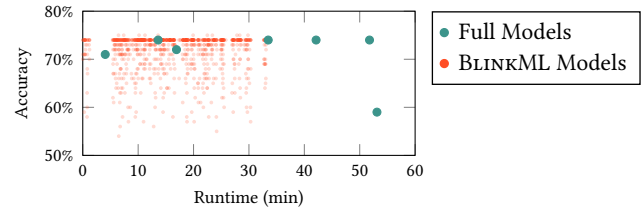
\begin{figure}[t]

  \pgfplotsset{hyperparam/.style={
    width=60mm,
    height=35mm,
    xmin=0,
    xmax=3600,
    ymin=0.5,
    ymax=0.8,
    xlabel=Runtime (min),
    ylabel=Accuracy,
    xlabel near ticks,
    ylabel near ticks,
    ylabel style={align=center},
    xtick={0, 600, 1200, 1800, 2400, 3000, 3600},
    xticklabels={0, 10, 20, 30, 40, 50, 60},
    ytick={0.5, 0.6, 0.7, 0.8},
    yticklabels={50\%, 60\%, 70\%, 80\%},
    % ytick={1, 10, 100, 1000, 10000, 100000},
    % yticklabels={1$\times$, 10$\times$, 100$\times$, $10^3\times$, $10^4\times$, $10^5\times$},
    % every y tick label/.append style={color=vintageorange},
    % yticklabels={0\%, 20\%, 40\%, 60\%, 80\%, 100\%},
    % yticklabels={0 hr, 0.5 hr, 1 hr, 1.5 hr, 2.0 hr},
    % extra y ticks={10, 20, 30, 40},
    % extra y tick style={yticklabels={}},
    ylabel shift=-2pt,
    xlabel shift=-2pt,
    legend style={
        at={(1.05,1)},anchor=north west,column sep=2pt,
        draw=black,fill=white,line width=.5pt,
        /tikz/every even column/.append style={column sep=10pt},
        font=\small,
    },
    legend columns=1,
    legend cell align={left},
    every axis/.append style={font=\footnotesize},
    ymajorgrids=false,
    % yminorgrids,
    minor grid style=lightgray,
    % nodes near coords,
    % every node near coord/.append style={font=\scriptsize,anchor=west,rotate=20}
  }}

\centering
\begin{tikzpicture}

  \begin{axis}[
    hyperparam
  ]

  \addlegendimage{only marks,mark=*,mark size=1.5,mark options={fill=vintagegreen,draw=vintagegreen},draw=none}
  \addlegendentry{Full Models}

  \addlegendimage{only marks,mark=*,mark size=1.5,mark options={fill=vintageorange,draw=vintageorange},draw=none}
  \addlegendentry{\system Models}

  \addplot[mark=*,mark size=0.5,mark options={fill=vintageorange,draw=vintageorange},
  draw=none,opacity=0.2]
  table[x=x,y=y] {
  x y
  0.51	0.70
  1.03	0.74
  1.78	0.71
  3.73	0.74
  4.82	0.74
  15.61	0.74
  16.64	0.59
  18.56	0.74
  18.85	0.71
  20.83	0.71
  21.99	0.71
  22.79	0.74
  23.16	0.74
  23.58	0.74
  23.89	0.74
  24.89	0.74
  27.37	0.74
  27.81	0.74
  28.12	0.71
  28.43	0.72
  30.24	0.73
  45.57	0.73
  46.45	0.58
  47.35	0.63
  48.27	0.60
  48.67	0.74
  48.96	0.74
  49.23	0.68
  49.51	0.71
  50.71	0.74
  51.67	0.67
  67.74	0.74
  69.16	0.73
  71.65	0.70
  72.02	0.74
  72.47	0.74
  72.96	0.74
  73.27	0.72
  323.90	0.74
  324.30	0.74
  325.93	0.74
  326.84	0.62
  327.14	0.72
  328.25	0.69
  329.16	0.67
  330.13	0.70
  333.97	0.67
  334.90	0.61
  336.78	0.74
  337.39	0.74
  339.48	0.74
  339.84	0.74
  340.25	0.72
  342.08	0.74
  342.42	0.73
  343.66	0.73
  344.04	0.74
  344.39	0.72
  344.94	0.74
  347.88	0.73
  350.02	0.74
  351.36	0.70
  351.72	0.74
  352.01	0.67
  352.35	0.71
  354.33	0.74
  355.70	0.66
  356.01	0.73
  357.78	0.72
  358.46	0.74
  358.84	0.72
  359.15	0.74
  360.06	0.64
  362.28	0.74
  364.00	0.74
  366.04	0.74
  367.51	0.65
  370.41	0.74
  371.59	0.69
  371.89	0.73
  372.22	0.74
  373.53	0.74
  373.92	0.74
  374.23	0.74
  376.13	0.74
  376.63	0.74
  377.01	0.73
  384.12	0.64
  386.01	0.74
  386.52	0.71
  386.84	0.74
  387.40	0.75
  387.79	0.72
  388.21	0.72
  388.79	0.74
  389.07	0.74
  390.06	0.54
  390.40	0.66
  391.64	0.74
  393.50	0.73
  394.89	0.74
  396.50	0.67
  397.79	0.72
  399.53	0.71
  400.01	0.73
  401.09	0.61
  402.60	0.63
  403.21	0.74
  403.74	0.74
  404.98	0.56
  405.31	0.71
  406.48	0.68
  406.95	0.74
  407.51	0.74
  408.78	0.73
  409.50	0.71
  410.60	0.68
  413.21	0.74
  414.59	0.68
  414.93	0.73
  415.85	0.63
  416.47	0.74
  416.80	0.74
  417.92	0.64
  418.93	0.66
  420.75	0.71
  421.50	0.74
  422.01	0.74
  422.42	0.73
  424.15	0.71
  424.60	0.72
  425.76	0.70
  426.08	0.74
  426.98	0.72
  427.99	0.73
  428.89	0.71
  429.47	0.71
  430.90	0.58
  432.27	0.64
  433.67	0.74
  434.42	0.74
  437.08	0.74
  437.54	0.74
  438.27	0.74
  440.50	0.74
  442.00	0.66
  443.53	0.75
  445.88	0.70
  447.30	0.67
  447.77	0.74
  450.60	0.74
  450.92	0.74
  451.33	0.74
  451.87	0.71
  452.56	0.74
  453.17	0.74
  453.60	0.74
  454.75	0.74
  456.22	0.63
  458.62	0.74
  460.18	0.71
  461.27	0.74
  461.62	0.74
  462.10	0.74
  462.90	0.73
  463.86	0.74
  464.23	0.71
  467.06	0.74
  468.47	0.73
  469.67	0.69
  470.15	0.74
  471.66	0.67
  473.81	0.65
  474.55	0.74
  476.31	0.68
  478.19	0.72
  478.81	0.74
  480.02	0.74
  481.89	0.74
  484.09	0.73
  484.77	0.74
  485.57	0.74
  486.23	0.72
  486.89	0.72
  488.92	0.72
  490.74	0.60
  492.59	0.70
  493.41	0.74
  494.69	0.57
  495.32	0.71
  496.15	0.74
  497.46	0.72
  498.94	0.72
  499.74	0.74
  500.73	0.74
  512.37	0.73
  512.95	0.74
  513.24	0.71
  514.21	0.60
  526.99	0.67
  529.34	0.74
  530.46	0.66
  532.83	0.74
  533.17	0.73
  535.28	0.74
  546.63	0.73
  547.93	0.72
  549.77	0.74
  550.86	0.70
  551.37	0.74
  551.70	0.74
  552.67	0.66
  562.52	0.65
  562.83	0.74
  563.37	0.74
  565.19	0.74
  565.49	0.74
  568.07	0.74
  569.12	0.58
  570.09	0.63
  570.42	0.74
  571.54	0.64
  573.43	0.72
  574.59	0.67
  574.92	0.73
  576.01	0.74
  576.38	0.72
  577.47	0.65
  578.33	0.59
  578.77	0.74
  579.21	0.73
  583.38	0.74
  583.69	0.74
  585.52	0.67
  587.63	0.74
  589.30	0.74
  589.69	0.72
  590.02	0.68
  590.38	0.74
  590.81	0.74
  591.20	0.72
  592.61	0.71
  593.18	0.74
  593.48	0.72
  594.91	0.72
  595.33	0.73
  595.67	0.71
  596.14	0.74
  598.39	0.74
  599.61	0.68
  599.96	0.74
  600.42	0.74
  600.73	0.66
  602.81	0.74
  603.25	0.74
  604.48	0.74
  604.82	0.74
  609.32	0.74
  609.79	0.75
  610.15	0.74
  610.50	0.74
  612.35	0.74
  619.99	0.69
  621.15	0.73
  621.48	0.71
  621.98	0.74
  622.27	0.69
  624.19	0.74
  625.43	0.71
  625.81	0.74
  626.24	0.74
  626.70	0.74
  627.77	0.63
  628.10	0.72
  628.43	0.74
  630.57	0.74
  631.12	0.74
  632.61	0.74
  633.22	0.74
  634.19	0.74
  634.73	0.74
  635.11	0.74
  635.64	0.74
  637.61	0.74
  639.73	0.74
  640.07	0.74
  641.17	0.66
  644.50	0.74
  645.72	0.69
  647.11	0.74
  647.63	0.74
  649.04	0.74
  649.39	0.74
  649.70	0.74
  651.03	0.66
  651.42	0.72
  651.80	0.74
  652.14	0.74
  654.90	0.74
  657.52	0.74
  658.63	0.57
  660.09	0.74
  660.41	0.74
  661.58	0.68
  661.96	0.74
  663.17	0.71
  663.53	0.74
  664.93	0.72
  665.27	0.74
  665.58	0.74
  665.89	0.72
  666.92	0.70
  669.16	0.74
  669.54	0.74
  670.08	0.74
  670.51	0.69
  670.82	0.74
  671.16	0.73
  671.42	0.69
  671.75	0.74
  672.27	0.74
  672.74	0.74
  674.00	0.75
  674.30	0.72
  675.29	0.68
  675.60	0.73
  676.72	0.72
  678.10	0.74
  680.87	0.74
  686.03	0.70
  691.94	0.74
  692.22	0.73
  693.77	0.74
  694.74	0.70
  711.52	0.73
  711.83	0.74
  717.80	0.74
  718.07	0.74
  718.41	0.74
  718.70	0.71
  721.20	0.74
  722.34	0.66
  722.66	0.69
  723.16	0.74
  724.33	0.68
  724.68	0.74
  726.02	0.72
  726.37	0.73
  726.66	0.73
  727.07	0.74
  728.48	0.74
  728.80	0.74
  729.17	0.74
  731.57	0.73
  732.92	0.66
  734.41	0.73
  735.39	0.69
  736.61	0.65
  736.91	0.74
  737.23	0.74
  737.58	0.74
  739.48	0.74
  739.76	0.74
  740.16	0.74
  740.67	0.74
  743.01	0.73
  744.14	0.73
  767.22	0.73
  768.12	0.56
  769.09	0.68
  772.23	0.74
  772.51	0.74
  773.69	0.74
  774.78	0.67
  775.06	0.70
  775.55	0.74
  777.25	0.74
  778.83	0.73
  779.20	0.74
  782.43	0.74
  783.31	0.60
  784.24	0.63
  784.53	0.73
  784.92	0.72
  786.29	0.73
  786.72	0.74
  787.91	0.68
  788.40	0.74
  789.27	0.62
  790.40	0.72
  790.84	0.74
  791.84	0.74
  792.19	0.70
  793.77	0.72
  796.50	0.74
  798.53	0.74
  812.56	0.70
  813.91	0.72
  814.25	0.74
  815.31	0.67
  815.91	0.74
  816.30	0.75
  816.77	0.74
  817.33	0.74
  819.28	0.74
  819.68	0.74
  821.41	0.74
  822.53	0.68
  836.30	0.72
  836.61	0.74
  837.09	0.73
  837.42	0.72
  838.51	0.72
  840.79	0.66
  843.12	0.74
  843.46	0.74
  844.70	0.70
  845.60	0.64
  845.89	0.69
  846.20	0.71
  846.51	0.69
  847.02	0.74
  847.30	0.71
  847.84	0.74
  849.27	0.72
  849.77	0.74
  853.47	0.74
  854.79	0.72
  856.79	0.74
  858.58	0.74
  861.64	0.74
  863.09	0.69
  864.86	0.74
  865.18	0.68
  865.54	0.74
  865.94	0.74
  866.49	0.74
  867.03	0.74
  867.46	0.74
  867.87	0.74
  868.23	0.73
  868.52	0.72
  868.81	0.71
  869.71	0.59
  870.05	0.74
  870.96	0.62
  873.77	0.73
  874.13	0.73
  876.53	0.71
  877.98	0.72
  878.33	0.74
  878.66	0.74
  879.04	0.74
  880.01	0.67
  880.30	0.74
  881.45	0.68
  882.57	0.63
  883.88	0.67
  884.23	0.73
  884.70	0.74
  885.06	0.72
  887.66	0.72
  888.11	0.72
  899.14	0.74
  901.25	0.74
  903.61	0.74
  904.86	0.74
  905.82	0.69
  909.01	0.74
  909.95	0.67
  910.28	0.74
  910.62	0.74
  911.50	0.64
  912.49	0.72
  912.79	0.74
  913.68	0.64
  913.96	0.74
  914.33	0.74
  928.44	0.72
  928.77	0.73
  930.72	0.74
  931.04	0.74
  932.11	0.70
  932.44	0.74
  933.38	0.64
  935.21	0.74
  936.55	0.71
  937.52	0.69
  938.08	0.74
  940.65	0.74
  940.95	0.74
  941.76	0.74
  942.04	0.74
  942.51	0.74
  946.42	0.74
  946.77	0.74
  948.57	0.74
  949.10	0.74
  949.43	0.74
  949.74	0.74
  950.09	0.74
  950.43	0.74
  950.80	0.74
  951.77	0.68
  953.01	0.74
  954.56	0.74
  954.85	0.71
  955.20	0.74
  972.62	0.73
  973.16	0.74
  974.15	0.71
  975.22	0.73
  975.55	0.74
  976.43	0.58
  977.97	0.74
  978.62	0.74
  980.17	0.73
  981.15	0.70
  981.42	0.74
  981.78	0.74
  982.15	0.74
  982.43	0.74
  984.10	0.74
  984.44	0.74
  986.54	0.74
  987.75	0.72
  988.77	0.69
  989.86	0.66
  990.15	0.74
  990.49	0.74
  990.80	0.74
  992.14	0.73
  992.67	0.74
  993.55	0.63
  994.00	0.74
  995.23	0.73
  995.55	0.73
  996.50	0.68
  998.40	0.73
  1041.57	0.74
  1042.84	0.72
  1043.85	0.66
  1056.85	0.74
  1057.38	0.74
  1080.69	0.71
  1081.59	0.70
  1081.99	0.74
  1082.87	0.61
  1083.40	0.74
  1083.75	0.74
  1084.84	0.74
  1085.35	0.74
  1086.56	0.74
  1086.88	0.74
  1087.83	0.70
  1088.15	0.73
  1089.25	0.74
  1090.77	0.74
  1091.08	0.74
  1091.43	0.74
  1092.49	0.74
  1093.98	0.74
  1107.65	0.71
  1107.93	0.74
  1108.20	0.72
  1110.35	0.74
  1110.64	0.74
  1110.99	0.74
  1111.25	0.71
  1112.43	0.74
  1112.72	0.74
  1113.64	0.66
  1115.85	0.74
  1116.34	0.73
  1116.67	0.74
  1117.02	0.74
  1117.40	0.74
  1118.30	0.65
  1118.64	0.74
  1119.05	0.74
  1120.98	0.74
  1121.24	0.69
  1121.69	0.74
  1122.05	0.71
  1122.58	0.74
  1122.89	0.74
  1123.18	0.70
  1125.08	0.74
  1125.61	0.74
  1126.48	0.58
  1126.87	0.74
  1127.28	0.73
  1128.54	0.63
  1129.35	0.57
  1129.68	0.74
  1131.74	0.74
  1132.68	0.67
  1133.00	0.73
  1134.01	0.71
  1134.39	0.74
  1134.67	0.68
  1137.08	0.74
  1137.47	0.72
  1138.49	0.67
  1140.57	0.74
  1141.55	0.72
  1141.93	0.74
  1142.21	0.74
  1142.66	0.74
  1142.98	0.74
  1150.45	0.69
  1170.26	0.73
  1171.90	0.74
  1172.95	0.68
  1173.25	0.74
  1173.53	0.70
  1174.56	0.74
  1176.82	0.69
  1177.23	0.74
  1177.73	0.74
  1178.00	0.69
  1178.41	0.74
  1178.69	0.72
  1179.03	0.74
  1179.38	0.73
  1180.38	0.59
  1180.70	0.73
  1182.22	0.74
  1183.24	0.72
  1183.58	0.74
  1187.52	0.74
  1187.96	0.74
  1188.32	0.74
  1190.46	0.74
  1192.37	0.74
  1192.96	0.74
  1193.27	0.69
  1195.04	0.74
  1195.46	0.75
  1195.79	0.74
  1196.11	0.74
  1197.32	0.73
  1197.64	0.74
  1198.14	0.74
  1200.39	0.74
  1200.72	0.70
  1201.01	0.71
  1201.43	0.74
  1203.45	0.74
  1203.80	0.73
  1204.11	0.74
  1204.39	0.66
  1205.71	0.74
  1206.58	0.57
  1207.58	0.64
  1207.85	0.73
  1209.02	0.69
  1210.64	0.74
  1210.90	0.68
  1211.85	0.68
  1217.80	0.62
  1218.09	0.73
  1245.09	0.74
  1245.46	0.74
  1246.39	0.65
  1248.32	0.74
  1248.63	0.74
  1249.17	0.74
  1249.47	0.71
  1249.80	0.65
  1250.92	0.64
  1251.30	0.73
  1252.37	0.61
  1252.79	0.74
  1253.11	0.72
  1253.64	0.74
  1255.63	0.74
  1255.95	0.73
  1259.39	0.73
  1265.27	0.66
  1266.61	0.74
  1267.00	0.74
  1267.27	0.73
  1275.86	0.71
  1276.79	0.61
  1277.99	0.73
  1278.27	0.72
  1296.76	0.74
  1297.64	0.64
  1298.74	0.74
  1299.05	0.74
  1300.32	0.72
  1300.60	0.70
  1301.62	0.71
  1302.59	0.71
  1303.45	0.65
  1305.33	0.71
  1306.35	0.60
  1306.67	0.74
  1308.71	0.74
  1310.88	0.74
  1311.29	0.72
  1312.74	0.63
  1313.72	0.69
  1314.64	0.67
  1314.95	0.73
  1316.08	0.73
  1316.47	0.74
  1316.81	0.74
  1317.83	0.66
  1319.10	0.74
  1321.04	0.74
  1322.19	0.74
  1322.49	0.72
  1323.60	0.67
  1325.63	0.74
  1328.04	0.74
  1329.60	0.71
  1330.50	0.68
  1331.69	0.70
  1333.24	0.74
  1333.57	0.72
  1333.85	0.72
  1334.79	0.66
  1335.82	0.74
  1342.46	0.71
  1342.89	0.74
  1344.65	0.73
  1345.53	0.66
  1346.66	0.67
  1347.57	0.64
  1348.62	0.64
  1348.91	0.73
  1349.23	0.74
  1349.59	0.75
  1350.09	0.74
  1351.03	0.69
  1358.39	0.73
  1359.40	0.71
  1359.72	0.74
  1360.73	0.74
  1361.06	0.74
  1361.33	0.74
  1361.68	0.74
  1361.96	0.73
  1363.45	0.73
  1364.30	0.58
  1364.66	0.73
  1365.00	0.73
  1365.28	0.74
  1365.61	0.74
  1365.92	0.74
  1366.84	0.68
  1367.14	0.74
  1368.17	0.66
  1369.87	0.74
  1385.02	0.73
  1389.39	0.74
  1389.85	0.73
  1391.59	0.71
  1392.48	0.61
  1393.92	0.74
  1395.07	0.73
  1395.35	0.70
  1395.66	0.72
  1402.28	0.72
  1404.02	0.75
  1404.29	0.71
  1404.82	0.74
  1405.09	0.72
  1410.03	0.74
  1412.84	0.65
  1415.51	0.74
  1415.87	0.73
  1416.81	0.69
  1417.25	0.74
  1418.16	0.66
  1418.49	0.74
  1418.95	0.74
  1420.48	0.74
  1420.87	0.74
  1421.19	0.74
  1423.03	0.74
  1423.41	0.74
  1425.37	0.74
  1425.71	0.74
  1426.20	0.74
  1426.56	0.74
  1440.57	0.74
  1441.63	0.74
  1441.94	0.74
  1442.88	0.66
  1443.18	0.73
  1444.80	0.74
  1445.71	0.69
  1446.05	0.74
  1446.30	0.73
  1447.17	0.58
  1449.38	0.74
  1450.27	0.55
  1452.52	0.74
  1452.86	0.74
  1453.13	0.72
  1487.32	0.74
  1488.49	0.68
  1488.79	0.74
  1494.31	0.73
  1494.57	0.70
  1495.80	0.74
  1496.20	0.74
  1496.72	0.74
  1497.00	0.71
  1499.31	0.74
  1499.58	0.72
  1499.89	0.72
  1500.28	0.74
  1500.64	0.74
  1503.44	0.74
  1503.72	0.72
  1505.95	0.73
  1506.21	0.74
  1507.23	0.62
  1508.18	0.73
  1508.85	0.74
  1510.30	0.74
  1511.25	0.68
  1511.79	0.74
  1512.12	0.74
  1512.56	0.74
  1512.83	0.70
  1513.10	0.74
  1513.44	0.74
  1514.39	0.67
  1514.66	0.72
  1515.68	0.65
  1516.69	0.69
  1516.98	0.73
  1517.49	0.74
  1517.83	0.74
  1518.76	0.66
  1519.75	0.71
  1521.36	0.74
  1523.41	0.74
  1523.70	0.74
  1524.55	0.62
  1525.48	0.64
  1525.81	0.74
  1526.22	0.74
  1526.53	0.73
  1527.03	0.74
  1528.76	0.74
  1529.68	0.69
  1530.12	0.74
  1531.08	0.60
  1531.41	0.74
  1550.05	0.74
  1615.55	0.73
  1615.89	0.74
  1616.77	0.63
  1618.94	0.74
  1620.02	0.65
  1622.35	0.74
  1623.81	0.74
  1624.72	0.74
  1625.96	0.74
  1627.16	0.73
  1628.43	0.72
  1628.82	0.74
  1629.75	0.59
  1630.06	0.74
  1630.47	0.74
  1630.77	0.74
  1631.05	0.70
  1633.05	0.74
  1633.40	0.73
  1634.56	0.70
  1636.72	0.74
  1639.34	0.74
  1639.80	0.74
  1641.07	0.74
  1641.52	0.74
  1642.79	0.69
  1643.93	0.68
  1644.54	0.74
  1645.76	0.69
  1646.05	0.66
  1647.14	0.63
  1654.09	0.71
  1663.72	0.74
  1666.57	0.74
  1670.68	0.75
  1671.09	0.74
  1672.75	0.73
  1675.40	0.74
  1675.89	0.74
  1678.65	0.74
  1681.08	0.74
  1681.72	0.74
  1682.07	0.74
  1685.62	0.74
  1687.31	0.74
  1689.74	0.69
  1691.85	0.72
  1694.00	0.74
  1695.99	0.74
  1696.37	0.74
  1697.54	0.66
  1697.89	0.74
  1699.91	0.75
  1702.61	0.74
  1703.06	0.70
  1704.26	0.64
  1704.70	0.73
  1705.10	0.70
  1705.48	0.74
  1705.95	0.74
  1707.27	0.62
  1708.00	0.74
  1708.43	0.74
  1709.00	0.74
  1709.60	0.74
  1712.15	0.74
  1713.87	0.74
  1716.16	0.74
  1716.61	0.73
  1717.16	0.74
  1749.59	0.73
  1750.25	0.74
  1750.55	0.72
  1760.70	0.73
  1762.74	0.74
  1764.86	0.74
  1765.32	0.74
  1766.60	0.67
  1767.05	0.72
  1767.48	0.74
  1769.13	0.74
  1770.29	0.63
  1772.44	0.64
  1772.80	0.74
  1774.05	0.60
  1775.29	0.66
  1775.95	0.74
  1776.33	0.73
  1776.78	0.72
  1777.17	0.72
  1779.60	0.74
  1782.11	0.73
  1783.26	0.59
  1784.75	0.74
  1785.92	0.68
  1788.76	0.73
  1789.15	0.74
  1789.60	0.75
  1791.03	0.70
  1792.92	0.74
  1793.25	0.70
  1794.47	0.66
  1794.90	0.74
  1795.93	0.68
  1797.67	0.74
  1800.36	0.74
  1804.15	0.74
  1805.32	0.65
  1806.84	0.72
  1815.13	0.74
  1815.42	0.74
  1816.46	0.60
  1816.86	0.74
  1817.33	0.74
  1818.37	0.57
  1819.67	0.68
  1821.24	0.74
  1959.94	0.74
  1962.05	0.74
  1962.52	0.74
  1963.65	0.70
  1965.12	0.69
  1965.71	0.74
  1966.86	0.69
  1967.21	0.70
  1967.70	0.73
  1969.77	0.74
  1971.83	0.74
  1973.21	0.64
  1974.48	0.69
  1974.91	0.74
  1975.36	0.74
  1977.24	0.69
  1977.84	0.74
  1978.47	0.74
  1978.82	0.70
  1979.98	0.65
  1981.16	0.69
  1981.84	0.74
  1982.23	0.74
  1983.38	0.66
  1983.84	0.73
  1984.98	0.72
  1986.31	0.74
  };
  % \addplot[draw=black!50!vintageorange,ultra thick]
  % table[x=x,y=y] {
  % x y
  % 0.51  0.70
  % 1.03  0.74
  % 386.84	0.74
  % 387.40	0.75
  % 3600 0.75
  % };

  % \addlegendentry{\system's Models}

  \addplot[mark=*,mark size=1.5,mark options={fill=vintagegreen,draw=vintagegreen},
  draw=none]
  table[x=x,y=y] {
  x y
  243.57	0.71
  816.51	0.74
  1015.29	0.72
  2008.85	0.74
  2527.32	0.74
  3106.12	0.74
  3186.63	0.59
  3650.85	0.74
  3844.85	0.71
  };

  \end{axis}
\end{tikzpicture}

\vspace{-4mm}
\caption{Hyperparameter optimization}
\label{fig:exp:hyperparam}
\end{figure}

%% file: figures/fig_exp_model_complexity.tex
\begin{figure}[t]

  \pgfplotsset{modelcomp/.style={
    width=45mm,
    height=30mm,
    xmin=0.5,
    xmax=6.5,
    ymin=0,
    ymax=500000,
    xlabel=Regularization Coefficient,
    ylabel=Est.~Sample Size,
    xlabel near ticks,
    ylabel near ticks,
    ylabel style={align=center},
    xtick={1,2,3,4,5,6},
    xticklabels={0, $10^{\minus 4}$, $10^{\minus 3}$, $10^{\minus 2}$, $10^{\minus 1}$, 1},
    xtick style={draw=none},
    ytick={0, 100000, 200000, 300000, 400000, 500000},
    yticklabels={0, 100K, 200K, 300K, 400K, 500K},
    ylabel shift=-3pt,
    xlabel shift=-2pt,
    legend style={
        at={(1.05,1)},anchor=north west,column sep=2pt,
        draw=black,fill=white,line width=.5pt,
        /tikz/every even column/.append style={column sep=10pt},
        font=\scriptsize,
    },
    legend columns=1,
    legend cell align={left},
    every axis/.append style={font=\footnotesize},
    ymajorgrids=false,
    % yminorgrids,
    minor grid style=lightgray,
    xticklabel style={yshift=2mm},
    scaled y ticks=false,
    % nodes near coords,
    xticklabel style={rotate=30,yshift=-2mm,anchor=east,xshift=1.5mm},
  }}

\centering
\begin{subfigure}[b]{0.48\linewidth}
  \begin{tikzpicture}
    \begin{axis}[
      modelcomp,
      ybar,
      bar width=1.5mm,
    ]

    \addplot[fill=vintageblack]
    table[x=x,y=y] {
    x y
    1 472031.9
    2 418375.2
    3 217667.9
    4 62399.6
    5 30303
    6 26048.9
    };

    \end{axis}
  \end{tikzpicture}

  \vspace{-2mm}
  \caption{Regularization}
  \label{fig:exp:model:a}
\end{subfigure}
\hfill
\begin{subfigure}[b]{0.48\linewidth}
  \begin{tikzpicture}
    \begin{axis}[
      modelcomp,
      ybar,
      bar width=1.5mm,
      xmin=0.5,
      xmax=7.5,
      xlabel=Number of Params,
      xtick={1,2,3,4,5,6,7},
      xticklabels={100, 500, 1K, 5K, 10K, 50K, 100K},
      xtick style={draw=none},
      ymin=0,
      ymax=200000,
      ytick={0, 50000, 100000, 150000, 200000},
      yticklabels={0, 50K, 100K, 150K, 20K},
    ]

    \addplot[fill=vintageblack]
    table[x=x,y=y] {
    x y
    1 24354
    2 92338
    3 128927
    4 174036
    5 167884
    6 157632
    7 164168
    };

    \end{axis}
  \end{tikzpicture}

  \vspace{-2mm}
  \caption{Number of Params}
  \label{fig:exp:model:b}
\end{subfigure}

\vspace{-2mm}
\caption{Model complexity vs.~estimated sample sizes}
\label{fig:exp:model}
\end{figure}
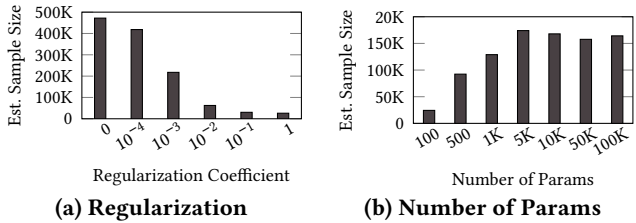

%% file: related.tex
%!TEX root = approx_ml.tex

\input{figures/tab_related}

\section{Related Work}
\label{sec:related}

% System for ML optimization: \cite{kaoudi2017cost}

In this section, we first discuss three types of work closely related to \system:
Sampling for ML,
Hyperparameter Optimization, and
Feature Selection.
We summarize their key contributions in \cref{tab:related}.

Next, we overview
the approaches inspired by familiar database techniques (DBMS-Inspired Optimization) and
the recent advances in statistical optimization methods (Faster Optimization Algorithms).

% \input{figures/fig_exp_model_complexity}

%  ML systems inspired by familiar database techniques (DBMS-Inspired Optimization),
%   ML systems employing different forms of approximation (Approximate ML Systems),
%   	and the recent advances in statistical optimization methods (Faster Optimization Algorithms).
% The developments in the last category are mostly orthogonal to \system (or ML systems in general).
%  However, ML systems can still benefit from employing these developments.

\ph{Sampling for ML}
Sampling for ML has been extensively studied in the literature. 
Typically, the sampling is optimized for a specific
type of model, such as linear regression~\cite{feldman2011unified,bachem2017practical,ghashami2014relative,cohen2015uniform,derezinski2017unbiased,drineas2011faster,drineas2012fast,drineas2006sampling,bhojanapalli2015tighter}, logistic regression~\cite{wang2018optimal,krishnapuram2005sparse}, clustering~\cite{feldman2013turning,jaiswal2014simple}, Kernel matrices~\cite{gittens2016revisiting,musco2017recursive}, point processes~\cite{li2015efficient}, and Gaussian mixture model~\cite{lucic2017training}.
Woodruf~\cite{woodruff2014sketching} overviews sketching techniques for least squares and low rank approximation.
Zombie~\cite{anderson2016input} employs a clustering-based active learning technique for training approximate models, but does not offer any error guarantees.
In contrast, 
% \tofix{by developing efficient techniques for applying MLE theories~\cite{fisher1925theory,efron1978assessing}}, 
% \barzan{not english} \yongjoo{updated}
\system offers probabilistic error guarantees for any MLE-based model.
Cohen~\cite{cohen2015uniform} studies uniform sampling for ML, but for only linear regression.
Bottou~\cite{bottou2008tradeoffs} and Shalev-Shwartz~\cite{shalev2008svm} study the optimization errors.

\ph{Hyperparameter Optimization}
Automatic hyperparameter optimization (i.e., AutoML) 
\barzan{is AutoML a well-known term?? answer: mentioned by reviewer}
has been pursued by AutoWeka~\cite{thornton2013auto}, auto-sklearn~\cite{feurer2015efficient}, and Google's Cloud AutoML~\cite{google_automl}.
These techniques have typically taken 
a Bayesian approach to hyperparameter optimization~\cite{shahriari2016taking}. The 
ML community has also studied how to optimize deep neural networks~\cite{bergstra2011algorithms,bergstra2013making,maclaurin2015gradient}.
What \system offers is orthogonal to these hyperparameter optimization methods, since \system speeds up the individual model training via approximation, while hyperparameter optimization methods focus on finding an optimal sequence of hyperparameters to test.

\ph{Feature Selection}
Choosing a relevant set of features is an important task in ML~\cite{guyon2003introduction,boyce2013optimal}. Feature selection can either be performed before training a final model~\cite{john1994irrelevant} or during model training relying on sparsity-inducing models~\cite{ng2004feature,yang2007feature}.
% \barzan{reword} \yongjoo{updated}
One approach to accelerate feature selection 
is preprocessing the data in advance~\cite{zhang2014materialization}.
% \barzan{what does this mean?} \yongjoo{updated}
\system can  be used in conjunction with these methods.

% Perhaps the most related to \system is Columbus~\cite{zhang2014materialization}, which
% 	 trains approximate models on a coreset, a representative subset of the training data.
% Although Columbus also uses a sample of the data to speed up training,
% 	they can only provide an accuracy guarantee for ordinary least squares methods.
% In contrast, \system supports a much wider class of problems; in fact, the classification methods and the factor analysis technique used in our experiments   include any non-linear objective functions, none of
% which can be approximated with guarantees by Columbus.

% MacroBase~\cite{bailis2017macrobase} develops several ML methods for speeding up prediction performance, which
% There are other methods that speed up the prediction time at the cost of increased preprocessing.
% NoScope \cite{kang2017noscope} first trains multiple deep neural networks and then composes a different cascade depending on the target accuracy. tKDC \cite{gan2017scalable} first builds a $k$-d tree and then conducts early pruning at the prediction stage of a kernel-density classifier.
% \system differs from these methods by focusing on speeding up the training phase, which is a crucial bottleneck in   early stages of model tuning.

% Next, we also overview the recent developments in non-linear optimizations proposed by the machine learning and the hardware community.
% These techniques are orthogonal to \system; \system's training time can be further shortened by incorporating those ideas.

\ph{DBMS-Inspired Optimization}
A salient feature of database systems is their declarative interface (SQL) and
the resulting optimization opportunities.
These ideas have been applied to speed up ML workloads.
SystemML~\cite{ghoting2011systemml,boehm2016systemml,elgohary2016compressed}
ScalOps~\cite{weimer2011machine},
Pig latin~\cite{olston2008pig},
and KeystoneML~\cite{sparks2017keystoneml}
propose
high-level ML languages for automatic parallelization and materialization, as well as easier programming.
Hamlet~\cite{kumar2016join} and others~\cite{kumar2015learning,schleich2016learning} avoid expensive denormalizations.
Hemingway~\cite{pan2017hemingway}, MLBase~\cite{kraska2013mlbase}, and TuPAQ~\cite{sparks2015automating} automatically choose an optimal plan for a given ML workload.
SciDB~\cite{stonebraker2013scidb,kersten2011sciql}, MADLib~\cite{cohen2009mad,hellerstein2012madlib}, and RIOT~\cite{zhang2010efficient} exploit in-database computing.
Kumar et al.~\cite{kumar2016model} uses a model selection management system to unify feature engineering \cite{anderson2013brainwash}, algorithm selection, and parameter tuning.
NoScope~\cite{kang2017noscope} and tKDC~\cite{gan2017scalable} speed up the prediction at the cost of more preprocessing.

\ph{Faster Optimization Algorithms}
Advances in optimization algorithms are largely orthogonal to ML systems; however, understanding their benefits is necessary for developing faster ML systems. Recent advances can be categorized into
software- and hardware-based approaches.

Software-based methods are mostly focused on improving gradient descent variants (e.g., SGD),
	where a key question is how to adjust the step size in order to accelerate the convergence rate
	towards the optimal solution~\cite{shallue2018measuring}. 
	Recent advances include adaptive rules for accelerating this  rate, e.g., Adagrad \cite{duchi2011adaptive}, Adadelta \cite{zeiler2012adadelta}, RMSprop \cite{tieleman2012lecture}, and Adam \cite{kingma2014adam}.
	Hogwild!~\cite{recht2011hogwild} and related techniques~\cite{zhang2013asynchronous,hadjis2016omnivore,gonzalez2015asynchronous} disable locks to speed up SGD via asynchronous updates.
	There is also recent work on rediscovering the benefits of quasi-Newton optimization methods, e.g., showing that minibatch variants of quasi-Newton methods (such as L-BFGS or CG) can be superior to SGD due to their higher parallelism~\cite{le2011optimization}.

 Hardware-based   techniques   speed up training by relaxing strict  precision requirements,
 e.g., DimmWitted~\cite{zhang2014dimmwitted} and BuckWild!~\cite{de2017understanding}.

%% file: figures/tab_related.tex
\begin{table*}[t]
\caption{Previous work on Sampling for ML / Hyperparameter Optimization / Feature Selection}
\label{tab:related}

% \vspace{-2mm}

\small
% \fontsize{8.5}{9}\selectfont
\renewcommand{\arraystretch}{1.2}
\begin{tabular}{p{23mm} p{43mm} p{92mm}}
  \toprule
    & \textbf{Approach} & \textbf{Key Contributions} \\
  \midrule
  \multirow{2}{*}{\parbox{23mm}{Sampling\\ for ML}}
    & Coreset (LR)~\cite{feldman2011unified,bachem2017practical,ghashami2014relative}
      & Proposes a coreset (non-uniform sample) for linear regression \\
    & Cohen~\cite{cohen2015uniform}
      & Approximates linear regression with uniform random sampling \\
    & Derezinski~\cite{derezinski2017unbiased}
      & Proposes \emph{volume sampling} for linear regression \\
    & Zombie~\cite{anderson2016input}
      & Applies active learning to sampling \\
    & Drineas~\cite{drineas2011faster,drineas2012fast,drineas2006sampling}, \newline Bhojanapalli~\cite{bhojanapalli2015tighter}
      & Develops non-uniform sampling (based on leverage scores) for linear regression  \\
    & Wang~\cite{wang2018optimal}, Krishnapuram~\cite{krishnapuram2005sparse}
      & Develops non-uniform sampling for logistic regression \\
    & Coreset (Clustering)~\cite{feldman2013turning,jaiswal2014simple}
      & Proposes a coreset (non-uniform sample) for clustering \\
    & Gittens~\cite{gittens2016revisiting}, Musco~\cite{musco2017recursive}
      & Approximates kernel matrices (i.e., matrices containing inner products) \\
    & Coreset (GMM)~\cite{lucic2017training}
      & Proposes a coreset (non-uniform sample) for Gaussian mixture models \\
    & Li~\cite{li2015efficient}
      & Proposes sampling for point processes \\
    & SafeScreening~\cite{ogawa2013safe}
      & Removes non-support vectors prior to training a SVM model \\
    % & Fisher~\cite{fisher1925theory}, Efron~\cite{efron1978assessing}
    %   & Assesses the accuracy of MLE models trained in a uniform random sample \\
    & \textbf{\system (ours)}
      & Develops efficient algorithms for training MLE models with error guarantees \\
  \midrule
  \multirow{2}{*}{\parbox{23mm}{Hyperparameter \\ Optimization}}
    & AutoWeka~\cite{thornton2013auto},
      auto-sklearn~\cite{feurer2015efficient}, \newline
      Cloud AutoML~\cite{google_automl}
      & Applies a wide range of feature selection \& hyperparameter optimization techniques to existing ML frameworks/libraries \\
    & Bergstra~\cite{bergstra2011algorithms},
      Bergstra~\cite{bergstra2013making}
      & Optimizes hyperparameter optimization for deep neural network \\
    & Maclaurin~\cite{maclaurin2015gradient}
      & Proposes gradient-based hypeparameter optimization \\
    & Shahriari~\cite{shahriari2016taking}
      & Reviews Bayesian approach to hyperparameter optimization \\
    & Bardenet~\cite{bardenet2013collaborative}
      & Proposes a colloborative hyperparameter tuning among similar tasks \\
    & TuPAQ~\cite{sparks2015automating}
      & Provides a systematic support for hyperparameter optimization \\
  \midrule
  \multirow{2}{*}{\parbox{23mm}{Feature \newline Selection}}
    & Guyon~\cite{guyon2003introduction}, Boyce~\cite{boyce2013optimal}, John~\cite{john1994irrelevant}
      & Introduces concepts related to feature selection \\
    & Ng~\cite{ng2004feature}, Yang~\cite{yang2007feature}
      & Performs feature selection via sparse models \\
    & AutoWeka~\cite{thornton2013auto}
      & See the description above \\
    & Columnbus~\cite{zhang2014materialization}
      & Optimizes feature selection by materialization \\
  \bottomrule
\end{tabular}

\end{table*}

%% file: conclusion.tex
\section{Conclusion}
\label{sec:conclusion}

In this work, we have developed \system,
an approximate machine learning system with probabilistic guarantees
for MLE-based models.
% approximate machine learning system
\system uses sampling 
to dramatically reduce time and computational costs,
which is particularly beneficial in
% which benefit 
the early stages of model tuning.
% during the early stages of model tuning.
% to train an approximate model in accordance to user's error tolerance,
%
% \system's techniques are applicable to a wide class of ML models, i.e., any algorithm that relies on maximum likelihood estimation for their training.
% The MLE models include
% Generalized Linear Models (e.g., linear regression (Lin), logistic regression (LR), max entropy classifier (ME), Poisson regression, etc.) and Probabilistic Principal Component Analysis (PPCA). 
Through an extensive set of experiments on several large-scale, real-world datasets, we showed that \system produced 95\% accurate models of linear regression, logistic regression, max entropy classifier, and Probabilistic Principal Component Analysis, while using only 0.16\%--15.96\% of the time needed for training the full model.
Our future plan is 
% to extend \system's capabilities
% \tofix{to more complex MLE-based models such as ,
% \barzan{this contradicts ur claim that what u have presented already applies to ANY MLE-based model!} \yongjoo{moved to down}
to extend
beyond the maximum likelihood estimation models, such as decision trees, Gaussian Process regression, Na\"ive Bayes classifiers, and Deep Boltzmann Machines~\cite{salakhutdinov2010efficient}.
We also plan to open-source \system, with wrappers for various popular ML libraries, including scikit-learn (Python), glm (R), and MLlib.
Finally, given that \system's underlying techniques are extensible to non-uniform sampling, we plan to further 
explore task- and model-specific non-uniform sampling strategies.
% Finally, to scale to extremely high-dimensional data (i.e., with millions of features),
% 	we plan to extend our   approach  to  avoid explicit materialization of $H$ and $J$.

\section{Acknowledgement}
\label{sec:ack}

This research is in part supported by National Science Foundation
through grants 1553169 and 1629397.

%% file: abstraction_example.tex
\section{Model Abstraction Examples}
\label{sec:model:abstract:example}

This section shows that \system-supported ML models can be cast into the abstract form in \cref{eq:support:obj2}. For illustration, we use logistic regression and PPCA.

\ph{Logistic Regression}
The objective function of logistic regression captures the difference between true class labels and the predicted class labels. An optional regularization term may be placed to prevent a model from being overfitted to a training set. For instance, the objective function of L2-regularized logistic regression is expressed as follows:
\[
f_n(\theta) = \minus \left[
\frac{1}{n} \sum_{i=1}^n t_i \log \sigma(\theta^\top \xx_i)
+ (1 - t_i) \log \sigma(1 - \theta^\top \xx_i)
\right]
+ \frac{\beta}{2} \| \theta \|^2
\]
where $\sigma(y) = 1 / (1 + \exp(y))$ is a sigmoid function, and $\beta$ is the coefficient that controls the strength of the regularization penalty. The observed class labels, i.e., $t_i$ for $i = 1, \ldots, n$, are either 0 or 1.
The above expression is minimized when $\theta$ is set to the value at which the gradient $\nabla f_n(\theta)$ of $f_n(\theta)$ becomes a zero vector; that is,
\[
\nabla f_n(\theta) = \left[ \frac{1}{n}
\sum_{i=1}^n (\sigma(\theta^\top \xx_i) - t_i)\, \xx_i
\right] + \beta\, \theta = \bm{0}
\]
It is straightforward to cast the above expression into \cref{eq:abstract:model}. That is,
$q(\theta; \xx_i, t_i) = (\sigma(\theta^\top \xx_i) - t_i)\, \xx_i$
and $r(\theta) = \beta\, \theta$.
% \cref{code:model:specification} shows an example of MCS for logistic regression.

\ph{PPCA}
The objective function of PPCA captures the difference between the covariance matrix $S$ of the training set and the covariance matrix $C = \Theta\, \Theta^\top + \sigma^2 I$ reconstructed from the $q$ number of extracted factors $\Theta$, as follows:
\[
f_n(\Theta) = \frac{1}{2} \left(
  d \log 2 \pi + \log |C| + \tr(C^{\minus 1} S)
\right)
\]
where $d$ is the dimension of feature vectors. $\Theta$ is a $d$-by-$q$ matrix in which each column represents a factor, and $\sigma$ is a real-valued scalar that represents the noise in data not explained by those factors. The optimal value for $\sigma$ can be obtained once the values for $\Theta$ are determined. The value of $q$, or the number of the factors to extract, is a user parameter. The above expression $f_n(\Theta)$ is minimized when its gradient $\nabla_\Theta f_n(\Theta)$ becomes a zero vector; that is,
\begin{align*}
\nabla_\Theta f(\Theta) &= C^{\minus1} (\Theta - S\, C^{\minus 1} \Theta) = \bm{0}
\end{align*}
The above expression can be cast into the form in \cref{eq:abstract:model} by observing $S = (1/n) \sum_{i=1}^n \xx_i \xx_i^\top$.\footnote{This sample covariance expression assumes that the training set is zero-centered.} That is, $q(\Theta; \xx_i) = C^{\minus1} \Theta - C^{\minus1} \xx_i \xx_i^\top C^{\minus1} \Theta$ and $r(\Theta) = 0$. $t_i$ is omitted on purpose since PPCA does not need observations (e.g., class labels). Although we used a matrix form $\Theta$ in the above expression for simplicity, \system internally uses a vector $\theta$ when passing parameters among components. The vector is simply flattened and unflattened as needed.

%% file: proofs.tex
\section{Deferred Proofs}
\label{sec:proofs}

\subsection{Generalization Error}

\begin{proof}[\textbf{Proof of \cref{lemma:gen}}]
  \system's error guarantee
%   i.e., $v(m_n) \le \varepsilon$, 
  is probabilistic because
  of the random sampling process for obtaining $D_n$.
%   the uniform random sample of the dataset $D$ 
  This proof exploits the fact that
  this random sampling process
  is independent of the distribution of $(\xx, y) \sim \mathcal{D}$.
  
  The generalization error of the full model
%   , i.e.,
%   $E_{(\xx,y) \sim \mathcal{D}} (\1{m_N(\xx) \ne y})$,
  can be bounded as follows:
  \begin{align*}
    &E_{(\xx,y) \sim \mathcal{D}} (\1{m_N(\xx) \ne y})
    = \int \1{m_N(\xx) \ne y} \, p(\xx)\, d\xx \\
    &= \int ( m_n(\xx) \ne y \land m_n(\xx) = m_N(\xx) ) \\
    &\qquad   \lor ( m_n(\xx) = y \land m_n(\xx) \ne m_N(\xx) ) \, p(\xx)\, d\xx \\
    &\le \int (m_n(\xx) \ne y) \lor ( m_n(\xx) = y \land m_n(\xx) \ne m_N(\xx) ) \, p(\xx)\, d\xx \\
    &\le \int (m_n(\xx) \ne y) \, p(\xx)\, d\xx \\
    &\quad  + \int ( m_n(\xx) = y \land m_n(\xx) \ne m_N(\xx) ) \, p(\xx)\, d\xx
  \end{align*}

  By definition, $\int (m_n(\xx) \ne y) \, p(\xx)\, d\xx = \varepsilon_g$.
  Based on the independence we stated above and $\pr( E_{\xx}(m_n(\xx) \ne m_N(\xx)) \le \varepsilon) \ge 1 - \delta$,
  \begin{align*}
    \int ( m_n(\xx) = y \land m_n(\xx) \ne m_N(\xx) ) \, p(\xx)\, d\xx
    \le (1 - \varepsilon_g) \cdot \varepsilon
  \end{align*}
  with probability at least $1 - \delta$.
  Thus,
  \[
  E_{(\xx,y) \sim \mathcal{D}} (\1{m_N(\xx) \ne y})
  \le \varepsilon_g + \varepsilon - \varepsilon_g \cdot \varepsilon
  \]
  with probability at least $1 - \delta$.
\end{proof}

\subsection{Model Parameter Distribution}

\begin{proof}[\textbf{Proof of \cref{thm:param_dist}}]
We first derive the distribution of $\hat{\theta}_n - \theta_\infty$, which will then be used to derive the distribution of $\hat{\theta}_n - \hat{\theta}_N$.
Our derivation is the generalization of the result in \cite{newey1994large}. The generalization is required since the original result does not include $r(\theta)$.

Let $\theta_\infty$ be the parameter values at which $g_\infty(\theta)$ becomes zero. Since the size of the training set is only $N$, $\theta_\infty$ exists only conceptually.
Since $\theta_n$ is the optimal parameter values, it satisfies $g_n(\theta_n) = \bm{0}$. According to the mean-value theorem, there exists $\bar{\theta}$ between $\theta_n$ and $\theta_\infty$ that satisfies:
\[
H(\bar{\theta}) (\theta_n - \theta_\infty) = g_n(\theta_n) - g_n(\theta_\infty)
 = \minus g_n(\theta_\infty)
\]
where $H(\bar{\theta})$ is the Jacobian of $g_n(\theta)$ evaluated at $\bar{\theta}$.
Note that $g_n(\theta_n)$ is zero since $\theta_n$ is obtained by finding the parameter at which $g_n(\theta)$ becomes zero.

Applying the multidimensional central limit theorem to the above equation produces the following:
\begin{align}
&\sqrt{n}\left( \hat{\theta}_n - \theta_\infty \right)
= - H(\bar{\theta})^{\minus1} \; \sqrt{n} \; g_n(\theta_\infty) \nonumber \\
&= - H(\bar{\theta})^{\minus1} \;
\frac{1}{\sqrt{n}}  \sum_{i=1}^n  \left(
  q(\theta_\infty; \xx_i, y_i) + r(\theta_\infty)
\right)  \label{eq:param_dist1} \\
&\xrightarrow{n \rightarrow \infty}
\mathcal{N}( \bm{0},\, H^{\minus1} J H^{\minus1})
\label{eq:param_dist}
\end{align}
where $H$ is a shorthand notation of $H(\bar{\theta})$.
To make a transition from \cref{eq:param_dist1} to \cref{eq:param_dist}, an important relationship called the \emph{information matrix equality} is used. According to the information matrix equality, the covariance of $q(\theta; \xx_i, y_i)$ is equal to the Hessian of the negative log-likelihood expression, which is equal to $J$.
% The asymptotic distribution of $\hat{\theta}_n - \theta_\infty$ is, therefore, $\mathcal{N}(\bm{0}, (1/n) H^{\minus1} J H^{\minus1})$.

Now, we derive the distribution of $\hat{\theta}_n - \hat{\theta}_N$.
We use the fact that $\hat{\theta}_N$ is the optimal parameter for $D_N$, which is a union of $D_n$ and $D_N - D_n$, where $\hat{\theta}_n$ is the optimal parameter for $D_n$. To separately capture the randomness stemming from $D_n$ and $D_N - D_n$, we introduce two random variables $X_1, X_2$ that independently follow $\mathcal{N}( \bm{0},\, H^{\minus1} J H^{\minus1} )$. From \cref{eq:param_dist}, $\hat{\theta}_n - \theta_\infty \rightarrow (1/\sqrt{n})\, X_1$. Also, let $q_i = q(\theta_\infty; \xx_i, y_i)$ for simplicity; then,
\begin{align*}
\sqrt{N} & \left( \hat{\theta}_N - \theta_\infty \right)
= - H^{\minus1} \frac{1}{\sqrt{N}} ( \sum_{i=1}^n q_i + \sum_{i=1}^{N-n} q_i ) \\
&= - H^{\minus1} \left[
     \frac{\sqrt{n}}{\sqrt{N}} \frac{1}{\sqrt{n}} \sum_{i=1}^n q_i
     + \frac{\sqrt{N-n}}{\sqrt{N}} \frac{1}{\sqrt{N-n}} \sum_{i=1}^{N-n} q_i
\right] \\
&\xrightarrow{n \rightarrow \infty \text{ and } N \rightarrow \infty}
\frac{\sqrt{n}}{\sqrt{N}} X_1 + \frac{\sqrt{N-n}}{\sqrt{N}} X_2
\end{align*}
Since $\hat{\theta}_n - \hat{\theta}_N = (\hat{\theta}_n - \theta_\infty) - (\hat{\theta}_N - \theta_\infty)$,
and the limit of each of $(\hat{\theta}_n - \theta_\infty)$
and $(\hat{\theta}_N - \theta_\infty)$ exists,
\begin{align*}
\hat{\theta}_n - \hat{\theta}_N
&\rightarrow \frac{1}{\sqrt{n}} X_1 - \frac{\sqrt{n}}{N} X_1 - \frac{\sqrt{N-n}}{N} X_2 \\
&= \left( \frac{1}{\sqrt{n}} - \frac{\sqrt{n}}{N} \right) X_1
   - \frac{\sqrt{N-n}}{N} X_2
\end{align*}
Note that $\hat{\theta}_n - \hat{\theta}_N$ asymptotically follows a normal distribution, since it is a linear combination of two random variables that independently follow normal distributions. Thus,
\begin{align*}
& \hat{\theta}_n - \hat{\theta}_N \xrightarrow{n \rightarrow \infty \text{ and } N \rightarrow \infty} \\
& \mathcal{N}\left( \bm{0}, \, \left( \frac{1}{\sqrt{n}} - \frac{\sqrt{n}}{N} \right)^2 H^{\minus1} J H^{\minus1} + \left( \frac{\sqrt{N-n}}{N} \right)^2 H^{\minus1} J H^{\minus1} \right) \\
&= \mathcal{N}\left( \bm{0}, \, \left( \frac{1}{n} - \frac{1}{N} \right) H^{\minus1} J H^{\minus1} \right)
\qedhere
\end{align*}
% the variance of $\hat{\theta}_n - \hat{\theta}_N$ is $ = \left(\frac{1}{n} - \frac{1}{N} \right) H^{\minus1} J H^{\minus1}$.
\end{proof}

\begin{proof}[\textbf{Proof of \cref{thm:cond_param_dist}}]
Observe that $\hat{\theta}_n - \hat{\theta}_N$ and $\hat{\theta}_N - \theta_\infty$ are independent because they are jointly normally distributed and the covariance between them is zero, as shown below:
\begin{align*}
&\cov( \hat{\theta}_n - \hat{\theta}_N,\, \hat{\theta}_N - \theta_\infty ) \\
&= \frac{1}{2} \left( \var(\hat{\theta}_n - \hat{\theta}_N + \hat{\theta}_N - \theta_\infty)
- \var(\hat{\theta}_n - \hat{\theta}_N) - \var(\hat{\theta}_N - \theta_\infty) \right) \\
&= \frac{1}{2}
   \left( \frac{1}{n} - \left( \frac{1}{n} - \frac{1}{N} \right) - \frac{1}{N} \right)
   \, H^{\minus1} J H^{\minus1} = \bm{0}
\end{align*}
Thus, $\var(\hat{\theta}_n - \theta_N) = \var(\hat{\theta}_n - \hat{\theta}_N \mid \theta_N) = \alpha \, H^{\minus1} J H^{\minus1}$, which implies
\begin{equation}
\hat{\theta}_n \sim
 \left( \theta_N,\,
 \alpha \, H^{\minus1} J H^{\minus1} \right)
\label{eq:cond:a}
\end{equation}
Using Bayes' theorem,
\[
\pr(\theta_N \mid \theta_n) = (1/Z)\, \pr(\theta_n \mid \theta_N)\, \pr(\theta_N)
\]
for some normalization constant $Z$. Since there is no preference on $\pr(\theta_N)$, we set a constant to $\pr(\theta_N)$. Then, from \cref{eq:cond:a}, $\hat{\theta}_N \mid \theta_n \sim \mathcal{N}(\theta_n, \alpha\, H^{\minus 1} J H^{\minus 1})$.
\end{proof}

\begin{proof}[\textbf{Proof of \cref{lemma:err}}]
  We first define three events $A$, $B$, and $C$ as follows:
\begin{align*}
&A: \quad v(m_n) \le \varepsilon \\
&B: \quad
\int \1{ v(m_n) \le \varepsilon_1 } \, h(\theta_N)\; d \theta_N
  \ge \frac{1 - \delta}{0.95} \\
&C: \quad
\frac{1}{k} \sum_{i=1}^k \1{ v(m_n; \theta_{N,i}) \le \varepsilon_1 }
  = \frac{1 - \delta}{0.95} + \sqrt{\frac{\log 0.95 }{\minus 2 k}}
\end{align*}
It suffices to show that $P(A \mid C) \ge 1 - \delta$, for which we use the relationship
$P(A \mid C) = P(A \mid B) \; P(B \mid C)$.

First, by the basic relationship between the probability and the indicator function,
$P(A \mid B) = (1 - \delta) / 0.95$.

Second, by the Hoeffding's inequality,
\begin{align*}
&\pr \left[ a \ge b - \sqrt{\frac{\log 0.95 }{\minus 2 k}} \right] \ge 0.95 \\
\text{where} \quad
&a = \int \1{ v(m_n; \theta_N) \le \varepsilon } \, h(\theta_N)\; d \theta_N \\
&b = \frac{1}{k} \sum_{i=1}^k \1{ v(m_n; \theta_{N,i}) \le \varepsilon_1 }
\end{align*}
since $b$ is the degree-$k$ U-statistics of $a$.
Thus, $P(B \mid C) \ge$ 0.95.

Therefore,
$P(A \mid C) = P(A \mid B) \; P(B \mid C) \ge (1 - \delta) / 0.95 \times 0.95 = 1 - \delta$.
\end{proof}

\subsection{Sample Size Estimation}

\noindent
\begin{proof}[\textbf{Proof of \cref{thm:decreasing}}]
Without loss of generality, we prove \cref{thm:decreasing} for the case where the dimension of training examples, $d$, is 2. It is straightforward to generalize our proof for the case with an arbitrary $d$.
Also, without loss of generality, we assume $B$ is the box area bounded by $(-1, -1)$ and $(1, 1)$. Then, $p_v(\gamma)$ is expressed as
\[
\int_{(-1,-1)}^{(1,1)}
\frac{1}{\sqrt{(2 \pi)^2 \, |\gamma C|}}
\exp \left( - \theta^\top (\gamma C)^{\minus1} \theta
\right)
\, d\theta
\]

To prove the theorem, it suffices to show that $\gamma_1 < \gamma_2 \Rightarrow p_v(\gamma_1) > p_v(\gamma_2)$ for arbitrary $\gamma_1$ and $\gamma_2$. By definition,
\[
p_v(\gamma_1) =
\int_{-(1, 1)}^{(1,1)}
\frac{1}{\sqrt{(2 \pi)^2 \, |\gamma_1 C|}}
\exp\left(-\theta^\top (\gamma_1 C)^{\minus1}\, \theta \right) \, d\theta
\]
By substituting $\sqrt{\gamma_1/\gamma_2}\, \theta$ for $\theta$,
\begin{align*}
&p_v(\gamma_1) =
\int_{-\sqrt{\gamma_2/\gamma_1}\, (1, 1)}^{\sqrt{\gamma_2/\gamma_1}\, (1,1)}
\frac{1}{\sqrt{(2 \pi)^2 \, |\gamma_2 C|}}
\exp\left(-\theta^\top (\gamma_2 C)^{\minus1}\, \theta \right) \, d\theta \\
&> \int_{-(1, 1)}^{(1,1)}
\frac{1}{\sqrt{(2 \pi)^2 \, |\gamma_2 C|}}
\exp\left(-\theta^\top (\gamma_2 C)^{\minus1}\, \theta \right) \, d\theta = p_v(\gamma_2)
\end{align*}
because the integration range for $p_v(\gamma_1)$ is larger.
\end{proof}

% \else
%
% \subsection{Sample Size Estimation}
%
% As stated early, the proof of \cref{thm:decreasing} is in our extended report~\cite{approxmltech}.
% \fi

%% file: model_sim_def.tex
\section{Model Similarities}
\label{sec:model:sim}

The model difference $v(m_n)$ defined for classification in \cref{sec:interface}
can be extended to regression and unsupervised learning in a straightforward way, as follows. The experiment results in \cref{sec:exp} used these definitions for corresponding models.

\ph{Regression}
For regression, $v(m_n)$ captures the expected prediction difference between $m_n$ and $m_N$.
\[
v(m_n) = \left( \E_{\xx \sim \mathcal{D}}
  [(m_n(\xx) - m_N(\xx))^2] \right)^{1/2}
\]

\ph{Unsupervised Learning}
For unsupervised learning, the model difference captures the difference between the model parameters.
For instance, \system uses the following expression for PPCA:
\[
v(m_n) = 1 - \text{cosine}(\theta_n, \, \theta_N)
\]
where 
$\theta_n$ and $\theta_N$ are the parameters of $m_n$ and $m_N$, respectively,
and
$\text{cosine}(\cdot, \cdot)$ indicates the cosine similarity.

%% file: raw_data.tex
\onecolumn

\pagebreak

\section{Raw Experiment Data}
\label{sec:raw_data}

Find the raw experiment data in the following tables.

\vspace{10mm}

\begin{table}[h]
\caption{Training time savings. This is the raw data for \cref{fig:exp:time_saving}.}
\label{tab:exp:time_saving}

\centering
\small
\begin{tabular}{rrr rrr}
\toprule
\multicolumn{3}{c}{\normalsize \textbf{Lin, \gas}} &
\multicolumn{3}{c}{\normalsize \textbf{Lin, \power}}  \\
\textbf{Requested Accuracy} & \textbf{Training Time} & \textbf{Ratio to Full Training} &
\textbf{Requested Accuracy} & \textbf{Training Time} & \textbf{Ratio} \\
\cmidrule(lr){1-3}
\cmidrule(lr){4-6}
80.0\% &   0.26 sec & 0.02\%  &  80.0\% &   1.28 sec & 0.15\% \\
85.0\% &   0.25 sec & 0.07\%  &  85.0\% &   1.39 sec & 0.16\% \\
90.0\% &   0.34 sec & 0.10\%  &  90.0\% &   1.39 sec & 0.16\% \\
95.0\% &   0.55 sec & 0.17\%  &  95.0\% &   6.67 sec & 0.76\% \\
96.0\% &   1.18 sec & 0.34\%  &  96.0\% &   9.83 sec & 1.12\% \\
97.0\% &   2.02 sec & 0.58\%  &  97.0\% &  22.85 sec & 2.61\% \\
98.0\% &   3.31 sec & 0.96\%  &  98.0\% &  47.63 sec & 5.44\% \\
99.0\% &  12.19 sec & 3.53\%  &  99.0\% & 161.66 sec & 18.45\% \\
\bottomrule
\end{tabular}

\vspace{4mm}

\begin{tabular}{rrr rrr}
\toprule
\multicolumn{3}{c}{\normalsize \textbf{LR, \criteo}} &
\multicolumn{3}{c}{\normalsize \textbf{LR, \higgs}}  \\
\textbf{Requested Accuracy} & \textbf{Training Time} & \textbf{Ratio} &
\textbf{Requested Accuracy} & \textbf{Training Time} & \textbf{Ratio} \\
\cmidrule(lr){1-3}
\cmidrule(lr){4-6}
80.0\% &   9.14 sec & 0.16\%  &  80.0\% &   0.36 sec & 0.07\% \\
85.0\% &   8.99 sec & 0.16\%  &  85.0\% &   0.36 sec & 0.07\% \\
90.0\% &   9.04 sec & 0.16\%  &  90.0\% &   1.92 sec & 0.36\% \\
95.0\% &  78.93 sec & 1.38\%  &  95.0\% &   4.39 sec & 0.83\% \\
96.0\% &  86.37 sec & 1.51\%  &  96.0\% &   9.21 sec & 1.74\% \\
97.0\% & 115.47 sec & 1.67\%  &  97.0\% &  19.89 sec & 3.75\% \\
98.0\% & 227.56 sec & 2.02\%  &  98.0\% &  53.39 sec & 10.07\% \\
99.0\% & 690.46 sec & 3.97\%  &  99.0\% & 231.08 sec & 43.60\% \\
\bottomrule
\end{tabular}

\vspace{4mm}

\begin{tabular}{rrr rrr}
\toprule
\multicolumn{3}{c}{\normalsize \textbf{ME, \mnist}} &
\multicolumn{3}{c}{\normalsize \textbf{ME, \yelp}}  \\
\textbf{Requested Accuracy} & \textbf{Training Time} & \textbf{Ratio} &
\textbf{Requested Accuracy} & \textbf{Training Time} & \textbf{Ratio} \\
\cmidrule(lr){1-3}
\cmidrule(lr){4-6}
80.0\% & 55.71 sec &  0.16\%   &  80.0\% & 18.41 sec &  1.06\% \\
85.0\% & 55.29 sec &  0.16\%   &  85.0\% & 30.98 sec &  1.79\% \\
90.0\% & 182.00 sec &  0.51\%   &  90.0\% & 70.87 sec &  4.09\% \\
95.0\% & 541.79 sec & 1.53\%   &  95.0\% & 301.86 sec &  17.43\% \\
96.0\% & 816.31 sec & 2.31\%   &  96.0\% & 441.40 sec & 25.48\% \\
97.0\% & 1,361.52 sec & 3.85\%   &  97.0\% & 735.26 sec & 42.45\% \\
98.0\% & 2,935.23 sec & 8.30\%   &  98.0\% & 1,130.53 sec & 65.27\% \\
99.0\% & 8,828.16 sec & 24.97\%   &  99.0\% & 1,589.77 sec & 91.79\% \\
\bottomrule
\end{tabular}

\vspace{4mm}

\begin{tabular}{rrr rrr}
\toprule
\multicolumn{3}{c}{\normalsize \textbf{PPCA, \mnist}} &
\multicolumn{3}{c}{\normalsize \textbf{PPCA, \higgs}}  \\
\textbf{Requested Accuracy} & \textbf{Training Time} & \textbf{Ratio} &
\textbf{Requested Accuracy} & \textbf{Training Time} & \textbf{Ratio} \\
\cmidrule(lr){1-3}
\cmidrule(lr){4-6}
0.9    & 2.20 sec & 6.29\%  &  0.9    &  0.145 sec &  7.25\% \\
0.95   & 2.24 sec & 6.40\%  &  0.95   &  0.144 sec &  7.20\% \\
0.99   & 2.39 sec & 6.83\%  &  0.99   &  0.145 sec &  7.25\% \\
0.995  & 2.65 sec & 7.57\%  &  0.995  &  0.150 sec &  7.50\% \\
0.999  & 4.39 sec & 12.54\%  &  0.999  &  0.167 sec &  8.35\% \\
0.9995 & 6.30 sec & 18.00\%  &  0.9995 &  0.190 sec &  9.50\% \\
0.9999 & 14.54 sec & 41.54\%  &  0.9999 & 0.388 sec & 19.40\% \\
\bottomrule
\end{tabular}
\end{table}

\twocolumn

\pagebreak

\begin{table*}[h]
\caption{This is the raw data for \cref{fig:exp:acc}.}
\label{tab:exp:acc}

\centering
\small
\begin{tabular}{rrrr rrrr}
\toprule
\multicolumn{4}{c}{\normalsize \textbf{Lin, \gas}} &
\multicolumn{4}{c}{\normalsize \textbf{Lin, \power}}  \\
\multirow{2}{*}{\textbf{Requested}} & \multicolumn{3}{c}{\textbf{Actual Accuracy}} &
\multirow{2}{*}{\textbf{Requested}} & \multicolumn{3}{c}{\textbf{Actual Accuracy}} \\
\cmidrule(lr){2-4}
\cmidrule(lr){6-8}
\textbf{Accuracy} & Mean & 5th Percentile & 95th Percentile &
\textbf{Accuracy} & Mean & 5th Percentile & 95th Percentile \\
\cmidrule(lr){1-4}
\cmidrule(lr){5-8}
80.0\% & 97.05\% & 96.04\% & 97.82\% &   80.0\% & 95.06\% & 93.83\% & 96.01\% \\
85.0\% & 97.11\% & 96.55\% & 97.71\% &   85.0\% & 94.84\% & 93.94\% & 95.85\% \\
90.0\% & 97.24\% & 96.03\% & 98.01\% &   90.0\% & 95.18\% & 94.31\% & 96.07\% \\
95.0\% & 97.44\% & 96.85\% & 98.00\% &   95.0\% & 96.28\% & 95.30\% & 96.88\% \\
96.0\% & 97.67\% & 96.98\% & 98.34\% &   96.0\% & 97.16\% & 96.60\% & 97.61\% \\
97.0\% & 98.28\% & 97.43\% & 98.94\% &   97.0\% & 97.83\% & 97.38\% & 98.25\% \\
98.0\% & 98.80\% & 98.45\% & 99.15\% &   98.0\% & 98.54\% & 98.29\% & 98.83\% \\
99.0\% & 99.38\% & 99.24\% & 99.52\% &   99.0\% & 99.25\% & 99.07\% & 99.43\% \\
\bottomrule
\end{tabular}

\vspace{4mm}

\begin{tabular}{rrrr rrrr}
\toprule
\multicolumn{4}{c}{\normalsize \textbf{LR, \criteo}} &
\multicolumn{4}{c}{\normalsize \textbf{LR, \higgs}}  \\
\multirow{2}{*}{\textbf{Requested}} & \multicolumn{3}{c}{\textbf{Actual Accuracy}} &
\multirow{2}{*}{\textbf{Requested}} & \multicolumn{3}{c}{\textbf{Actual Accuracy}} \\
\cmidrule(lr){2-4}
\cmidrule(lr){6-8}
\textbf{Accuracy} & Mean & 5th Percentile & 95th Percentile &
\textbf{Accuracy} & Mean & 5th Percentile & 95th Percentile \\
\cmidrule(lr){1-4}
\cmidrule(lr){5-8}
80.0\% & 96.16\% & 95.33\% & 96.51\% &   80.0\% & 94.11\% & 93.04\% & 95.05\% \\
85.0\% & 96.16\% & 95.33\% & 96.51\% &   85.0\% & 94.71\% & 93.16\% & 96.14\% \\
90.0\% & 96.16\% & 95.33\% & 96.51\% &   90.0\% & 95.98\% & 93.28\% & 96.91\% \\
95.0\% & 97.46\% & 97.13\% & 98.20\% &   95.0\% & 98.11\% & 97.79\% & 98.36\% \\
96.0\% & 97.95\% & 97.24\% & 98.43\% &   96.0\% & 98.58\% & 98.20\% & 98.99\% \\
97.0\% & 98.13\% & 97.64\% & 98.49\% &   97.0\% & 98.99\% & 98.72\% & 99.21\% \\
98.0\% & 98.51\% & 98.07\% & 98.95\% &   98.0\% & 99.32\% & 99.13\% & 99.50\% \\
99.0\% & 98.86\% & 98.42\% & 99.27\% &   99.0\% & 99.71\% & 99.63\% & 99.82\% \\
\bottomrule
\end{tabular}

\vspace{4mm}

\begin{tabular}{rrrr rrrr}
\toprule
\multicolumn{4}{c}{\normalsize \textbf{ME, \mnist}} &
\multicolumn{4}{c}{\normalsize \textbf{ME, \yelp}}  \\
\multirow{2}{*}{\textbf{Requested}} & \multicolumn{3}{c}{\textbf{Actual Accuracy}} &
\multirow{2}{*}{\textbf{Requested}} & \multicolumn{3}{c}{\textbf{Actual Accuracy}} \\
\cmidrule(lr){2-4}
\cmidrule(lr){6-8}
\textbf{Accuracy} & Mean & 5th Percentile & 95th Percentile &
\textbf{Accuracy} & Mean & 5th Percentile & 95th Percentile \\
\cmidrule(lr){1-4}
\cmidrule(lr){5-8}
80.0\% & 94.15\% & 94.14\% & 94.15\% &   80.0\% & 82.44\% & 81.71\% & 83.09\% \\
85.0\% & 94.15\% & 94.14\% & 94.15\% &   85.0\% & 86.00\% & 85.49\% & 86.49\% \\
90.0\% & 95.83\% & 95.77\% & 95.89\% &   90.0\% & 90.19\% & 89.89\% & 90.50\% \\
95.0\% & 97.89\% & 97.87\% & 97.91\% &   95.0\% & 95.29\% & 95.00\% & 95.53\% \\
96.0\% & 98.35\% & 98.35\% & 98.37\% &   96.0\% & 96.50\% & 96.30\% & 96.69\% \\
97.0\% & 98.78\% & 98.77\% & 98.80\% &   97.0\% & 97.54\% & 97.40\% & 97.65\% \\
98.0\% & 99.28\% & 99.26\% & 99.29\% &   98.0\% & 98.52\% & 98.38\% & 98.65\% \\
99.0\% & 99.64\% & 99.63\% & 99.66\% &   99.0\% & 99.43\% & 99.38\% & 99.51\% \\
\bottomrule
\end{tabular}

\vspace{4mm}

\begin{tabular}{rrrr rrrr}
\toprule
\multicolumn{4}{c}{\normalsize \textbf{PPCA, \mnist}} &
\multicolumn{4}{c}{\normalsize \textbf{PPCA, \higgs}}  \\
\multirow{2}{*}{\textbf{Requested}} & \multicolumn{3}{c}{\textbf{Actual Accuracy}} &
\multirow{2}{*}{\textbf{Requested}} & \multicolumn{3}{c}{\textbf{Actual Accuracy}} \\
\cmidrule(lr){2-4}
\cmidrule(lr){6-8}
\textbf{Accuracy} & Mean & 5th Percentile & 95th Percentile &
\textbf{Accuracy} & Mean & 5th Percentile & 95th Percentile \\
\cmidrule(lr){1-4}
\cmidrule(lr){5-8}
0.9    & 0.9815 & 0.9794 & 0.9825   & 0.9    & 0.9931 & 0.9922 & 0.9941 \\
0.95   & 0.9909 & 0.9908 & 0.9912   & 0.95   & 0.9981 & 0.9974 & 0.9989 \\
0.99   & 0.9982 & 0.9980 & 0.9984   & 0.99   & 0.9997 & 0.9996 & 0.9999 \\
0.995  & 0.9989 & 0.9988 & 0.9989   & 0.995  & 0.9998 & 0.9997 & 0.9998 \\
0.999  & 0.9999 & 0.9998 & 0.9999   & 0.999  & 0.9999 & 0.9999 & 0.9999 \\
0.9995 & 0.9999 & 0.9999 & 0.9999   & 0.9995 & 1.0000 & 1.0000 & 1.0000 \\
0.9999 & 1.0000 & 1.0000 & 1.0000   & 0.9999 & 1.0000 & 1.0000 & 1.0000 \\
\bottomrule
\end{tabular}
\end{table*}

\begin{table*}
\caption{This is the raw data for \cref{fig:exp:sss:a}.}
\label{tab:exp:sss:a}

\centering
\small
\begin{tabular}{rrrrr rrrrr}
\toprule
\multicolumn{5}{c}{\normalsize \textbf{Lin, \power}} &
\multicolumn{5}{c}{\normalsize \textbf{LR, \criteo}}  \\
\multirow{2}{*}{\textbf{Requested}} & \multicolumn{4}{c}{\textbf{Actual Accuracy}} &
\multirow{2}{*}{\textbf{Requested}} & \multicolumn{4}{c}{\textbf{Actual Accuracy}} \\
\cmidrule(lr){2-5}
\cmidrule(lr){7-10}
\textbf{Accuracy} & FixedRatio & RelativeRatio & \gradual & \system &
\textbf{Accuracy} & FixedRatio & RelativeRatio & \gradual & \system \\
\cmidrule(lr){1-5}
\cmidrule(lr){6-10}
80.0\% & 96.10\% & 98.68\% & 94.84\% & 95.06\% &   80.0\% & 98.63\% & 99.01\% & 93.95\% & 96.16\% \\
85.0\% & 96.10\% & 98.74\% & 94.84\% & 94.84\% &   85.0\% & 98.63\% & 98.98\% & 93.94\% & 96.16\% \\
90.0\% & 96.10\% & 98.75\% & 94.83\% & 95.18\% &   90.0\% & 98.63\% & 99.11\% & 96.82\% & 96.16\% \\
95.0\% & 96.10\% & 98.85\% & 97.48\% & 96.28\% &   95.0\% & 98.63\% & 98.98\% & 97.00\% & 97.46\% \\
96.0\% & 96.10\% & 98.78\% & 97.48\% & 97.16\% &   96.0\% & 98.63\% & 98.97\% & 97.37\% & 97.95\% \\
97.0\% & 96.10\% & 98.82\% & 98.54\% & 97.83\% &   97.0\% & 98.63\% & 98.99\% & 98.19\% & 98.13\% \\
98.0\% & 96.10\% & 98.79\% & 98.85\% & 98.54\% &   98.0\% & 98.63\% & 99.05\% & 97.99\% & 98.51\% \\
99.0\% & 96.10\% & 98.84\% & 99.29\% & 99.25\% &   99.0\% & 98.63\% & 98.82\% & 99.35\% & 98.86\% \\
\bottomrule
\end{tabular}
\end{table*}

\begin{table*}
\vspace{10mm}
\caption{This is the raw data for \cref{fig:exp:sss:b}.}
\label{tab:exp:sss:b}

\begin{tabular}{rrrrrr}
\toprule
\multicolumn{6}{c}{\normalsize \textbf{Lin, \power}}  \\
\multirow{2}{*}{\textbf{Requested}} & \multicolumn{5}{c}{\textbf{Runtime}} \\
\cmidrule(lr){2-6}
\textbf{Accuracy} & FixedRatio & RelativeRatio & Incestimator & \system & \system's pure training time \\
\cmidrule(lr){1-6}
80.0\% & 2.40 sec & 54.35 sec & 1.79 sec & 1.28 sec & 1.79 sec \\
85.0\% & 2.40 sec & 63.92 sec & 2.38 sec & 1.39 sec & 2.38 sec \\
90.0\% & 2.40 sec & 60.16 sec & 1.86 sec & 1.39 sec & 1.86 sec \\
95.0\% & 2.40 sec & 66.10 sec & 40.07 sec & 6.67 sec & 3.67 sec \\
96.0\% & 2.40 sec & 61.02 sec & 42.41 sec & 9.83 sec & 6.83 sec \\
97.0\% & 2.40 sec & 60.27 sec & 115.09 sec & 22.85 sec & 19.85 sec \\
98.0\% & 2.40 sec & 61.99 sec & 201.80 sec & 47.63 sec & 44.63 sec \\
99.0\% & 2.40 sec & 63.55 sec & 735.97 sec & 161.66 sec & 158.66 sec \\
\bottomrule
\end{tabular}

\vspace{4mm}

\begin{tabular}{rrrrrr}
\toprule
\multicolumn{6}{c}{\normalsize \textbf{LR, \criteo}}  \\
\multirow{2}{*}{\textbf{Requested}} & \multicolumn{5}{c}{\textbf{Runtime}} \\
\cmidrule(lr){2-6}
\textbf{Accuracy} & FixedRatio & RelativeRatio & Incestimator & \system & \system's pure training time \\
\cmidrule(lr){1-6}
80.0\% & 78.16 sec & 700.04 sec & 9.10 sec & 9.14 sec & 9.14 sec \\
85.0\% & 78.16 sec & 710.02 sec & 12.39 sec & 8.99 sec & 8.99 sec \\
90.0\% & 78.16 sec & 773.97 sec & 32.80 sec & 9.04 sec & 9.04 sec \\
95.0\% & 78.16 sec & 721.37 sec & 115.34 sec & 78.93 sec & 18.93 sec \\
96.0\% & 78.16 sec & 709.27 sec & 101.21 sec & 86.37 sec & 26.37 sec \\
97.0\% & 78.16 sec & 812.69 sec & 212.78 sec & 95.45 sec & 35.45 sec \\
98.0\% & 78.16 sec & 786.36 sec & 822.78 sec & 115.47 sec & 55.47 sec \\
99.0\% & 78.16 sec & 719.83 sec & 5703.52 sec & 227.56 sec & 167.56 sec \\
\bottomrule
\end{tabular}

\end{table*}

\begin{table*}
\caption{This is the raw data for \cref{fig:exp:overhead}.}
\label{tab:exp:overhead}

\centering
\small
\begin{tabular}{rrrrrr}
\toprule
\multicolumn{6}{c}{\normalsize \textbf{\system's Runtime Overhead}}  \\
\multirow{2}{*}{\textbf{Number of}} & \multicolumn{4}{c}{\textbf{\system's Runtime}} & \multirow{2}{*}{Full Model} \\
\cmidrule(lr){2-5}
\textbf{Features} & Initial Training & Statistics Computation & Sample Size Searching & Final Training & Training \\
\cmidrule(lr){1-6}
100 & 0.34 sec & 0.02 sec & 0.65 sec & 1.99 sec & 1915 sec \\
500 & 0.64 sec & 0.08 sec & 0.70 sec & 6.29 sec & 4156 sec \\
1K  & 0.71 sec & 0.22 sec & 0.78 sec & 7.20 sec & 6801 sec \\
5K  & 0.59 sec & 1.22 sec & 1.28 sec & 12.38 sec & 1432 sec \\
10K & 0.70 sec & 1.80 sec & 1.67 sec & 8.74 sec & 11307 sec \\
50K & 0.89 sec & 6.92 sec & 3.72 sec & 14.46 sec & 5267 sec \\
100K & 1.36 sec & 11.84 sec & 7.93 sec & 16.06 sec & 5473 sec \\
500K & 1.74 sec & 81.74 sec & 42.77 sec & 16.03 sec & 6578 sec \\
998K & 2.38 sec & 130.84 sec & 84.40 sec & 17.42 sec & 6196 sec \\
\bottomrule
\end{tabular}
\end{table*}

\begin{table*}
\vspace*{10mm}
\caption{This is the raw data for \cref{fig:exp:generror,fig:exp:itrcount}.}
\label{tab:exp:generror}
\centering
\small
\begin{tabular}{rrrr rrr}
\toprule
\multicolumn{4}{c}{\normalsize \textbf{Generalization Error}} &
\multicolumn{3}{c}{\normalsize \textbf{Number of Iterations}}  \\
\textbf{\# of Features} & \textbf{Full Training} & \textbf{\system} & \textbf{Predicted Gen. Error Bound} &
\textbf{\# of Features} & \textbf{Full Training} & \textbf{\system} \\
\cmidrule(lr){1-4}
\cmidrule(lr){5-7}
100 & 26.31 \% & 26.40 \% & 27.12 \%  &  100 & 14 & 13 \\
500 & 21.70 \% & 21.80 \% & 22.45 \%  &  500 & 24 & 27 \\
1K  & 21.01 \% & 21.04 \% & 21.52 \%  &  1K  & 28 & 22 \\
5K  & 20.86 \% & 20.77 \% & 21.26 \%  &  5K  & 24 & 22  \\
10K & 20.83 \% & 20.68 \% & 21.04 \%  &  10K & 22 & 21 \\
50K & 20.84 \% & 20.73 \% & 20.98 \%  &  50K & 27 & 23 \\
100K & 20.83 \% & 20.71 \% & 21.03 \%  &  100K & 24 & 24 \\
\bottomrule
\end{tabular}
\end{table*}